\documentclass[]{fairmeta}
\usepackage{todonotes}
\usepackage{amssymb}
\usepackage{amsfonts}
\usepackage{multicol}
\usepackage{enumitem}
\usepackage{longtable} 
\usepackage{booktabs}
\usepackage{multirow}
\usepackage{enumitem}
\usepackage{tabularx}
\usepackage[table]{xcolor}   % the [table] option enables \rowcolor
\usepackage{graphicx}
\usepackage{subcaption}

\usepackage{xspace}
\newcommand{\glimmermodelname}{Muse Glimmer\xspace}
\newcommand{\teachermodelname}{Muse Spark\xspace}
\newcommand{\gemma}{Gemma4-31B\xspace}
\newcommand{\qwen}{Qwen3.6-27B\xspace}
\newcommand{\dataset}[1]{\textsc{#1}}

\usepackage{listings}
\usepackage{xcolor}
\lstdefinestyle{json}{ basicstyle=\ttfamily\scriptsize, breaklines=true, frame=single, backgroundcolor=\color{gray!5}, numbers=left, numberstyle=\tiny\color{gray}, numbersep=5pt, showstringspaces=false, tabsize=2, columns=fullflexible, }

\title{\glimmermodelname 30B: An On-Device Agentic Model}

\author[1]{Muse Glimmer team}
\affiliation[1]{Meta Superintelligence Labs}

\abstract{We introduce \glimmermodelname, an open-source 30B parameter causal language model with perception encoder, 
purpose-built for autonomous agentic tasks on consumer hardware. \glimmermodelname integrates multi-step reasoning, reliable tool use and failure recovery into a single model that runs locally on consumer devices  
without requiring cloud infrastructure. The model accepts text and image inputs and produces text outputs, was trained on data from over a hundred languages,
and is quantized to approximately 4-bit precision for efficient single-GPU inference. \glimmermodelname achieves state-of-the-art performance for its size class on agentic benchmarks, setting a new standard compared to the models of its size: \gemma and \qwen \color{black}
on tool use, multi-step planning, and agentic task completion\color{black}.  
We release the model under the Apache 2.0 license to foster open-source development of capable, safe, local AI agents.}

\date{\today}
\metadata[Model]{\url{https://huggingface.co/meta-models/Muse-Glimmer-30B}}
\metadata[Blogpost]{\url{https://research.meta.ai/blog/introducing-muse-glimmer-open-agentic-model}}

\begin{document}

\maketitle

\section{Introduction} 
\label{sec:introduction}

Recent large language models have demonstrated strong performance on complex reasoning, program synthesis, and tool-augmented problem solving \citep{meta2025llama4,openai2024gpt4ocard,openai2026openaio1card,Guo_2025}. Work on open-weight models has shown that architectures with substantially fewer parameters can recover much of the performance of their larger counterparts when given sufficient data or targeted optimization \citep{gemmateam2026gemma4technicalreport,qwen2025qwen3,abdin2024phi4}. Nevertheless, many publicly available models aimed at consumer hardware are trained as general-purpose systems and lack the specialization required for reliable autonomous agent behavior, such as sustained multi-step reasoning, robust tool invocation, and principled handling of private user context. This leaves an unaddressed gap between the capabilities of proprietary cloud-hosted systems and those available to developers building local AI agents.

We present \glimmermodelname, a 30-billion parameter language model purpose-built for autonomous agentic tasks suitable for running entirely on consumer hardware. \glimmermodelname is distilled from \teachermodelname, Meta's frontier-class model \citep{menghini2026musesparksafety}, inheriting its reasoning depth and tool-use reliability while being compressed into a form factor that fits on devices that have 32GB of dedicated memory available.

The result is a model that delivers strong intelligence for agentic on-device workloads: multi-step planning, sequential tool invocation, failure recovery, and long-horizon task execution. The model supports $\approx 140$ languages, with robust multilingual knowledge and reasoning and demonstrated multilingual agentic capabilities.

Our main contributions are as follows:
\begin{itemize}
    \item We release \glimmermodelname, a 30B model that sets a new standard for its size class across tool use, multi-step reasoning, and task completion and performs strongly compared with \gemma and \qwen. 
    % \item \todo[inline]{Architecture novelty formulation}

    \item We demonstrate that targeted distillation from a frontier-class teacher can produce a capable small model, achieving good performance with fewer parameters and lower inference compute cost. 
    \item We demonstrate that a dense architecture delivers high capabilities given our memory constraint on the size of the model as compared to mixture-of-experts. 
    \item We show that through a combination of 4-bit quantization and speculative decoding, the model achieves suitable speeds for usage on edge hardware while preserving its quality.
    \item We release artifacts under Apache 2.0: full-precision weights, 4-bit quantized weights, and a DFlash~\citep{chen2026dflash} head for speculation, enabling the community to deploy, fine-tune, and extend the model.
\end{itemize}

\section{Architecture} 
\label{sec:architecture} 

\glimmermodelname is a 30B parameter causal language model augmented with a dedicated perception encoder for multimodal input. The model accepts interleaved text and image tokens and produces text output. We summarize the architecture in Table~\ref{tab:arch} and detail the key design choices below. \begin{table}[h] \centering \caption{Architecture of \glimmermodelname.} \label{tab:arch} \begin{tabular}{ll} \toprule \textbf{Property} & \textbf{Value} \\ \midrule Architecture & Dense Causal Transformer \\ Total parameters & $\sim29.6$B \\ \quad Language model (transformer blocks) & $\sim25.16$B \\ \quad Token embedding (untied) & $\sim1.34$B \\ \quad Output projection (untied) & $\sim1.34$B \\ \quad Perception encoder & $\sim1.8$B \\ Input modalities & Text + Image \\ Output modality & Text \\ Hidden dimension & 6656 \\ Layers & 52 \\ Attention heads (Q / KV) & 32 / 2 (GQA ratio 16:1) \\ Head dimension & 128 \\ FFN type & SwiGLU \\ FFN intermediate dimension & 19{,}968 \\ Sliding window size & 2048 \\ Attention pattern & [Local, Local, Local, Global] repeating \\ Position encoding & RoPE ($\theta = 500{,}000$), local layers only \\ Gated attention & Yes \\ Vocabulary size & 202{,}048 \\ Context length & 128K+ \\ \bottomrule \end{tabular} \end{table}

\subsection{Language model} The language backbone comprises 52 transformer layers with a hidden dimension of 6656, totaling $\sim$25.16B parameters (excluding embedding and output layers). We use untied input and output embeddings over a vocabulary of 202{,}048 tokens, each contributing $\sim$1.34B parameters ($202{,}048 \times 6656$). 
\paragraph{Attention.} Each layer uses grouped-query attention (GQA) with 32 query heads and 2 key-value heads (ratio 16:1) at a head dimension of 128. We employ a sliding window of 2048 tokens in a repeating pattern of three local-attention layers followed by one global-attention layer, bounding KV-cache growth while preserving long-range information flow. Gated attention is applied throughout. Rotary position embeddings (RoPE) with $\theta = 500{,}000$ are applied exclusively to the local attention layers; the global layers rely on the accumulated positional signal from surrounding local layers, making the model extendable.

\paragraph{Feed-forward network.} Each layer's FFN uses SwiGLU activation~\citep{shazeer2020glu} with an intermediate dimension of 19{,}968. The SwiGLU formulation uses three weight matrices---gate projection $W_1 \in \mathbb{R}^{d \times d_{\mathrm{ff}}}$, up projection $W_2 \in \mathbb{R}^{d \times d_{\mathrm{ff}}}$, and down projection $W_3 \in \mathbb{R}^{d_{\mathrm{ff}} \times d}$---yielding an effective expansion ratio of $\sim$3$\times$ the hidden dimension.

\subsection{Multimodal input and the perception encoder}

To support interleaved text-image input, we adopt the Perception Encoder~\citep{bolya2025perception} which leverages a NaViT-style~\citep{dehghani2023patch} vision backbone. The encoder is a $\sim$1.8B-parameter ViT-G/14 with 50 layers, width 1536, and patch size 14 that processes images at native resolution using a 448-px reference grid. Its output tokens are spatially downsampled via a $2\times2$ pixel-shuffle, passed through a two-layer MLP adapter, and then projected into the language model's embedding space by a single bias-free linear layer followed by parameter-free RMSNorm.

Unlike approaches that rely on dynamic tiling with a thumbnail / global crop~\citep{cho2026perceptionlm,liu2024llavanext}, we use a single aspect-ratio-preserving resize to a valid token grid. After the $2\times2$ spatial downsampling, one output visual token corresponds to a $28\times28$ pixel cell originating from $14\times14$ ViT patches. Thus, a $448\times448$ image yields $32\times32=1024$ raw ViT patches and $16\times16=256$ output visual tokens. For higher-resolution images, we increase the grid up to a maximum of 4096 output tokens per image.

We keep the perception encoder frozen throughout training. 

While \glimmermodelname can ingest video as a sequence of frames, it is not explicitly optimized for video. Instead, we optimize for interleaved text and image understanding, including multi-image contexts.
 
\subsection{Design for on-device efficiency}
\label{sec:ondevice}

Our primary design constraint for \glimmermodelname was to operate reliably within a strict 32\,GB VRAM envelope, which is the practical ceiling for consumer-grade hardware.

\begin{figure}
    \centering
    \includegraphics[width=0.8\linewidth]{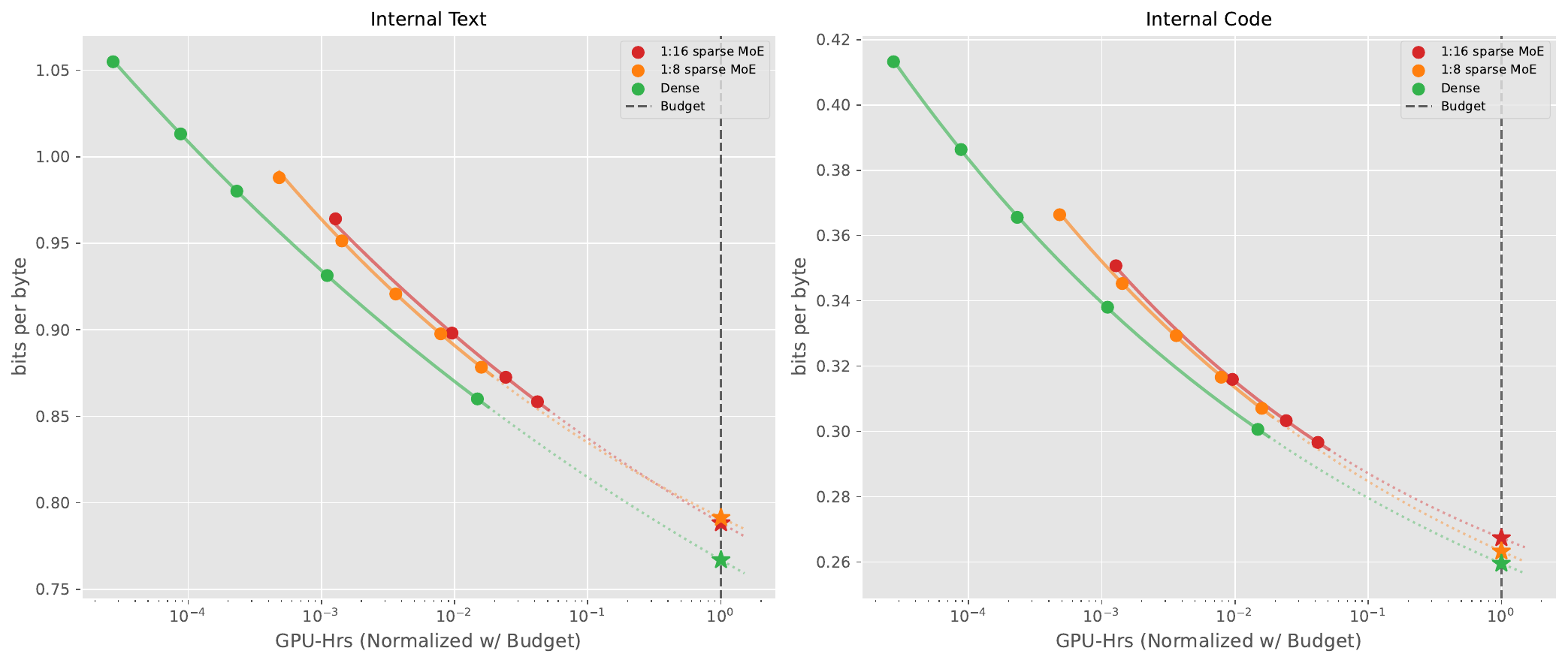}
    \caption{Power law fits of bits-per-byte with respect to GPU-Hrs for the three ladders under consideration (GPU-Hrs is estimated based on final target model MFU). The dense ladder uses 550 TPP, while the MoE ladders use 2,800 TPP (1:8 sparsity) and 6,000 TPP (1:16 sparsity). The total parameter count for the target model in each of these options was chosen such that they can fit into 32\,GB VRAM.}
    \label{fig:ladder}
\end{figure}

To determine the optimal architecture under this hardware limit, we evaluated candidate configurations by training a series of progressively scaled architectural ladders that compared a dense baseline against two Mixture-of-Experts (MoE) designs featuring different routing sparsity (1:8 and 1:16 active experts). We standardized this comparison by keeping the total memory footprint at inference fixed. Under this constant memory ceiling, we estimated the target tokens-per-parameter (TPP) based on our allocated training GPU-hours and the estimated Model FLOP Utilization (MFU) for each configuration. Given our compute budget, all of our ladders operate in over-trained regimes.
Scaling curves (Fig~\ref{fig:ladder}) demonstrated that within constraints of training compute and parameter size, dense configurations yielded a 2$\times$-2.5$\times$ increase in compute efficiency compared to the MoE counterparts. Consequently, we selected a dense 30B parameter design for the architecture.

Beyond static parameter weights, the memory footprint of long-horizon agentic sessions is heavily dominated by the size of the key-value (KV) cache \citep{gemmateam2026gemma4technicalreport}. To mitigate this during multi-step agentic execution, \glimmermodelname incorporates techniques, such as the sliding-window attention to bound KV-cache growth for long contexts, and using grouped-query attention with only 2 KV heads per layer with head dimension of 128. 
To further compress the KV cache, we evaluated a simplified Multi-Head Latent Attention (MLA) variant \citep{liu2024deepseek} in global layers by projecting a single joint key-value vector. We tested two configurations: matching either the baseline’s head count (halving KV cache size) or its total KV footprint (doubling head count). Both underperformed standard Grouped Query Attention (GQA) \citep{ainslie2023gqa} on long-context benchmarks such as RULER \citep{hsieh2024ruler}, leading us to retain GQA.

\section{Model Evaluation} 
\label{sec:model_card} 

\begin{table}[!t]
\centering
\footnotesize
\setlength{\tabcolsep}{5pt}
\renewcommand{\arraystretch}{1.15}
\caption{
\textbf{\glimmermodelname capability scorecard.} We compare \glimmermodelname %\textcolor{red}{(RC2-10)}
at \emph{high} reasoning strength against the competitive open models \gemma and \qwen across agentic, tool-use, reasoning, multimodal, memory, and privacy/safety evaluations. Best capability per row in \textbf{bold}. Higher is better unless marked $\downarrow$. All benchmarks were screened for contamination.
}
\label{tab:main-results}
\begin{tabularx}{\textwidth}{>{\raggedright\arraybackslash}p{3.5cm}l*{3}{>{\centering\arraybackslash}X}}
\toprule
\rowcolor{blue!18}
\multicolumn{5}{c}{\textbf{Capabilities Scorecard}}
\\
\midrule
\rowcolor{blue!8}
\textbf{Domain}
&
\textbf{Evaluation}
&
\textbf{\shortstack{Muse Glimmer\\(high)}}
&
\textbf{\shortstack{\gemma\\(thinking)}}
&
\textbf{\shortstack{\qwen\\(thinking)}}
\\
\midrule
\multirow{9}{=}{Agentic \& Coding}
&
$\tau$³-Banking
&
\textbf{23.5}
&
15.1
&
16.7
\\
&
WildClawBench
&
\textbf{47.6}
&
37.6
&
43.2
\\
&
GAIA2
&
\textbf{43.3}
&
36.4
&
40.0
\\
&
GDPval-AA
&
953
&
811
&
\textbf{1141}
\\
&
SciCode
&
\textbf{43.6}
&
43.4
&
39.8
\\
&
OSWorld-Verified
&
65.9
&
58.5
&
\textbf{75.6}
\\
&
SWE-Bench Verified
&
76.0
&
66.6
&
\textbf{77.2}
\\
&
SWE-Bench Pro
&
\textbf{51.2}
&
36.9
&
50.2
\\
&
TerminalBench 2.1 (Terminus2)
&
51.7
&
43.4
&
\textbf{60.7}
\\
\midrule
Tool Use
&
MCPAtlas
&
\textbf{75.5}
&
54.2
&
62.5
\\
\midrule
\multirow{3}{=}{Reasoning \& Knowledge}
&
DeepSearchQA
&
\textbf{74.6}
&
61.7
&
71.1
\\
&
AIME 2026
&
\textbf{94.7}
&
89.2
&
94.1
\\
&
GPQA Diamond
&
83.5
&
\textbf{85.7}
&
84.2
\\
&
HLE
&
22.0
&
\textbf{23.6}
&
23.1
\\
\midrule
\multirow{2}{=}{Instruction Following}
&
IFBench
&
\textbf{77.0}
&
76.0
&
70.8
\\
&
SkillsBench
&
44.3
&
32.4
&
\textbf{46.6}
\\
\midrule
\multirow{5}{=}{Multimodal}
&
CharXiv Reasoning
&
\textbf{78.8}
&
77.7
&
78.4
\\
&
MMMU-Pro
&
74
&
73
&
\textbf{75}
\\
&
OmniDocBench v1.5
&
75.8
&
72.5
&
\textbf{77.8}
\\
&
Screenspot Pro
&
75.4
&
75.9
&
\textbf{76.1}
\\
\midrule
\multirow{2}{=}{Memory \& Long Context}
&
Beam 128K
&
\textbf{65.1}
&
58.2
&
63.0
\\
&
AA-LCR
&
\textbf{80.0}
&
68.3
&
73.3
\\
\midrule
\multirow{2}{=}{Multilingual}
&
Omnilingual Gaia2
&
\textbf{60.4}
&
37.5
&
45.0
\\
&
Linguini
&
\textbf{8.8}
&
3.8
&
2.5
\\
\midrule
\multirow{4}{=}{Privacy \& Security}
&
CIMemories (violation $\downarrow$)
&
26.4
&
\textbf{12.1}
&
53.4
\\
&
CIMemories (coverage $\uparrow$)
&
64.8
&
53.0
&
\textbf{66.9}
\\
&
Siren AgentDojo (ASR@5 $\downarrow$)
&
28.4
&
\textbf{25.6}
&
40.3
\\
&
Siren AgentDojo (utility $\uparrow$)
&
\textbf{94.2}
&
90.8
&
92.7
\\
\bottomrule
\end{tabularx}
\end{table}

\begin{figure}
    \centering
    \includegraphics[width=0.8\linewidth]{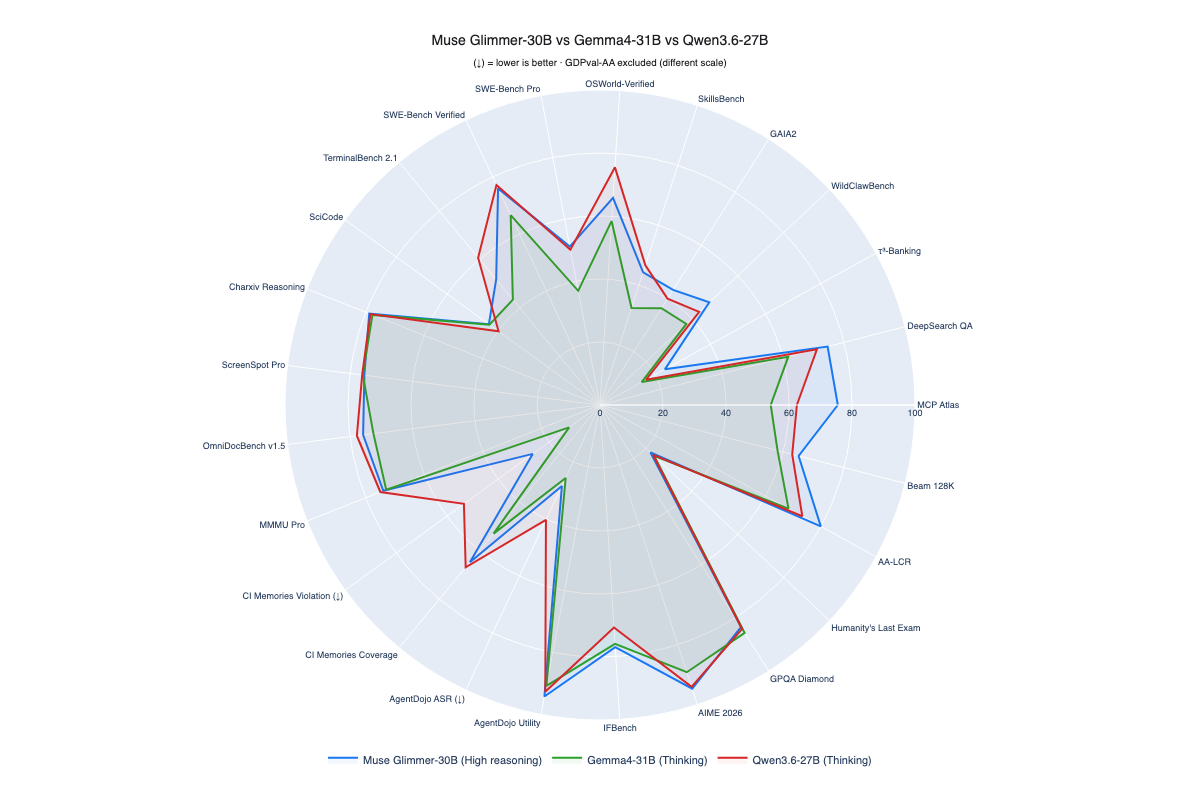}
    \caption{Muse Glimmer capability radar chart. We compare Muse Glimmer at high reasoning strength against the competitive open models Gemma4-31B and Qwen3.6-27B }
    \label{fig:stats}
\end{figure}

Effective autonomous agency requires diverse capabilities. Agentic task completion such as navigating popular scaffolds, writing and debugging code or resolving multi-turn requests requires the model to chain reasoning over long horizons, invoke tools with precise schema, interpret multimodal observations (screenshots, charts, documents), and retain state across sessions. Tool-use and function calling serve as the execution interface through which reasoning is translated into environment actions, while instruction following and steerability determine whether the model can adhere to complex, compositional directives that real-world orchestration systems impose. Long-context memory underpins many of the above: an agent that cannot recall earlier turns may fail at non-trivial multi-step tasks even if it exhibits strong reasoning. We also evaluate multilingual agentic capabilities and a variety of non-agentic covering multilingual reasoning, reading comprehension, knowledge and linguistic competence. 
The model results are presented in~\Cref{tab:main-results} and~\Cref{fig:stats}.
Benchmark descriptions and evaluation details are presented in~\Cref{sec:benchmarks}. 
%Diagram with capabilities needed

\section{Pre-training} 
\label{sec:pre-training}

\subsection{Tokenizer}

\glimmermodelname employs the Llama 4~\citep{meta2025llama4} Byte-Pair Encoding (BPE) tokenizer with a total vocabulary of 202,048 tokens, comprising 200,000 base tokens and 2,048 special tokens. The base vocabulary was trained on a balanced data mix of 35\% code, 35\% multilingual text, 15\% mathematics, and 15\% English, optimized for compression efficiency across diverse domains. 

The special token budget is shared across modalities, unifying text,  vision, reasoning, and agentic capabilities under a single vocabulary space.

\subsection{Pre-training and Mid-training Stages} 
\label{sec:pre-training_stage} 

Training proceeds in multiple stages, a pre-training phase followed by two mid-training phases.  
Each stage serves a distinct purpose: the pre-training phase acquires broad knowledge and reasoning through logit distillation from the teacher; mid-training progressively introduces domain-specific capabilities and extends the effective context length.

\paragraph{Pre-training.}

The base model is trained by logit distillation from \teachermodelname . 

We employ co-distillation, logging the top 256 logits from \teachermodelname during its pre-training. Unlike prior formulations that continuously blend soft and hard targets \citep{meta2025llama4}, our objective consists of only KL divergence for the first 95\% of pre-training. For the final 5\%, we transition to a cross-entropy-only (CE) loss on the ground-truth targets to address the truncation bias inherent to top-$k$ logit distillation \citep[e.g.,][]{anshumann2025sparselogitsamplingaccelerating}.

The pre-training data comprises a large-scale, diverse collection spanning web documents, code, mathematics, multilingual text, and images, with a knowledge cutoff of 2026-01-04.

\paragraph{Checkpoint averaging.}
At the conclusion of pre-training, we apply checkpoint averaging ("souping")~\citep{izmailov2018averaging, wortsman2022model} to produce the initialization for mid-training. Following recent work demonstrating the effectiveness of weight averaging during LLM pre-training~\citep{li2025pma, sanyal2023early, groeneveld2025olmo2}, we save checkpoints every 200 steps during the final 5\% of training (84 checkpoints total) and sweep over merge strategies: uniform average, weighted moving average (WMA), and exponential moving average (EMA)—as well as the number of checkpoints and stride between them. We find that merging over a short window (200-step spacing) consistently outperforms wider strides, and that EMA and WMA provide slightly more stable results than uniform averaging, though the best-performing configuration is similar across merge types. The final souping recipe we chose averages the last checkpoint with three prior checkpoints spaced 600 steps apart using a uniform average. This yields consistent improvements across all perplexity evaluations relative to the single final checkpoint, with larger gains on smoother benchmarks (e.g., ML Papers, Multilingual Documents).

The resulting souped checkpoint serves as the initialization for mid-training Phase~1.

\paragraph{Mid-training Phase 1: Domain infusion.}
Starting from the souped pre-training checkpoint, Phase~1 introduces targeted synthetic and domain-specific data to improve the starting point for post-training. In addition to organic data (similar to that used in pre-training), the data in this phase also contains synthetic data, including reasoning traces and structured problem-solving demonstrations. The learning rate we use for this phase is lower than the peak pre-training learning rate, so as to to balance gains on synthetic-data objectives against acceptable regression on organic perplexity benchmarks.

\paragraph{Mid-training Phase 2: Agentic capabilities.}
The second phase of mid-training extends the context window from 32K to 128K tokens and enriches the model with agentic reasoning data. For this phase we have used a cross-entropy loss, as the distributional shift introduced by long-context and agentic data makes teacher logits less reliable at this stage. The data mix differs from that of the first mid-training phase by including tool-use and agentic trajectory data and purpose-generated agentic demonstrations from our synthetic pipeline (Section~\ref{sec:data_pipeline}). The learning rate for this phase is set to lower than the peak pre-training learning rate.

\section{Post-training} 
\label{sec:posttraining} 

Post-training converts the distilled base into an aligned agent through a sequence of supervised fine-tuning (SFT), reinforcement learning (RL) and on-policy distillation stages described below. 
\begin{enumerate} 
\item \textbf{SFT} Align the base checkpoint to supervised targets: task formatting, instruction following, and reasoning style. 
\item \textbf{On-policy Distillation} We apply on-policy distillation~\cite{agarwal2024policy} where the student generates rollouts and the teacher provides per-token distributional targets on the student's visited prefixes. 
\item \textbf{Multi-objective RL.} Optimize toward reward models, verifiers, and task-success signals: reasoning, agentic, reinforcement learning from human feedback (RLHF). 
\end{enumerate}

\subsection{Supervised Fine-tuning (SFT)}
We use a supervised fine-tuning (SFT) recipe. This stage extends the capabilities acquired during mid-training into a general instruction-following setting. It uses a mixture of synthetic and human-generated data covering reasoning, coding, mathematics, general instruction following, and agentic scaffolds such as tool-use and multi-step task execution (Section \ref{sec:data_pipeline}). 
We train for approximately one epoch on each dataset, prioritizing response formatting, data quality, and coverage over repeated exposure to maintain entropy.

\subsection{Rollouts extraction}
\label{sec:rkld_warmup}
We collected teacher trajectories from \teachermodelname, focusing on reasoning, agentic, and tool-use capabilities across multiple unique RL and RLHF runs. For each source run, we selected a range of training checkpoints and retained trajectories where the reward was higher than a threshold value. We prioritized trajectories generated at later checkpoints as these reflected more developed teacher policies. We retained repeated tasks and prompts when they occurred at different checkpoints or produced different trajectories. Each trajectory was converted into a common conversational training format containing system context, user messages, assistant reasoning, tool interactions, and the final response. This conversion preserved the sequence of model and environment interactions needed to reproduce the teacher's behavior, while operational metadata that was not visible to the model was removed.

\textbf{Filtering and quality controls.} The extracted trajectories underwent extensive cleaning and filtering to ensure high-quality teacher rollouts. We stripped evaluation metadata (e.g., rewards, grader outputs), occasional prompt echoes, and artifacts from unsupported tool interfaces or collection infrastructure. Trajectories containing references that revealed the teacher's model identity were excluded to avoid transferring unintended identity behavior to the student. We then validated each sample against the training conversation schema and the model's input renderer, rejecting malformed message sequences, invalid tool-call structures, unsupported fields, and examples that could not be rendered into the target token sequence. 

\subsection{On-Policy Distillation and Reinforcement Learning} 
\label{sec:distillation} 
\label{sec:rkld} 

In the later stages of post-training we leverage on-policy distillation to train \glimmermodelname from the larger \teachermodelname model \citep{kim2016sequence, gu2024minillm, agarwal2024onpolicy}, together with reinforcement learning. During on-policy distillation, The student, \glimmermodelname as it is being trained, actively samples rollouts. As the student generates these autoregressive trajectories, the teacher evaluates the student's generated distributions step by step to produce fine-grained feedback signals. This setup resolves the train–inference distribution shift by training the model on its own output distribution, teaching it to recover when generation paths drift. We find that this intermediate stage consistently improves downstream RL outcomes relative to proceeding directly from SFT to RL.

\subsection{Reasoning Modes}

\glimmermodelname supports varying reasoning strengths. We trained this capability during RL by splitting per prompt samples into subgroups, where each sub group has a length penalty applied corresponding to its assigned reasoning strength. As applying more thinking leads to higher performance, this may reduce the model's capabilities. We found that including the full spectrum did not harm performance at higher reasoning strengths and in fact had a beneficial regularizing effect on their chain of thought length. On the other hand, applying full spectrum training does provide less exposure to higher strength samples in each step; these primarily drive capability acquisition, subsequently propagated to lower strengths. Consequently, balancing performance vs step count led us to adopt a routing scheme, whereby a certain proportion of prompts are routed solely to higher strengths while the remainder use the full spectrum. 

At inference time, \glimmermodelname supports configurable reasoning depth through a reasoning strength parameter specified in the system prompt. Four levels are defined: \textit{low}, which triggers minimal internal deliberation for simple factual lookup; \textit{medium}, which allocates a moderate thinking budget for tasks requiring some planning; \textit{high}, which enables full chain-of-thought reasoning for complex multi-step problems such as debugging or research; and \textit{xhigh} for maximum capability. The reasoning strength is rendered as a single directive (\texttt{Reasoning strength: <level>}) appended to the system prompt, allowing deployment-time control over the trade-off between latency and reasoning quality without retraining. High reasoning is the default for agentic tasks, where the model must plan tool-use sequences, interpret intermediate results, and backtrack when necessary.

\label{sec:agentic_alignment} 

\subsection{Trustworthy Local Agentic Workflows} 
\label{sec:privacy_safety_training} 
%\todo[inline]{summarise}
We applied SFT and RL to enable \glimmermodelname to better support trustworthy local agentic workflows. 

\paragraph{Appropriate Information Flows:} 
\glimmermodelname is trained to better respect the privacy of information when interacting with third parties on an individual's behalf. The underlying principle behind this notion of privacy is inspired by Contextual Integrity (CI) theory~\citep{nissenbaum2004privacy}. Here, the privacy of information depends upon the context surrounding the exchange, i.e. the sender/recipient of the information, the type of data being considered, and transmission principles governing the exchange. We first introduce the ability to reason about the context of information exchange (e.g., when composing emails, or WhatsApp messages) during the SFT stages, through synthetic traces from \teachermodelname. These traces capture interactions between an intricate network of individuals (see Section~\ref{sec:data_pipeline}); we induce behavior via carefully crafted system prompt instructions (also referred to as prompt distillation\citep{snell2022learning, askell2021general}) during trace generation with \teachermodelname, and remove them from the final trace before training. We also discard traces where, despite system prompt instructions, \teachermodelname either (a) includes unnecessary information, (b) doesn't include necessary information, or (c) doesn't complete the task at hand. Our traces span both simpler non-agentic, single turn settings as well as more complex, multi-turn tool-use settings in mock personal assistant environments. We finally amplify this behavior during RL by penalizing oversharing, while rewarding correct sharing of necessary information. 

\paragraph{Safety via Adversarial Robustness and Principled Refusal:}

As an agentic assistant operating over tool-mediated environments, \glimmermodelname must remain robust to adversarial manipulation while maintaining its ability to complete legitimate tasks. We train \glimmermodelname to exhibit complementary safety behaviors such as resisting indirect prompt injections hidden in tool-returned data, refusing harmful or policy-violating requests, and respecting a principal hierarchy that governs how conflicting instructions are resolved. We leverage the synthetic data generation pipeline from Section~\ref{sec:data_pipeline} to produce agentic safety traces via \teachermodelname, using rejection sampling under a charter-aware constitution to ensure the agent correctly ignores adversarial content, refuses unsafe sub-tasks while completing safe ones, and offers defensive alternatives. We discard traces where the agent either (a) succumbs to an adversarial instruction, (b) over-refuses a legitimate task, or (c) fails to complete the benign objective. Our safety traces span both simpler single-turn settings and complex multi-turn tool-use scenarios across diverse domains. During RL, we further amplify robustness by deterministically penalizing trajectories where the agent executes an attacker-seeded action, while jointly rewarding successful task completion and explicit recognition of untrusted content in the model's chain-of-thought reasoning.

\section{Data} 
\label{sec:pre-training_data} 

\subsection{Pre-training Data} 
Our pre-training dataset is a large-scale, diverse collection spanning a broad range of domains and modalities, including web documents, code, STEM content, multilingual text, and images. We filter each domain using a combination of rule-based heuristics, classifier-based quality scoring, and aggressive deduplication. In addition, we decontaminate the corpus against benchmark and evaluation sets to minimize train–test overlap. The knowledge cutoff date is 2026-01-04. List of supported languages is listed in ~\ref{sec:app:multilingual}.

\subsection{Synthetic Data} 
\label{sec:data_pipeline}

Training LLMs as effective tool-using agents requires high-fidelity trajectories that demonstrate complex reasoning, privacy awareness, and proficiency in tool selection and execution. Given the scalability and resource challenges associated with collecting real-world data, we used a pipeline for generating synthetic supervised fine-tuning data. This pipeline uses an always-on AI agent harness operating within Docker-container with simulated environments that mimics applications such as popular email services, calendar, documents, WhatsApp, code repository management tools and other common CLI tools and access to the real web.

We adhere to the design principles established in Agents Research Environments~\citep{froger2025arescalingagentenvironments}. This design considers a "universe" as a pre-populated digital life—a persona supported by a network of 10–14 contacts, featuring weeks of correspondence, calendar events, documents, and messaging history. By integrating these personas into simulated environments, we obtain cohesive digital lives, ensuring the agent remains agnostic to the simulated nature of its interactions. Tools accessible to the agent utilize the same argument structures and JSON return formats as their real-world counterparts. \footnote{For instance, when an agent executes a query such as {\bf gh issue list --label} critical, the mock environment returns matching issues derived from the synthetic universe definition. Similarly, commands like {\bf wacli send --to "Alice" --message "status update?"} trigger an LLM to generate a contextually appropriate response based on the persona’s profile.}

The synthetic data framework supports complex conversations, modeling user and environment responses, leverage time shifts to represent long running (multiple days) scenarios, multiple types of memory (native, memory embeddings, cli tools) and other features.

{\bf Personas:}
We first generate diverse personas to simulate potential agent users. Each persona starts from sampled demographic and Big Five personality attributes \citep{mccrate1992personality} (age, region, and trait scores), which the model expands into a full profile including occupation, hobbies, and habits.
To maximize diversity beyond temperature-based sampling, we additionally ground a subset of profiles in randomly sampled text, prompting the model to generate a persona plausibly matching the text's author—an approach related to work deriving personas from real text to approximate population-level trait distributions \citep{hu2025population}.
Personas are generated in batches and organized into relationship graphs that ground interactions across email, messaging, and calendar platforms.
See Appendix~\ref{app:personas} for further details.

{\bf Universes:}
For each persona, we materialize a complete suite of mock tool states, packaged within a dedicated directory.
The pipeline selects a realistic subset of applications and populates them in dependency order to ensure that subsequent content—such as emails referencing specific contacts or calendar events involving established participants—remains grounded in the earlier state.
Finally, the pipeline generates all necessary workspace files and memory entries.

{\bf Scenarios:} Once a universe is established, we generate task sequences in two stages. 
First, given a concise universe summary (contacts, recent communications, document history), the model proposes high-level objectives spanning categories such as execution, search, adaptability, and time management—constrained to only the tools and contacts actually present to avoid hallucinating unavailable resources.
Second, these objectives are expanded into runnable scenarios: tasks are rewritten as natural user interactions and any missing setup content (e.g., email threads or calendar invites) is planned as needed.
Finally, verifiable rubrics are generated to evaluate agent performance.

{\bf Post-processing: }
We apply a rigorous post-processing pipeline to ensure dataset quality:
\begin{itemize}
    \item \textbf{Deterministic Quality Filtering}: We remove trajectories that fail to meet baseline standards, such as trajectories where the agent failed to complete the goals, duplicated entries, incoherent text, or "fourth-wall" breaks.
    \item \textbf{Undesirable behavioral masking}: We intentionally disable training on any parts of the trajectories that we judge as “undesirable” even if the agent achieved the prescribed goal. This includes bad tool calls, bad memory search queries, and other cases.
    \item \textbf{Grounding Verification}: We employ a post-hoc judging pipeline to perform aggressive filtering based on grounding checks. Any trajectory exhibiting hallucinations, false successes, or unvalidated outcomes is removed from the final dataset.
\end{itemize}

\section{Inference Optimization}
\label{sec:inference} 

Our focus for inference is on low batch size regime for a range of edge devices, so we primarily cover: weight-only quantization to reduce the memory footprint of the weights and the data transfer during decode; training and deployment of a DFlash drafter for speculative decoding. 

\subsection{Quantization}
\label{sec:quantization}

The weights of \glimmermodelname occupy 55.7\,GB in BF16. We target deployments with 24 or 32\,GB of memory while reserving capacity for the unquantized KV cache, draft model, perception encoder.

We therefore target a language-model weight size below 20\,GB. We compare post-training quantization (PTQ) variants using the change in negative log-likelihood relative to BF16, denoted $\Delta\mathrm{NLL}$, on held-out text and code data. We use $\Delta\mathrm{NLL}$ to select candidates for downstream reasoning and agentic evaluations.

\paragraph{\textbf{Static per-tensor allocations.}}

We start from the popular Q4\_K\_M profile from \texttt{llama.cpp}\footnote{\url{https://github.com/ggml-org/llama.cpp/blob/23e0d70bacaaca1429d365a44aa9e7434f17823b/src/llama.cppv\#L5188}}, where the Value, FFN-down and Output projections are upcasted to higher precision in a subset of the layers. To be more conservative, we apply the upcasting on all layers and we test  different precision formats beyond K-quant from \texttt{llama.cpp}. 

These formats include GPTQ~\citep{frantar2023gptq} and AWQ~\citep{lin2024awq}, where the lower-precision tensors use INT4 and the upcasted tensors use INT6, all with group size 128. We apply the same tensor allocation to floating-point formats, using NVFP4 with 16-weight blocks for the lower-precision tensors and MXFP8 with 32-weight blocks for the upcasted tensors~\citep{abecassis2025pretraining,rouhani2023microscaling}. We calibrate this allocation using our weight-only adaptation of AWQ-Lite from NVIDIA TensorRT Model Optimizer~\citep{nvidia-modelopt}. The calibration uses activation statistics to select rescaling factors across adjacent projections. We fold these factors into the BF16 weights before casting each weight once to its target format, while keeping activations in BF16. Lastly, we include uniform NVFP4 and uniform MXFP8 variants as reference points.

For our \texttt{Q4KM} variant, we keep the upcasting applied only to a subset of layers matching \texttt{llama.cpp}. We make two changes compared to the original scheme: our output projection uses Q5\_K as opposed to Q6\_K, and we use a larger group size 128 for both the Q6\_K tensors and the Q5\_K output projection, compared to 16 for Q6\_K and 32 for Q5\_K. The larger groups avoid the group-size-16 limitation in MLX and reduce scale overhead when converting K-quant weights to the affine representation used by our MLX implementation in ExecuTorch\footnote{\url{https://pytorch.org/blog/fast-ondevice-agentic-ai-with-executorch/}}. Both changes show negligible quality degradation.

\paragraph{\textbf{Dynamic allocations.}}
We also test whether calibration measurements support a more flexible allocation than the static Q4KM schedule. Motivated by the per-tensor assignments of Unsloth Dynamic 2.0,\footnote{\url{https://huggingface.co/unsloth/gemma-4-31B-it-GGUF}} we introduce \texttt{Greedy-KL}, which starts from Q6\_K and lowers tensor precision according to the increase in calibration KL divergence per byte saved. It uses the remaining capacity to move the most sensitive Q4\_K tensors to Q5\_K. For comparison, we adapt \texttt{EvoPress}, an evolutionary method for mixed-precision quantization~\citep{sieberling2025evopress}, to the same Q4\_K, Q5\_K, and Q6\_K search space. EvoPress evolves precision assignments under a memory constraint using KL divergence from BF16 as its search objective.

Both searches expose the query, key, value, and output-gate slices of the fused attention projection independently. The FFN gate and up projections are likewise exposed independently. Together with the attention output, FFN-down, token embedding, and output projection, this gives 418 weight units. No tensor role is fixed at high precision. The winner EvoPress profile assigns 51, 130, and 237 units to Q4\_K, Q5\_K, and Q6\_K, respectively, all with group size 32. The search selects Q4\_K for the token embedding and Q5\_K for the output projection, with a total of 19.65\,GB for the model weights.

\begin{figure}[t!]
  \centering
  \begin{minipage}[t]{0.32\linewidth}
    \centering
    \includegraphics[width=\linewidth]{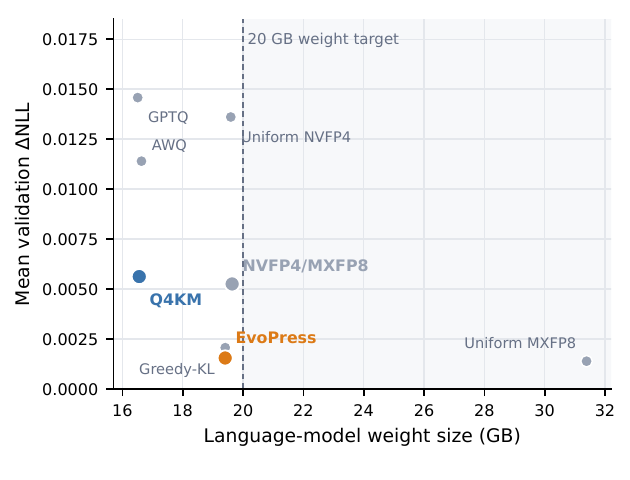}
  \end{minipage}\hfill
  \begin{minipage}[t]{0.32\linewidth}
    \centering
    \includegraphics[width=\linewidth]{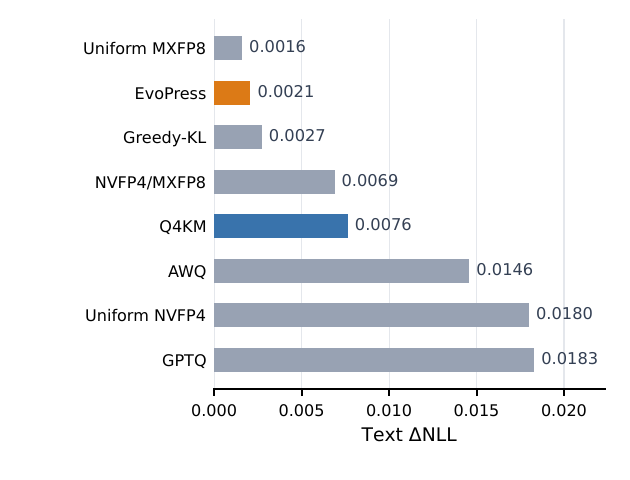}
  \end{minipage}\hfill
  \begin{minipage}[t]{0.32\linewidth}
    \centering
    \includegraphics[width=\linewidth]{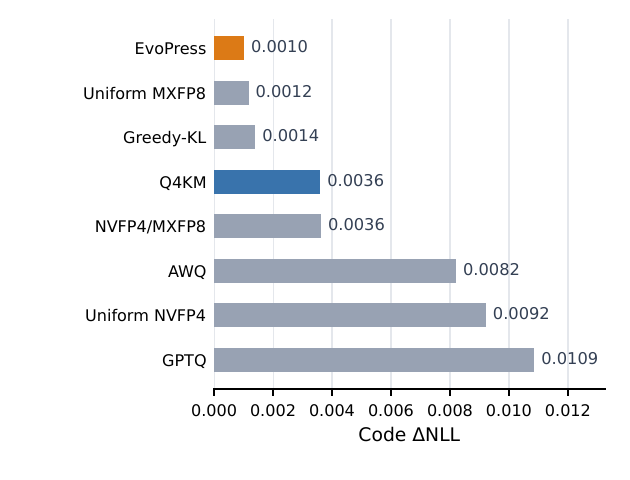}
  \end{minipage}
  \caption{\textbf{Memory and held-out validation quality of the evaluated PTQ methods.} \textbf{(left)} Language-model weight size against mean $\Delta\mathrm{NLL}$ across held-out text and code data. The chosen deployment profiles are highlighted. \textbf{(center--right)} Text and code $\Delta\mathrm{NLL}$ for the same methods, ordered from lowest to highest within each panel. Lower values are better.}
  \label{fig:quantization-memory-quality}
\end{figure}

\paragraph{\textbf{Results.}}
Figure~\ref{fig:quantization-memory-quality} summarizes the memory and validation trade-offs. EvoPress provides the lowest validation loss, and Q4KM provides the smallest memory footprint, so we select these two variants for our 32\,GB and 24\,GB VRAM deployments, respectively.

\paragraph{\textbf{Agentic and reasoning evaluations.}}
We evaluate the candidates on 15 downstream tasks to compare against our NLL-based results. Table~\ref{tab:quant-evals} shows our results.
We see that $\Delta\mathrm{NLL}$ aligns reasonably well with the average accuracy across all tasks combined. 
Related work has shown that aggressive quantization at low bit widths can increase chain-of-thought length and its associated inference cost~\citep{liu2025quantization,lotfi2026quantized}. However due to the conservative nature of our chosen quantization schemes, we observe no significant variation in chain-of-thought length relative to BF16.

\begin{table}[t] \centering 
\caption{\textbf{Downstream accuracy for quantization.} Average accuracy and response length in tokens for different quantized variants across a mix of 15 downstream benchmarks.}\begin{tabular}{@{}lcc@{}} \toprule \textbf{Model Variant} & \textbf{Average Accuracy (\%)} & \textbf{Average Response Length} \\ \midrule BF16 & 75.7 & 11070 \\ EvoPress & 75.6 & 11417 \\ Greedy-KL & 75.3 & 10754 \\ Q4KM & 74.8 & 10774 \\ NVFP4/MXFP8 & 74.4 & 10988 \\ \bottomrule \end{tabular} \label{tab:quant-evals} \end{table}

\subsection{Speculative Decoding}

We adopt DFlash~\citep{chen2026dflash} as our speculative decoding strategy, matching the current open source leading standard. Our drafter
is a lightweight block-diffusion model~\citep{arriola2025blockdiffusion} that
predicts in a single forward pass an entire block of marginal probabilities that are sampled independently to build the draft block. Table~\ref{tab:dflash-config} summarizes our DFlash configuration.

\begin{table}[t]
\centering
\caption{\textbf{DFlash drafter configuration.}}
\begin{tabular}{@{}ll@{}}
\toprule
\textbf{Component} & \textbf{Setting} \\
\midrule
Draft layers          & 5 \\
Block size            & 16 \\
Attention             & Sliding-window, 2048, all layers \\
Attention heads       & 32 query (matches target) / 8 KV (GQA) \\
Sequence length       & 128k \\
Hidden-feature layers & 5, uniform over target: $\{1,13,25,37,49\}$ of 52 \\
\bottomrule
\end{tabular}
\label{tab:dflash-config}
\end{table}

We train the DFlash drafter offline in a supervised manner using on-policy rollouts from the target model, drawn from a combined pool of SFT and RL prompts.
During training, starting anchors for the masked blocks are sampled at a rate of $1/16$ from the response tokens, and the model is optimized using the D-PACE (Dynamic Position-Aware Cross-Entropy) objective~\citep{wu2026dpace}.

\section{Trust Considerations}
\label{sec:trust_considerations} 

We address catastrophic risk assessment, content safety, agentic-specific risks and other trustworthiness considerations through a combination of train-time mitigations built into the model weights.
\subsection{Safety and Privacy Policies}

\glimmermodelname is evaluated and mitigated across three risk axes—content safety, agentic risk (policies for irreversible-action confirmation, data minimization, scaffold boundary respect, and indirect prompt-injection resistance, enforced through dedicated training data), and catastrophic risk (chemical, biological, cyber, and loss-of-control assessments), using both train-time alignment built into the weights and a lightweight inference-time safety mechanism.

\subsection{Train-Time Mitigations}

Safety behavior in \glimmermodelname is established through three stages of post-training:

\begin{enumerate}
\item \textbf{Safety SFT.} Curated examples demonstrating correct safety behavior, including refusal of harmful requests, calibrated responses to borderline prompts, and proper handling of agentic safety scenarios, are included in the supervised fine-tuning data mix. We reuse established alignment recipes while augmenting with \glimmermodelname-specific agentic safety data covering tool-use boundaries, prompt injection resistance, and permission handling.

\item \textbf{Safety RL.} Reinforcement learning stages incorporate safety-specific reward signals that penalize policy violations while rewarding helpful, non-refusing responses to legitimate requests. This stage hillclimbs on content safety evaluations through iterative checkpoint evaluation and mitigation.

\item \textbf{Appropriate information flows} As described in Section~\ref{sec:privacy_safety_training}, principles of data sensitivity recognition, minimization, and local-first execution are embedded directly into the model weights through dedicated synthetic training data, providing defense-in-depth beyond what system prompts alone can achieve.

\end{enumerate}

\section{Related Work}

\paragraph{On-device small language models.}
Running capable language models directly on consumer hardware has been enabled by a wave of compact, openly released model families. 
The Llama model series~\citep{touvron2023llama,touvron2023llama2,grattafiori2024llama3} showed good performance as compared with frontier models while spawning a large open-source ecosystem of fine-tunes, quantization tooling, and local inference engines (llama.cpp, Ollama, LM~Studio). This ecosystem established the practical infrastructure such as GGUF quantization formats, speculative decoding support, Apple Silicon optimizations, on which models like \glimmermodelname now rely for consumer deployment.

Microsoft's \textsc{Phi-4}~\citep{abdin2024phi4} is a 14B-parameter model trained largely on curated synthetic data. Google DeepMind's recent \textsc{Gemma~4}~\citep{gemmateam2026gemma4technicalreport} release spans dense models from 2.3B to 31B parameters (alongside a Mixture-of-Experts variant) and relies on quantization-aware training and KV cache optimizations to fit tight memory and latency budgets. Finally, the \textsc{Qwen3}~\citep{qwen2025qwen3} family offers dense and MoE models spanning 0.6B--235B parameters, unifies fast (non-thinking) and deliberative (thinking) inference within a single model, and extends multilingual coverage to 119 languages. Such model families define the small-LLM landscape against which we position our 30B model.

\paragraph{Teacher-student Distillation.} Distilling the capabilities of large teacher models into smaller, deployable students has emerged as a dominant paradigm for building capable compact models. DeepSeek-R1 \citep{Guo_2025} demonstrated that reinforcement learning alone can elicit sophisticated chain-of-thought reasoning in large models, and that the resulting reasoning traces can then be distilled into much smaller students (1.5B–70B parameters) that substantially outperform models trained from scratch at equivalent scale. The Qwen3 family \citep{qwen2025qwen3} extends this idea through strong-to-weak on-policy distillation, in which the student generates its own rollouts and receives dense token-level KL supervision from a larger teacher, yielding gains over off-policy distillation that merely imitates teacher-generated outputs. Concurrent work on Self-Distilled Reasoner \citep{zhao2025selfdistilledreasoner} and AsyncOPD \citep{kang2025asyncopd} further explores the efficiency frontier of on-policy distillation, showing that student-generated data with teacher feedback is more sample-efficient than traditional supervised fine-tuning. Our post-training recipe (Section ~\ref{sec:distillation}) employs on-policy distillation from the Muse Spark teacher, an approach we found superior to a reinforcement-learning-only recipe for our target model size.

\paragraph{Agentic LLM frameworks.} The ReAct paradigm \citep{yao2023react} established the interleaved reasoning-and-acting loop that underpins most modern LLM agent architectures, showing that language models can effectively chain verbal reasoning with tool invocations to solve multi-step tasks. Subsequent work has focused on training models for reliable tool use rather than relying solely on prompting. Toolformer \citep{schick2023toolformer} introduced self-supervised API-call insertion during pre-training, while Gorilla \citep{patil2023gorilla} demonstrated that fine-tuning on large-scale API documentation enables models to generate correct function calls with high fidelity even as APIs evolve. FireAct \citep{chen2023fireact} and Agent-FLAN \citep{chen2024agentflan} showed that fine-tuning on multi-step agent trajectories produces models that significantly outperform prompted baselines in interactive environments. On the deployment side, open-source orchestration frameworks such as OpenClaw \citep{openclaw2025} have popularized locally-hosted agents that chain model inference with tool execution on consumer hardware, creating demand for compact models that can sustain reliable sequential tool invocation across long sessions. Coding agents like SWE-agent \citep{yang2024sweagent} and OpenHands \citep{wang2024openhands} demonstrate that agentic tool use in software engineering requires sustained multi-step planning and failure recovery—capabilities that our model targets through its agentic alignment training (Section ~\ref{sec:data_pipeline}).

\section{Conclusion} 
\label{sec:conclusion} 

This technical report presents \glimmermodelname, a 30B dense model purpose-built for autonomous agentic operation on consumer hardware. Through targeted distillation from \teachermodelname, it surpasses comparably-sized open models on agentic benchmarks. Combined with 4-bit quantization and speculative decoding via a Dflash head, \glimmermodelname runs at practical speeds on a single RTX 5090 with minimal quality loss. Multimodal understanding of screenshots, documents, and charts equips the model to operate natively in agentic and information-rich environments. Muse Glimmer covers over a hundred language varieties and leads comparably-sized open models up to 27 and 23 pass@1 points on multilingual agentic evaluation.

All artifacts including model weights and the speculation head
are released under Apache 2.0.

\clearpage
\newpage

\section{Authors List}
\subsection{Core Contributors}
Arash Ebrahim Abtahizadeh, Muna Aghamelu, Adil Ahmed, Juan C. Angeles Acuna, Alexis Audran-Reiss, Asaf Avrahamy, Yoram Bachrach, Can Balioglu, Amar Budhiraja, Onur Çelebi, Gaurav Chaurasia, Joy Chen, Lynn Cherif, Lucia Cipolina-Kun, Gal Cohen, Timothée Darcet, Beto de Paola, Digant Desai, Sam Devlin, Derek Dunfield, Ivan Evtimov, Maryam Fazel-Zarandi, Rob Fergus, Jean-Christophe Gagnon-Audet, Grant Gardner, Bhavul Gauri, Paris Giampouras, Sujan Kumar Gonugondla, Antoine Grosnit, Young Han, Michael Hassid, Monuril Islam, Masoud Jalili Sabet, Songhao Jia, Narine Kokhlikyan, Svetlana Kurtsman, Sandra Lefdal, Chee Hau Leow, Noam Levi, Mike Lewis, Sanae Lotfi, Xiyang Luo, Despoina Magka, Saeed Mahloujifar, Neal Mangaokar, Ishita Mediratta, Alexander H. Miller, Hela Momand, Saba Nazir, Kelvin Niu, Mattia Opper, Parth Pathak, Alberto Pepe, Danielle Pintz, Mikhail Plekhanov, Sidhartha Reddy Potu, Anton Protopopov, Shengyi Qian, Shashank Rajput, Joseph Rocca, Scott Roy, Luca Sbordone, Amin Shakeri, Tatiana Shavrina, Michael Shvartsman, Ruan Silva, Anya Sims, Eric Michael Smith, Hugo Touvron, Mabel Ubong, Isabel Urrego, Yanzheng Xiang, Zhiyuan Yao, Alex Yarmula, He Ye, Jade Yu, Luke Zettlemoyer, Arman Zharmagambetov

\subsection{Contributors}
Perpetua Fathima Ahlijah, Tremo Ahmed, Enrique Alfonseca, Pedram Bashiri, Nikolay Bashlykov, Amine Benhalloum, Aaron Berman, Rohith Bhatia, Siddharth Bhattacharya, Prashant Bhende, Chloe Bi, Shuchao Bi, Clement Anthonioz Blanc, Allen Bolourchi, Iva Simon-Bubalo, Pere-Lluís Huguet Cabot, Andrea Caciolai, Erica Cai, Xiang "Tony" Cao, Yue Cao, Andrew Caples, Jinzhao Chang, Lele Chen, Moya Chen, Weijie Chen, Wenhu Chen, Xiaobai Chen, Yinyin Chen, Zhengxing Chen, Wei Cheng, Pavel Chikulaev, Edward Chou, Sy Choudhury, Brian Clerkin, Marta R. Costa-jussà, Hanjun Dai, Curtis Davis, Myra Deng, Tiffany Siqi Deng, Dujian Ding, Hantian Ding, Shuangrui Ding, Jesse Dodge, Carl Domond, Shweta Doshi, Shirin Eghtesadi, Faith Eischen, Athyu Eleti, Jordi Armengol-Estape, Alex Fabbri, Han Fang, Paula Farani, Aiman Farooq, Darius Feher, Christoph Feichtenhofer, Amanda Felix, Romain Froger, Carl Chengyan Fu, Sean Gahagan, Yifan Gao, Itai Gat, Jannes Gladrow, Akashdeep Goel, Ahuva Goldstand, Bo He, David Heineman, Dylan Herman, Irene Hjelt, Joonas Hjelt, Alexander Holz, Rui Hou, Annie Hu, Yuan-Ting Hu, Bernie Huang, Jing Huang, Kun Huang, Ashley Hunter, Srinivasan Iyer, Somya Jain, Bei Jia, Michael Jiang, Song Jiang, Wenwen Jiang, Lin Jin, Jacob Kahn, Melanie Kambadur, Yingxin Kang, Justine Kao, Stephane Kasriel, Hossein Kavianihamedani, Madian Khabsa, Ameya Khandekar, Yash Khare, Gefen Kohavi, Kavya Kopparapu, Praveen Krishnan, Venu Kuchibhotla, Ananya Kumar, Leona Lan, Jack Lanchantin, Hunter Lang, Nikolay Pavlovich Laptev, Yuanzhi Li, Yunlu Li, Asad Liaqat, Johnson Lin, Runwei Lin, Zhichao Lin, Shaun Lindsay, Yaron Lipman, Frederick Liu, Mengchen Liu, Qiang Liu, Yinghai Lu, Yujie Lu, Christopher Luc, Faye Ma, Pingchuan Ma, Tengyu Ma, Wufei Ma, Xiaodong Ma, Damir Maksyutov, Josh Malone, Udeep Manchanda, Thomas Mann, Kalyani Marathe, Akhil Mathur, Roberto-Rafael Maura-Rivero, Ruslan Mavlyutov, Thomas Mesnard, Gregoire Mialon, Julian Michael, Sagar Miglani, Todor Mihaylov, Naureen Mitha, Archi Mitra, Leo Mok, Naila Murray, Muthukumaravel Muthuraman, Mergen Nachin, Nese Demir Nalci, Saaketh Narayan, Alyssa Newcomb, Peter Ney, Joycelyn Ng, Devin Norder, Maxwell Nye, Rajvardhan Oak, Cody Ohlsen, Artidoro Pagnoni, Yi Pan, Nilesh Pandey, Guan Pang, Richard Yuanzhe Pang, Mahesh Pasupuleti, Parth Patel, Yash Patel, Vladan Petrovic, Leslie St. Pierre, Norman Ponte, Yundi Qian, Zengyi Qin, Jack Rae, Ramya Raghavendra, Emmanuel Ramirez, Nikhila Ravi, Hongyu Ren, Riandy Riandy, Harrison Rudolph, Kalyan Saladi, Eduardo Sánchez, Karthik Abinav Sankararaman, Alan Schelten, Thomas Scialom, Aman Shankar, Paresh Shenoy, Alana Shine, Hamid Shojanazeri, Mitesh Kumar Singh, Pawan Sisodiya, Ali Sol, Sten Sootla, DiJia Su, Suraj Subramanian, Brian Sun, Tony Sun, Zhiqing Sun, Natacha Supper, Jason Talpalar, Ellen Xiaoqing Tan, Liang Tan, Buğra Tekin, Jean-Christophe Testud, Parth Thakkar, Joseph Tighe, Ty Toledano, T Tse, Igor Tufanov, Alex Vaughan, Josh Walters, Alex Wang, Bethany Wang, Meng Wang, Rocky Wang, Weiyue Wang, Xiaolan Wang, Xin Wang, Zijian Wang, Bram Wasti, Quentin Wellens, Yeming Wen, Mary Williamson, Yuncheng Wu, Fei Xia, Tian Xie, Anlu Xing, David Xu, Jacob Xu, Puxin Xu, Zheng Xu, Shen Yan, Megan Yang, Yining Yang, Nick Yannopoulos, Kevin Yao, Michihiro Yasunaga, Haoxuan You, Michael You, Dingli Yu, Tianhe Yu, Yue Yu, Summer Yue, Xiang Yue, Shuwei Zeng, Hanwen Zha, Chen Zhang, Fred Zhang, Hongming Zhang, Kaiyu Zhang, Meng Zhang, Mingyang Zhang, Yi Zhang, Shengjia Zhao, Chao Zhou, Claire Zhou, Shuyan Zhou, Yanjun Zhou, Yaxuan Zhou, Jiacheng Zhu, Rosalie Zhu, Yun Zhu
% Shengyi: Tianbao Xie wants to use T Tse as the preferred name

\clearpage
\newpage
\bibliographystyle{assets/plainnat}
\bibliography{paper}

\begin{thebibliography}{93}
\providecommand{\natexlab}[1]{#1}
\providecommand{\url}[1]{\texttt{#1}}
\expandafter\ifx\csname urlstyle\endcsname\relax
  \providecommand{\doi}[1]{doi: #1}\else
  \providecommand{\doi}{doi: \begingroup \urlstyle{rm}\Url}\fi

\bibitem[Abdin et~al.(2024)Abdin, Aneja, Behl, Bubeck, Eldan, Gunasekar,
  Harrison, Hewett, Javaheripi, Kauffmann, et~al.]{abdin2024phi4}
Marah Abdin, Jyoti Aneja, Harkirat Behl, S{\'e}bastien Bubeck, Ronen Eldan,
  Suriya Gunasekar, Michael Harrison, Russell~J Hewett, Mojan Javaheripi, Piero
  Kauffmann, et~al.
\newblock Phi-4 technical report.
\newblock \emph{arXiv preprint arXiv:2412.08905}, 2024.

\bibitem[Abecassis et~al.(2025)Abecassis, Agrusa, Ahn, Alben, Alborghetti,
  Andersch, Arayandi, Bjorlin, Blakeman, Briones,
  et~al.]{abecassis2025pretraining}
Felix Abecassis, Anjulie Agrusa, Dong Ahn, Jonah Alben, Stefania Alborghetti,
  Michael Andersch, Sivakumar Arayandi, Alexis Bjorlin, Aaron Blakeman, Evan
  Briones, et~al.
\newblock Pretraining large language models with nvfp4.
\newblock \emph{arXiv preprint arXiv:2509.25149}, 2025.

\bibitem[Agarwal et~al.(2024{\natexlab{a}})Agarwal, Vieillard, Zhou, Stanczyk,
  Garea, Geist, and Bachem]{agarwal2024onpolicy}
Rishabh Agarwal, Nino Vieillard, Yongchao Zhou, Piotr Stanczyk, Sabela~Ramos
  Garea, Matthieu Geist, and Olivier Bachem.
\newblock On-policy distillation of language models: Learning from
  self-generated mistakes.
\newblock In \emph{International Conference on Learning Representations},
  volume 2024, pages 21246--21263, 2024{\natexlab{a}}.

\bibitem[Agarwal et~al.(2024{\natexlab{b}})Agarwal, Vieillard, Zhou, Stanczyk,
  Ramos~Garea, Geist, and Bachem]{agarwal2024policy}
Rishabh Agarwal, Nino Vieillard, Yongchao Zhou, Piotr Stanczyk, Sabela
  Ramos~Garea, Matthieu Geist, and Olivier Bachem.
\newblock On-policy distillation of language models: Learning from
  self-generated mistakes.
\newblock In \emph{International Conference on Learning Representations},
  volume 2024, pages 21246--21263, 2024{\natexlab{b}}.

\bibitem[Ainslie et~al.(2023)Ainslie, Lee-Thorp, De~Jong, Zemlyanskiy,
  Lebr{\'o}n, and Sanghai]{ainslie2023gqa}
Joshua Ainslie, James Lee-Thorp, Michiel De~Jong, Yury Zemlyanskiy, Federico
  Lebr{\'o}n, and Sumit Sanghai.
\newblock Gqa: Training generalized multi-query transformer models from
  multi-head checkpoints.
\newblock In \emph{Proceedings of the 2023 Conference on Empirical Methods in
  Natural Language Processing}, pages 4895--4901, 2023.

\bibitem[Andrews et~al.(2025)Andrews, Artetxe, Meglioli, Costa-juss{\`a},
  Chuang, Dale, Duppenthaler, Ekberg, Gao, Licht, Maillard, Mourachko, Ropers,
  Saleem, S{\'a}nchez, Tsiamas, Turkatenko, Ventayol-Boada, and Yates]{bouquet}
Pierre Andrews, Mikel Artetxe, Mariano~Coria Meglioli, Marta~R.
  Costa-juss{\`a}, Joe Chuang, David Dale, Mark Duppenthaler, Nathanial~Paul
  Ekberg, Cynthia Gao, Daniel~Edward Licht, Jean Maillard, Alexandre Mourachko,
  Christophe Ropers, Safiyyah Saleem, Eduardo S{\'a}nchez, Ioannis Tsiamas,
  Arina Turkatenko, Albert Ventayol-Boada, and Shireen Yates.
\newblock {BOUQ}u{ET} : dataset, benchmark and open initiative for universal
  quality evaluation in translation.
\newblock In Christos Christodoulopoulos, Tanmoy Chakraborty, Carolyn Rose, and
  Violet Peng, editors, \emph{Proceedings of the 2025 Conference on Empirical
  Methods in Natural Language Processing}, pages 27515--27535, Suzhou, China,
  November 2025. Association for Computational Linguistics.
\newblock ISBN 979-8-89176-332-6.
\newblock \doi{10.18653/v1/2025.emnlp-main.1400}.
\newblock \url{https://aclanthology.org/2025.emnlp-main.1400/}.

\bibitem[Anshumann et~al.(2025)Anshumann, Zaidi, Kedia, Ahn, Kwon, Lee, Lee,
  and Lee]{anshumann2025sparselogitsamplingaccelerating}
Anshumann, Mohd~Abbas Zaidi, Akhil Kedia, Jinwoo Ahn, Taehwak Kwon, Kangwook
  Lee, Haejun Lee, and Joohyung Lee.
\newblock Sparse logit sampling: Accelerating knowledge distillation in llms,
  2025.
\newblock \url{https://arxiv.org/abs/2503.16870}.

\bibitem[Arriola et~al.(2025)Arriola, Gokaslan, Chiu, Yang, Qi, Han, Sahoo, and
  Kuleshov]{arriola2025blockdiffusion}
Marianne Arriola, Aaron Gokaslan, Justin~T. Chiu, Zhihan Yang, Zhixuan Qi,
  Jiaqi Han, Subham~Sekhar Sahoo, and Volodymyr Kuleshov.
\newblock Block diffusion: Interpolating between autoregressive and diffusion
  language models, 2025.
\newblock \url{https://arxiv.org/abs/2503.09573}.

\bibitem[Askell et~al.(2021)Askell, Bai, Chen, Drain, Ganguli, Henighan, Jones,
  Joseph, Mann, DasSarma, Elhage, Hatfield-Dodds, Hernandez, Kernion, Ndousse,
  Olsson, Amodei, Brown, Clark, McCandlish, Olah, and
  Kaplan]{askell2021general}
Amanda Askell, Yuntao Bai, Anna Chen, Dawn Drain, Deep Ganguli, Tom Henighan,
  Andy Jones, Nicholas Joseph, Ben Mann, Nova DasSarma, Nelson Elhage, Zac
  Hatfield-Dodds, Danny Hernandez, Jackson Kernion, Kamal Ndousse, Catherine
  Olsson, Dario Amodei, Tom Brown, Jack Clark, Sam McCandlish, Chris Olah, and
  Jared Kaplan.
\newblock A general language assistant as a laboratory for alignment.
\newblock \emph{arXiv preprint arXiv:2112.00861}, 2021.

\bibitem[Bandarkar et~al.(2024)Bandarkar, Liang, Muller, Artetxe, Shukla, Husa,
  Goyal, Krishnan, Zettlemoyer, and Khabsa]{belebele}
Lucas Bandarkar, Davis Liang, Benjamin Muller, Mikel Artetxe, Satya~Narayan
  Shukla, Donald Husa, Naman Goyal, Abhinandan Krishnan, Luke Zettlemoyer, and
  Madian Khabsa.
\newblock The belebele benchmark: a parallel reading comprehension dataset in
  122 language variants.
\newblock In Lun-Wei Ku, Andre Martins, and Vivek Srikumar, editors,
  \emph{Proceedings of the 62nd Annual Meeting of the Association for
  Computational Linguistics (Volume 1: Long Papers)}, pages 749--775, Bangkok,
  Thailand, August 2024. Association for Computational Linguistics.
\newblock \doi{10.18653/v1/2024.acl-long.44}.
\newblock \url{https://aclanthology.org/2024.acl-long.44/}.

\bibitem[Bandi et~al.(2026)Bandi, Dumitru, Hertzberg, Agarwal, Boo, Polakam,
  Hassaan, Da, Kim, Gupta, et~al.]{bandi2026mcp}
Chaithanya Bandi, Razvan-Gabriel Dumitru, Ben Hertzberg, Divyansh Agarwal,
  Geobio Boo, Tejas Polakam, Sami Hassaan, Jeff Da, HiJae Kim, Vipul Gupta,
  et~al.
\newblock Mcp-atlas: A large-scale benchmark for tool-use competency with real
  mcp servers.
\newblock \emph{arXiv preprint arXiv:2602.00933}, 2026.

\bibitem[Bolya et~al.(2025)Bolya, Huang, Sun, Cho, Madotto, Wei, Ma, Zhi,
  Rajasegaran, Rasheed, Wang, Monteiro, Xu, Dong, Ravi, Li, Doll{\'a}r, and
  Feichtenhofer]{bolya2025perception}
Daniel Bolya, Po-Yao Huang, Peize Sun, Jang~Hyun Cho, Andrea Madotto, Chen Wei,
  Tengyu Ma, Jiale Zhi, Jathushan Rajasegaran, Hanoona Rasheed, Junke Wang,
  Marco Monteiro, Hu~Xu, Shiyu Dong, Nikhila Ravi, Daniel Li, Piotr Doll{\'a}r,
  and Christoph Feichtenhofer.
\newblock Perception encoder: The best visual embeddings are not at the output
  of the network.
\newblock In \emph{Advances in Neural Information Processing Systems},
  volume~38, 2025.

\bibitem[Caciolai et~al.(2026)Caciolai, Cabot, Cheng, Ventayol-Boada, Gonzalez,
  Ropers, Bandarkar, Ruder, Sakakihara, Yun, Andrews, Mialon, Froger, and
  Costa-jussà]{omnigaia2}
Andrea Caciolai, Pere-Lluís~Huguet Cabot, Chierh Cheng, Albert Ventayol-Boada,
  Gabriel~Mejia Gonzalez, Christophe Ropers, Lucas Bandarkar, Sebastian Ruder,
  Darlene Sakakihara, Elliot Yun, Pierre Andrews, Grégoire Mialon, Romain
  Froger, and Marta~R. Costa-jussà.
\newblock Omnilingualgaia2: Evaluating the multilingual gap in frontier ai
  agents, 2026.
\newblock \url{https://arxiv.org/abs/2608.08775}.

\bibitem[{Center for AI Safety} et~al.(2026){Center for AI Safety}, {Scale AI},
  and {HLE Contributors Consortium}]{phan2025lastexam}
{Center for AI Safety}, {Scale AI}, and {HLE Contributors Consortium}.
\newblock A benchmark of expert-level academic questions to assess {AI}
  capabilities.
\newblock \emph{Nature}, 649:\penalty0 1139--1146, 2026.
\newblock \doi{10.1038/s41586-025-09962-4}.
\newblock \url{https://arxiv.org/abs/2501.14249}.

\bibitem[Chen et~al.(2023)Chen, Shu, Shareghi, Collier, Narasimhan, and
  Yao]{chen2023fireact}
Baian Chen, Chang Shu, Ehsan Shareghi, Nigel Collier, Karthik Narasimhan, and
  Shunyu Yao.
\newblock Fireact: Toward language agent fine-tuning.
\newblock \emph{arXiv preprint arXiv:2310.05915}, 2023.

\bibitem[Chen et~al.(2026)Chen, Liang, and Liu]{chen2026dflash}
Jian Chen, Yesheng Liang, and Zhijian Liu.
\newblock Dflash: Block diffusion for flash speculative decoding, 2026.
\newblock \url{https://arxiv.org/abs/2602.06036}.

\bibitem[Chen et~al.(2024)Chen, Du, Chen, Liu, Yang, Liu, and
  Sun]{chen2024agentflan}
Zehui Chen, Kuikun Du, Wenquan Chen, Yining Liu, Zhengkai Yang, Zhiyuan Liu,
  and Maosong Sun.
\newblock Agent-flan: Designing data and methods of effective agent tuning for
  large language models.
\newblock \emph{arXiv preprint arXiv:2403.12881}, 2024.

\bibitem[Cho et~al.(2026)Cho, Madotto, Mavroudi, Afouras, Nagarajan, Maaz,
  Song, Ma, Hu, Jain, et~al.]{cho2026perceptionlm}
Jang~Hyun Cho, Andrea Madotto, Effrosyni Mavroudi, Triantafyllos Afouras,
  Tushar Nagarajan, Muhammad Maaz, Yale Song, Tengyu Ma, Shuming Hu, Suyog
  Jain, et~al.
\newblock Perceptionlm: Open-access data and models for detailed visual
  understanding.
\newblock \emph{Advances in Neural Information Processing Systems},
  38:\penalty0 23475--23537, 2026.

\bibitem[Chollet(2019)]{chollet2019arc}
Fran{\c{c}}ois Chollet.
\newblock On the measure of intelligence.
\newblock \emph{arXiv preprint arXiv:1911.01547}, 2019.

\bibitem[De~Benedetti et~al.(2024)De~Benedetti, Carlini, and
  Tram{\`e}r]{debenedetti2024agentdojo}
Edoardo De~Benedetti, Nicholas Carlini, and Florian Tram{\`e}r.
\newblock Agentdojo: A dynamic environment to evaluate prompt injection attacks
  and defenses for llm agents.
\newblock \emph{arXiv preprint arXiv:2406.13352}, 2024.

\bibitem[Dehghani et~al.(2023)Dehghani, Mustafa, Djolonga, Heek, Minderer,
  Caron, Steiner, Puigcerver, Geirhos, Alabdulmohsin, Oliver, Padlewski,
  Gritsenko, Lu{\v{c}}i{\'{c}}, and Houlsby]{dehghani2023patch}
Mostafa Dehghani, Basil Mustafa, Josip Djolonga, Jonathan Heek, Matthias
  Minderer, Mathilde Caron, Andreas Steiner, Joan Puigcerver, Robert Geirhos,
  Ibrahim Alabdulmohsin, Avital Oliver, Piotr Padlewski, Alexey Gritsenko,
  Mario Lu{\v{c}}i{\'{c}}, and Neil Houlsby.
\newblock Patch n' pack: Navit, a vision transformer for any aspect ratio and
  resolution.
\newblock In \emph{Advances in Neural Information Processing Systems},
  volume~36, pages 2252--2274, 2023.

\bibitem[Dekoninck et~al.(2026)Dekoninck, Jovanović, Gehrunger, Rögnvaldsson,
  Petrov, Sun, and Vechev]{dekoninck2026matharena}
Jasper Dekoninck, Nikola Jovanović, Tim Gehrunger, Kári Rögnvaldsson, Ivo
  Petrov, Chenhao Sun, and Martin Vechev.
\newblock Beyond benchmarks: Matharena as an evaluation platform for
  mathematics with llms.
\newblock 2026.
\newblock \url{https://arxiv.org/abs/2605.00674}.

\bibitem[Deutsch et~al.(2025)Deutsch, Briakou, Caswell, Finkelstein, Galor,
  Juraska, Kovacs, Lui, Rei, Riesa, Rijhwani, Riley, Salesky, Trabelsi,
  Winkler, Zhang, and Freitag]{deutsch-etal-2025-wmt24}
Daniel Deutsch, Eleftheria Briakou, Isaac~Rayburn Caswell, Mara Finkelstein,
  Rebecca Galor, Juraj Juraska, Geza Kovacs, Alison Lui, Ricardo Rei, Jason
  Riesa, Shruti Rijhwani, Parker Riley, Elizabeth Salesky, Firas Trabelsi,
  Stephanie Winkler, Biao Zhang, and Markus Freitag.
\newblock {WMT}24++: Expanding the language coverage of {WMT}24 to 55 languages
  {\&} dialects.
\newblock In Wanxiang Che, Joyce Nabende, Ekaterina Shutova, and Mohammad~Taher
  Pilehvar, editors, \emph{Findings of the Association for Computational
  Linguistics: ACL 2025}, pages 12257--12284, Vienna, Austria, July 2025.
  Association for Computational Linguistics.
\newblock ISBN 979-8-89176-256-5.
\newblock \doi{10.18653/v1/2025.findings-acl.634}.
\newblock \url{https://aclanthology.org/2025.findings-acl.634/}.

\bibitem[Ding et~al.(2026)Ding, Dai, Xing, Ding, Liu, JingYi, Yang, Zhang, Wei,
  Fang, Ma, Duan, Shao, Wang, Lin, Chen, and
  Zang]{ding2026wildclawbenchbenchmarkrealworldlonghorizon}
Shuangrui Ding, Xuanlang Dai, Long Xing, Shengyuan Ding, Ziyu Liu, Yang JingYi,
  Penghui Yang, Zhixiong Zhang, Xilin Wei, Xinyu Fang, Yubo Ma, Haodong Duan,
  Jing Shao, Jiaqi Wang, Dahua Lin, Kai Chen, and Yuhang Zang.
\newblock Wildclawbench: A benchmark for real-world, long-horizon agent
  evaluation, 2026.
\newblock \url{https://arxiv.org/abs/2605.10912}.

\bibitem[Frantar et~al.(2023)Frantar, Ashkboos, Hoefler, and
  Alistarh]{frantar2023gptq}
Elias Frantar, Saleh Ashkboos, Torsten Hoefler, and Dan Alistarh.
\newblock {GPTQ}: Accurate post-training quantization for generative
  pre-trained transformers.
\newblock \emph{arXiv preprint arXiv:2210.17323}, 2023.

\bibitem[Froger et~al.(2025)Froger, Andrews, Bettini, Budhiraja, Cabral, Do,
  Garreau, Gaya, Laurençon, Lecanu, Malkan, Mekala, Ménard, Bertran,
  Piterbarg, Plekhanov, Rita, Rusakov, Vorotilov, Wang, Yu, Benhalloum, Mialon,
  and Scialom]{froger2025arescalingagentenvironments}
Romain Froger, Pierre Andrews, Matteo Bettini, Amar Budhiraja, Ricardo~Silveira
  Cabral, Virginie Do, Emilien Garreau, Jean-Baptiste Gaya, Hugo Laurençon,
  Maxime Lecanu, Kunal Malkan, Dheeraj Mekala, Pierre Ménard, Gerard
  Moreno-Torres Bertran, Ulyana Piterbarg, Mikhail Plekhanov, Mathieu Rita,
  Andrey Rusakov, Vladislav Vorotilov, Mengjue Wang, Ian Yu, Amine Benhalloum,
  Grégoire Mialon, and Thomas Scialom.
\newblock Are: Scaling up agent environments and evaluations.
\newblock \emph{arXiv preprint arXiv:2509.17158}, 2025.
\newblock \url{https://arxiv.org/abs/2509.17158}.

\bibitem[Froger et~al.(2026)Froger, Andrews, Bettini, Budhiraja, Cabral, Do,
  Garreau, Gaya, Laurençon, Lecanu, Malkan, Mekala, Ménard, Bertran,
  Piterbarg, Plekhanov, Rita, Rusakov, Vorotilov, Wang, Yu, Benhalloum, Mialon,
  and Scialom]{froger2026gaia2benchmarkingllmagents}
Romain Froger, Pierre Andrews, Matteo Bettini, Amar Budhiraja, Ricardo~Silveira
  Cabral, Virginie Do, Emilien Garreau, Jean-Baptiste Gaya, Hugo Laurençon,
  Maxime Lecanu, Kunal Malkan, Dheeraj Mekala, Pierre Ménard, Gerard
  Moreno-Torres Bertran, Ulyana Piterbarg, Mikhail Plekhanov, Mathieu Rita,
  Andrey Rusakov, Vladislav Vorotilov, Mengjue Wang, Ian Yu, Amine Benhalloum,
  Grégoire Mialon, and Thomas Scialom.
\newblock Gaia2: Benchmarking llm agents on dynamic and asynchronous
  environments, 2026.
\newblock \url{https://arxiv.org/abs/2602.11964}.

\bibitem[{Gemma Team} et~al.(2026){Gemma Team}, Abd, Aggarwal, Algayres,
  Andreev, Bachem, Ballantyne, Brick, Cărbune, Casbon, Chaturvedi, Cotruta,
  Coucke, Culliton, Dadashi, Dixon, Elhawaty, Evci, Farabet, Ferret, Galgani,
  Girgin, Grill, Grootendorst, Guo, Hardin, He, Hernandez, Homburger, Hussenot,
  Ji, Joulin, Kamath, Kassraie, Lacombe, Lahoti, Liu, Martins, Martins,
  Matejovicova, Merhej, Momchev, Mondal, Mullins, Panyam, Pathak, Perrin,
  Pinto, Pot, Pouget, Ramé, Ramos, Reid, Rim, Rivière, Roth, Rouillard,
  Sanseviero, Sessa, Settle, Sinopalnikov, Smoot, Stanczyk, Steiner, Stewart,
  Tolstikhin, Tschannen, Tsitsulin, Vieillard, Wu, Xu, Yang, Yvinec, Zhang,
  Zou, Aagnes, Abdelhamed, Agrawal, Agrawal, Alabdulmohsin, Alayrac, Alon,
  Amarnath, Anand, Anastasiou, Ariafar, Aubet, Axiotis, Barbero, Barral,
  Bendebury, Bergmann, Bileschi, Black, Blondel, Borgeaud, Bražinskas,
  Burnell, Busa-Fekete, Cai, Cameron, Caucheteux, Chadha, Chan, Chawla, Chen,
  Chen, Chen, Chen, Cheng, hsiang Chien, Chinaev, Chou, Chu, Coleman, Consul,
  Conway-Rahman, Crowell, Cutler, Dani, Daruki, Das, Deutsch, Dikkala, Ding,
  Ding, Dodhia, Donhauser, Doshi, Dragan, Druinsky, Dua, Egyed, Eisenbud,
  Eppens, Fan, Fatemi, Fathullah, Feinberg, Ferev, Fujimoto, Galatzer-Levy,
  Gante, Geisler, Ghosal, Girgis, Go, Gokhale, Grills, Gu, Gupta, Guruganesh,
  Hadsell, Harkous, Harlalka, Hassabis, Hauth, Heyward, Hosseini, Hsia, Hsu,
  Huang, Huang, Hui, Hutter, I, Iliopoulos, Jain, Jawahar, Ji, Jin, Johnson,
  Joshi, Kandoor, Kang, Kavukcuoglu, Kazemi, Kenealy, Khalifa, Kirk, Kothawade,
  Kovalev, Kovelamudi, Kraft, Kumar, Kuppam, Lannin, Lee, Lee, Lepikhin, Li,
  Li, Liévin, Lin, Lin, Liu, Liu, Liu, Liu, Lunayach, Ma, Madan, Maksai,
  Malmi, Matuszak, McDuff, Menghani, Mirylenka, Misiunas, Misra, Mitran,
  Mohamed, Mukha, Noland, O'Donnell, Olszewska, Orlando, Pan, Panigrahy,
  Parekh, Park, Paskie, Peng, Petrini, Petrov, Pfeiffer, Piot, Plomecka, Poder,
  Ponce, Pramanik, Racz, Rajan, Ramanovich, Rao, Ritter, Rodrigues, Rosen,
  Rybiński, Sachdeva, Sander, Sathyanarayana, Savla, Schmidgall, Schuster,
  Seguin, Sellergren, Severyn, Shafran, Shah, Shangguan, Shenoy, Shenoy,
  Shivanna, Sho, Spangher, Stokowiec, Strother, Su, Sun, Sundararajan,
  Tacchetti, Taege, Tafti, Tekur, Thapa, Traverse, Treven, Tu, Tung,
  Veličković, Venkat, Venkatesh, Venkiteswaran, Visin, Vitvitskyi,
  Vodrahalli, Wang, Wang, Warkentin, Wassenberg, Wieting, Xiao, Xu, Xu, Xue,
  Yadav, Yan, Yang, Yang, Yang, Ying, Yoo, Zafar, Zhang, Zhang, Zhang, Zhang,
  Zhao, Zhou, and Zou]{gemmateam2026gemma4technicalreport}
{Gemma Team}, Sherif~El Abd, Vaibhav Aggarwal, Robin Algayres, Alek Andreev,
  Olivier Bachem, Ian Ballantyne, Cormac Brick, Victor Cărbune, Michelle
  Casbon, Mayank Chaturvedi, Victor Cotruta, Alice Coucke, Phil Culliton,
  Robert Dadashi, Lucas Dixon, Mohamed Elhawaty, Utku Evci, Clément Farabet,
  Johan Ferret, Filippo Galgani, Sertan Girgin, Jean-Bastien Grill, Maarten
  Grootendorst, Jiaxian Guo, Cassidy Hardin, Yanzhang He, Steven~M. Hernandez,
  Omri Homburger, Léonard Hussenot, Juyeong Ji, Armand Joulin, Aishwarya
  Kamath, Parnian Kassraie, Olivier Lacombe, Preethi Lahoti, Gaël Liu, Gus
  Martins, Luciano Martins, Tatiana Matejovicova, Ramona Merhej, Nikola
  Momchev, Sneha Mondal, Ryan Mullins, Sindhu~Raghuram Panyam, Shreya Pathak,
  Sarah Perrin, André~Susano Pinto, Etienne Pot, Angéline Pouget, Alexandre
  Ramé, Sabela Ramos, Douglas Reid, David Rim, Morgane Rivière, Karsten Roth,
  Louis Rouillard, Omar Sanseviero, Pier~Giuseppe Sessa, Shane Settle, Danila
  Sinopalnikov, Sara Smoot, Piotr Stanczyk, Andreas Steiner, Lawrence Stewart,
  Ilya Tolstikhin, Michael Tschannen, Anton Tsitsulin, Nino Vieillard, Renjie
  Wu, Pingmei Xu, Haichuan Yang, Edouard Yvinec, Li~Zhang, Joe Zou, Nicolas
  Aagnes, Abdelrahman Abdelhamed, Shivani Agrawal, Shubham Agrawal, Ibrahim
  Alabdulmohsin, Jean~Baptiste Alayrac, Uri Alon, Chandramouli Amarnath, Ankesh
  Anand, Chrysovalantis Anastasiou, Setareh Ariafar, François-Xavier Aubet,
  Kyriakos Axiotis, Federico Barbero, Joelle Barral, Alexei Bendebury, Urs
  Bergmann, Stanley Bileschi, Kat Black, Mathieu Blondel, Sebastian Borgeaud,
  Arthur Bražinskas, Ryan Burnell, Robert Busa-Fekete, Mu~Cai, Glenn Cameron,
  Charlotte Caucheteux, Garima Chadha, Jetha Chan, Aditya Chawla,
  Blake~Jianhang Chen, Jesse Chen, Lin Chen, Xu~Chen, Derek Cheng, Tzu hsiang
  Chien, Nikolai Chinaev, Yi~Chou, Zhaohui Chu, Benjamin Coleman, Pooja Consul,
  Sam Conway-Rahman, Scott Crowell, Dylan Cutler, Vivek Dani, Samira Daruki,
  Anil Das, Daniel Deutsch, Nishanth Dikkala, Li~Ding, Qiuhan Ding, Shenil
  Dodhia, Konstantin Donhauser, Tulsee Doshi, Anca Dragan, Alex Druinsky, Sahil
  Dua, Zoltan Egyed, Danielle Eisenbud, Daniel Eppens, Cindy Fan, Bahare
  Fatemi, Yassir Fathullah, Vlad Feinberg, Milen Ferev, Takumi Fujimoto, Isaac
  Galatzer-Levy, João Gante, Simon Geisler, Soham Ghosal, Antonious~M. Girgis,
  Alec Go, Alhaad Gokhale, Alex Grills, Yiming Gu, Pramod Gupta, Guru
  Guruganesh, Raia Hadsell, Hamza Harkous, Jitendra Harlalka, Demis Hassabis,
  Anja Hauth, Joe Heyward, Arian Hosseini, Chih-Yang Hsia, I-Hung Hsu, Xiaopeng
  Huang, Yangsibo Huang, Kevin Hui, Adrian Hutter, Te~I, Fotis Iliopoulos,
  Advait Jain, Ganesh Jawahar, Ziwei Ji, Qilin Jin, Melvin Johnson, Kandarp
  Joshi, Arun Kandoor, Wang-Cheng Kang, Koray Kavukcuoglu, Mehran Kazemi,
  Kathleen Kenealy, Amr Khalifa, Phoebe Kirk, Suraj Kothawade, Vitaly Kovalev,
  Neel Kovelamudi, Adam Kraft, Ravin Kumar, Harish Kuppam, Justin Lannin,
  Chen-Yu Lee, Seungji Lee, Dmitry Lepikhin, Dongdong Li, Qiujia Li, Valentin
  Liévin, Ethan Lin, Ziqian Lin, Casper Liu, Tianlin Liu, Tianqi Liu, Xin Liu,
  Mayank Lunayach, Min Ma, Gagan Madan, Andrii Maksai, Eric Malmi, Michal
  Matuszak, Daniel McDuff, Gaurav Menghani, Daniil Mirylenka, Karolis Misiunas,
  Vedant Misra, Andreea Mitran, Kareem Mohamed, Maksim Mukha, Eric Noland,
  James O'Donnell, Kate Olszewska, Bernett Orlando, Wanqiong Pan, Rina
  Panigrahy, Unnati Parekh, Chunjong Park, Eric Paskie, Liqian Peng, Bryce
  Petrini, Slav Petrov, Jonas Pfeiffer, Bilal Piot, Martyna Plomecka, Siim
  Poder, Octavio Ponce, Arijit Pramanik, David Racz, Anish Rajan, Michelle
  Ramanovich, Anand Rao, Marvin Ritter, Vitor Rodrigues, Evan Rosen, Mikołaj
  Rybiński, Noveen Sachdeva, Michaël~E. Sander, Rohit Sathyanarayana, Sagar
  Savla, Samuel Schmidgall, Tal Schuster, Benoit Seguin, Andrew Sellergren,
  Aliaksei Severyn, Izhak Shafran, Dhruv Shah, Yuan Shangguan, Ashish Shenoy,
  Pradeep Shenoy, Rakesh Shivanna, Pauline Sho, Lucas Spangher, Wojciech
  Stokowiec, Tim Strother, Yao Su, Yinghao Sun, Mukund Sundararajan, Andrea
  Tacchetti, Mor~Hazan Taege, Pouya Tafti, Chetan Tekur, Rahul Thapa, Madeleine
  Traverse, Lenart Treven, Tao Tu, Chien~Te Tung, Petar Veličković,
  Malini~Pooni Venkat, Sagar~Gubbi Venkatesh, Vidya Venkiteswaran, Francesco
  Visin, Alex Vitvitskyi, Kiran Vodrahalli, Weiyi Wang, Xin Wang, Tris
  Warkentin, Jan Wassenberg, John Wieting, Lechao Xiao, Hao Xu, Yuhui Xu,
  Fuzhao Xue, Arun Yadav, Jun Yan, Antoine Yang, Lin Yang, Ming-Hsuan Yang,
  Ziyu Ying, Jae~Hyeon Yoo, Sajjad Zafar, Fred Zhang, Jiageng Zhang, Jianyi
  Zhang, Xiaofan Zhang, Chao Zhao, David Zhou, and Chen Zou.
\newblock Gemma 4 technical report, 2026.
\newblock \url{https://arxiv.org/abs/2607.02770}.

\bibitem[Grattafiori et~al.(2024)Grattafiori, Dubey, Jauhri,
  et~al.]{grattafiori2024llama3}
Aaron Grattafiori, Abhimanyu Dubey, Abhinav Jauhri, et~al.
\newblock The llama 3 herd of models.
\newblock \emph{arXiv preprint arXiv:2407.21783}, 2024.

\bibitem[Groeneveld et~al.(2025)]{groeneveld2025olmo2}
Dirk Groeneveld et~al.
\newblock {OLMo} 2: Furious.
\newblock \emph{arXiv preprint arXiv:2501.00656}, 2025.

\bibitem[Gu et~al.(2024)Gu, Dong, Wei, and Huang]{gu2024minillm}
Yuxian Gu, Li~Dong, Furu Wei, and Minlie Huang.
\newblock {MiniLLM}: Knowledge distillation of large language models.
\newblock In \emph{International Conference on Learning Representations
  (ICLR)}, 2024.
\newblock \url{https://arxiv.org/abs/2306.08543}.

\bibitem[Guo et~al.(2025)Guo, Yang, Zhang, Song, Wang, Zhu, Xu, Zhang, Ma, Bi,
  Zhang, Yu, Wu, Wu, Gou, Shao, Li, Gao, Liu, Xue, Wang, Wu, Feng, Lu, Zhao,
  Deng, Ruan, Dai, Chen, Ji, Li, Lin, Dai, Luo, Hao, Chen, Li, Zhang, Xu, Ding,
  Gao, Qu, Li, Guo, Li, Chen, Yuan, Tu, Qiu, Li, Cai, Ni, Liang, Chen, Dong,
  Hu, You, Gao, Guan, Huang, Yu, Wang, Zhang, Zhao, Wang, Zhang, Xu, Xia,
  Zhang, Zhang, Tang, Zhou, Li, Wang, Li, Tian, Huang, Zhang, Wang, Chen, Du,
  Ge, Zhang, Pan, Wang, Chen, Jin, Chen, Lu, Zhou, Chen, Ye, Wang, Yu, Zhou,
  Pan, Li, Zhou, Wu, Yun, Pei, Sun, Wang, Zeng, Liu, Liang, Gao, Yu, Zhang,
  Xiao, An, Liu, Wang, Chen, Nie, Cheng, Liu, Xie, Liu, Yang, Li, Su, Lin, Li,
  Jin, Shen, Chen, Sun, Wang, Song, Zhou, Wang, Shan, Li, Wang, Wei, Zhang, Xu,
  Li, Zhao, Sun, Wang, Yu, Zhang, Shi, Xiong, He, Piao, Wang, Tan, Ma, Liu,
  Guo, Ou, Wang, Gong, Zou, He, Xiong, Luo, You, Liu, Zhou, Zhu, Huang, Li,
  Zheng, Zhu, Ma, Tang, Zha, Yan, Ren, Ren, Sha, Fu, Xu, Xie, Zhang, Hao, Ma,
  Yan, Wu, Gu, Zhu, Liu, Li, Xie, Song, Pan, Huang, Xu, Zhang, and
  Zhang]{Guo_2025}
Daya Guo, Dejian Yang, Haowei Zhang, Junxiao Song, Peiyi Wang, Qihao Zhu,
  Runxin Xu, Ruoyu Zhang, Shirong Ma, Xiao Bi, Xiaokang Zhang, Xingkai Yu,
  Yu~Wu, Z.~F. Wu, Zhibin Gou, Zhihong Shao, Zhuoshu Li, Ziyi Gao, Aixin Liu,
  Bing Xue, Bingxuan Wang, Bochao Wu, Bei Feng, Chengda Lu, Chenggang Zhao,
  Chengqi Deng, Chong Ruan, Damai Dai, Deli Chen, Dongjie Ji, Erhang Li,
  Fangyun Lin, Fucong Dai, Fuli Luo, Guangbo Hao, Guanting Chen, Guowei Li,
  H.~Zhang, Hanwei Xu, Honghui Ding, Huazuo Gao, Hui Qu, Hui Li, Jianzhong Guo,
  Jiashi Li, Jingchang Chen, Jingyang Yuan, Jinhao Tu, Junjie Qiu, Junlong Li,
  J.~L. Cai, Jiaqi Ni, Jian Liang, Jin Chen, Kai Dong, Kai Hu, Kaichao You,
  Kaige Gao, Kang Guan, Kexin Huang, Kuai Yu, Lean Wang, Lecong Zhang, Liang
  Zhao, Litong Wang, Liyue Zhang, Lei Xu, Leyi Xia, Mingchuan Zhang, Minghua
  Zhang, Minghui Tang, Mingxu Zhou, Meng Li, Miaojun Wang, Mingming Li, Ning
  Tian, Panpan Huang, Peng Zhang, Qiancheng Wang, Qinyu Chen, Qiushi Du, Ruiqi
  Ge, Ruisong Zhang, Ruizhe Pan, Runji Wang, R.~J. Chen, R.~L. Jin, Ruyi Chen,
  Shanghao Lu, Shangyan Zhou, Shanhuang Chen, Shengfeng Ye, Shiyu Wang,
  Shuiping Yu, Shunfeng Zhou, Shuting Pan, S.~S. Li, Shuang Zhou, Shaoqing Wu,
  Tao Yun, Tian Pei, Tianyu Sun, T.~Wang, Wangding Zeng, Wen Liu, Wenfeng
  Liang, Wenjun Gao, Wenqin Yu, Wentao Zhang, W.~L. Xiao, Wei An, Xiaodong Liu,
  Xiaohan Wang, Xiaokang Chen, Xiaotao Nie, Xin Cheng, Xin Liu, Xin Xie,
  Xingchao Liu, Xinyu Yang, Xinyuan Li, Xuecheng Su, Xuheng Lin, X.~Q. Li,
  Xiangyue Jin, Xiaojin Shen, Xiaosha Chen, Xiaowen Sun, Xiaoxiang Wang, Xinnan
  Song, Xinyi Zhou, Xianzu Wang, Xinxia Shan, Y.~K. Li, Y.~Q. Wang, Y.~X. Wei,
  Yang Zhang, Yanhong Xu, Yao Li, Yao Zhao, Yaofeng Sun, Yaohui Wang, Yi~Yu,
  Yichao Zhang, Yifan Shi, Yiliang Xiong, Ying He, Yishi Piao, Yisong Wang,
  Yixuan Tan, Yiyang Ma, Yiyuan Liu, Yongqiang Guo, Yuan Ou, Yuduan Wang, Yue
  Gong, Yuheng Zou, Yujia He, Yunfan Xiong, Yuxiang Luo, Yuxiang You, Yuxuan
  Liu, Yuyang Zhou, Y.~X. Zhu, Yanping Huang, Yaohui Li, Yi~Zheng, Yuchen Zhu,
  Yunxian Ma, Ying Tang, Yukun Zha, Yuting Yan, Z.~Z. Ren, Zehui Ren, Zhangli
  Sha, Zhe Fu, Zhean Xu, Zhenda Xie, Zhengyan Zhang, Zhewen Hao, Zhicheng Ma,
  Zhigang Yan, Zhiyu Wu, Zihui Gu, Zijia Zhu, Zijun Liu, Zilin Li, Ziwei Xie,
  Ziyang Song, Zizheng Pan, Zhen Huang, Zhipeng Xu, Zhongyu Zhang, and Zhen
  Zhang.
\newblock Deepseek-r1 incentivizes reasoning in llms through reinforcement
  learning.
\newblock \emph{Nature}, 645\penalty0 (8081):\penalty0 633–638, 2025.
\newblock ISSN 1476-4687.
\newblock \doi{10.1038/s41586-025-09422-z}.
\newblock \url{http://dx.doi.org/10.1038/s41586-025-09422-z}.

\bibitem[Gupta et~al.(2026)Gupta, Chatterjee, Haas, Tao, Wang, Liu, Oiwa,
  Gribovskaya, Ackermann, Blitzer, et~al.]{gupta2026deepsearchqa}
Nikita Gupta, Riju Chatterjee, Lukas Haas, Connie Tao, Andrew Wang, Chang Liu,
  Hidekazu Oiwa, Elena Gribovskaya, Jan Ackermann, John Blitzer, et~al.
\newblock Deepsearchqa: Bridging the comprehensiveness gap for deep research
  agents.
\newblock \emph{arXiv preprint arXiv:2601.20975}, 2026.

\bibitem[Hsieh et~al.(2024)Hsieh, Sun, Kriman, Acharya, Rekesh, Jia, Zhang, and
  Ginsburg]{hsieh2024ruler}
Cheng-Ping Hsieh, Simeng Sun, Samuel Kriman, Shantanu Acharya, Dima Rekesh, Fei
  Jia, Yang Zhang, and Boris Ginsburg.
\newblock Ruler: What's the real context size of your long-context language
  models?
\newblock \emph{arXiv preprint arXiv:2404.06654}, 2024.

\bibitem[Hu et~al.(2025)Hu, Lian, Xiao, Xiong, Lei, Wang, Ding, Xiao, Yuan, and
  Xie]{hu2025population}
Zhengyu Hu, Jianxun Lian, Zheyuan Xiao, Max Xiong, Yuxuan Lei, Tianfu Wang,
  Kaize Ding, Ziang Xiao, Nicholas~Jing Yuan, and Xing Xie.
\newblock Population-aligned persona generation for llm-based social
  simulation.
\newblock \emph{arXiv preprint arXiv:2509.10127}, 2025.

\bibitem[Izmailov et~al.(2018)Izmailov, Podoprikhin, Garipov, Vetrov, and
  Wilson]{izmailov2018averaging}
Pavel Izmailov, Dmitrii Podoprikhin, Timur Garipov, Dmitry Vetrov, and
  Andrew~Gordon Wilson.
\newblock Averaging weights leads to wider optima and better generalization.
\newblock In \emph{Uncertainty in Artificial Intelligence (UAI)}, 2018.

\bibitem[Jimenez et~al.(2024)Jimenez, Yang, Wettig, Yao, Pei, Press, and
  Narasimhan]{jimenez2024swebench}
Carlos~E Jimenez, John Yang, Alexander Wettig, Shunyu Yao, Kexin Pei, Ofir
  Press, and Karthik Narasimhan.
\newblock {SWE-bench}: Can language models resolve real-world {GitHub} issues?
\newblock \emph{arXiv preprint arXiv:2310.06770}, 2024.

\bibitem[Kang et~al.(2025)Kang, Galim, Oh, Kang, Park, Kim, Lee, Kim, Tiwari,
  Zeng, Koo, and Lee]{kang2025asyncopd}
Wonjun Kang, Kevin Galim, Seunghyuk Oh, Minjun Kang, Sanghyun Park, Donghoon
  Kim, Minjae Lee, Minseo Kim, Rishabh Tiwari, Yuchen Zeng, Hyung~Il Koo, and
  Kangwook Lee.
\newblock Asyncopd: How stale can on-policy distillation be?
\newblock \emph{arXiv preprint arXiv:2606.24143}, 2025.

\bibitem[{Kilo AI}(2026)]{pinchbench2026}
{Kilo AI}.
\newblock Pinchbench: Real-world benchmarks for ai coding agents, 2026.
\newblock \url{https://pinchbench.com}.
\newblock Accessed: 2026-07-27.

\bibitem[Kim and Rush(2016)]{kim2016sequence}
Yoon Kim and Alexander~M. Rush.
\newblock Sequence-level knowledge distillation.
\newblock In \emph{Proceedings of the 2016 Conference on Empirical Methods in
  Natural Language Processing}, pages 1317--1327, Austin, Texas, November 2016.
  Association for Computational Linguistics.
\newblock \doi{10.18653/v1/D16-1139}.
\newblock \url{https://aclanthology.org/D16-1139/}.

\bibitem[Li et~al.(2025{\natexlab{a}})Li, Meng, Lin, Luo, Tian, Ma, Huang, and
  Chua]{li2025screenspotpro}
Kaixin Li, Ziyang Meng, Hongzhan Lin, Ziyang Luo, Yuchen Tian, Jing Ma, Zhiyong
  Huang, and Tat-Seng Chua.
\newblock Screenspot-pro: Gui grounding for professional high-resolution
  computer use.
\newblock In \emph{Proceedings of the 33rd ACM International Conference on
  Multimedia}, pages 8778--8786, 2025{\natexlab{a}}.

\bibitem[Li et~al.(2026)Li, Liu, Chen, You, Di, He, Zheng, Choe, Sun, Wang,
  et~al.]{li2026skillsbench}
Xiangyi Li, Yimin Liu, Wenbo Chen, Bingran You, Zonglin Di, Yifeng He, Shenghan
  Zheng, Kyoung~Whan Choe, Jiankai Sun, Shuyi Wang, et~al.
\newblock Skillsbench: Benchmarking how well agent skills work across diverse
  tasks.
\newblock \emph{arXiv preprint arXiv:2602.12670}, 2026.

\bibitem[Li et~al.(2025{\natexlab{b}})]{li2025pma}
Yuelin Li et~al.
\newblock Model merging in pre-training of large language models.
\newblock \emph{arXiv preprint arXiv:2505.12082}, 2025{\natexlab{b}}.

\bibitem[Lin et~al.(2024)Lin, Tang, Tang, Yang, Chen, Wang, Xiao, Dang, Gan,
  and Han]{lin2024awq}
Ji~Lin, Jiaming Tang, Haotian Tang, Shang Yang, Wei-Ming Chen, Wei-Chen Wang,
  Guangxuan Xiao, Xingyu Dang, Chuang Gan, and Song Han.
\newblock {AWQ}: Activation-aware weight quantization for {LLM} compression and
  acceleration.
\newblock \emph{arXiv preprint arXiv:2306.00978}, 2024.

\bibitem[Liu et~al.(2024{\natexlab{a}})Liu, Feng, Wang, Wang, Liu, Zhao, Dengr,
  Ruan, Dai, Guo, et~al.]{liu2024deepseek}
Aixin Liu, Bei Feng, Bin Wang, Bingxuan Wang, Bo~Liu, Chenggang Zhao, Chengqi
  Dengr, Chong Ruan, Damai Dai, Daya Guo, et~al.
\newblock Deepseek-v2: A strong, economical, and efficient mixture-of-experts
  language model.
\newblock \emph{arXiv preprint arXiv:2405.04434}, 2024{\natexlab{a}}.

\bibitem[Liu et~al.(2024{\natexlab{b}})Liu, Li, Li, Li, Zhang, Shen, and
  Lee]{liu2024llavanext}
Haotian Liu, Chunyuan Li, Yuheng Li, Bo~Li, Yuanhan Zhang, Sheng Shen, and
  Yong~Jae Lee.
\newblock Llavanext: Improved reasoning, ocr, and world knowledge,
  2024{\natexlab{b}}.

\bibitem[Liu et~al.(2025)Liu, Sun, Zhang, Bai, Yu, Yu, Yuan, and
  Hou]{liu2025quantization}
Ruikang Liu, Yuxuan Sun, Manyi Zhang, Haoli Bai, Xianzhi Yu, Tiezheng Yu, Chun
  Yuan, and Lu~Hou.
\newblock Quantization hurts reasoning? an empirical study on quantized
  reasoning models.
\newblock \emph{arXiv preprint arXiv:2504.04823}, 2025.

\bibitem[Lotfi et~al.(2026)Lotfi, Kirichenko, Li, and Liu]{lotfi2026quantized}
Sanae Lotfi, Polina Kirichenko, Steven Li, and Zechun Liu.
\newblock Quantized reasoning models think they need to think longer, but they
  do not.
\newblock \emph{arXiv preprint arXiv:2606.00206}, 2026.

\bibitem[McCrae and John(1992)]{mccrate1992personality}
Robert~R. McCrae and Oliver~P. John.
\newblock An introduction to the five-factor model and its applications.
\newblock \emph{Journal of Personality}, 60\penalty0 (2):\penalty0 175--215,
  1992.
\newblock \doi{https://doi.org/10.1111/j.1467-6494.1992.tb00970.x}.
\newblock
  \url{https://onlinelibrary.wiley.com/doi/abs/10.1111/j.1467-6494.1992.tb00970.x}.

\bibitem[Menghini et~al.(2026)Menghini, Ney, Kwisaba, Wang, Turpin, Binder,
  Testud, Boyd, Li, Evtimov, et~al.]{menghini2026musesparksafety}
Cristina Menghini, Peter Ney, Hamza Kwisaba, Zifan Wang, Miles Turpin, Felix
  Binder, Jean-Christophe Testud, Aidan Boyd, Nathaniel Li, Ivan Evtimov,
  et~al.
\newblock Muse spark safety \& preparedness report.
\newblock \emph{arXiv preprint arXiv:2606.12429}, 2026.

\bibitem[Merrill et~al.(2026)Merrill, Shaw, Carlini, Li, Raj, Bercovich, Shi,
  Shin, Walshe, Buchanan, et~al.]{merrill2026terminal}
Mike~A Merrill, Alexander~G Shaw, Nicholas Carlini, Boxuan Li, Harsh Raj, Ivan
  Bercovich, Lin Shi, Jeong~Yeon Shin, Thomas Walshe, E~Kelly Buchanan, et~al.
\newblock Terminal-bench: Benchmarking agents on hard, realistic tasks in
  command line interfaces.
\newblock \emph{arXiv preprint arXiv:2601.11868}, 2026.

\bibitem[{Meta AI}(2025)]{meta2025llama4}
{Meta AI}.
\newblock The llama 4 herd: The beginning of a new era of natively multimodal
  ai innovation.
\newblock \url{https://ai.meta.com/blog/llama-4-multimodal-intelligence/},
  2025.
\newblock Accessed: 2025.

\bibitem[Mireshghallah et~al.(2026)Mireshghallah, Mangaokar, Kokhlikyan,
  Zharmagambetov, Zaheer, Mahloujifar, and
  Chaudhuri]{mireshghallah2026cimemories}
Niloofar Mireshghallah, Neal Mangaokar, Narine Kokhlikyan, Arman
  Zharmagambetov, Manzil Zaheer, Saeed Mahloujifar, and Kamalika Chaudhuri.
\newblock Cimemories: A compositional benchmark for contextual integrity in
  llms.
\newblock In \emph{International Conference on Learning Representations},
  volume 2026, pages 95513--95532, 2026.

\bibitem[Nissenbaum(2004)]{nissenbaum2004privacy}
Helen Nissenbaum.
\newblock Privacy as contextual integrity.
\newblock \emph{Washington Law Review}, 79\penalty0 (1):\penalty0 119--158,
  2004.

\bibitem[{NLLB Team} et~al.(2024){NLLB Team}, Costa-juss{\`a}, Cross,
  {\c{C}}elebi, Elbayad, Heafield, Heffernan, Kalbassi, Lam, Licht, Maillard,
  Sun, Wang, Wenzek, Youngblood, Akula, Barrault, Gonzalez, Hansanti, Hoffman,
  Jarrett, Sadagopan, Rowe, Spruit, Tran, Andrews, Ayan, Bhosale, Edunov, Fan,
  Gao, Goswami, Guzm{\'a}n, Koehn, Mourachko, Ropers, Saleem, Schwenk, and
  Wang]{nllb-24}
{NLLB Team}, Marta~R. Costa-juss{\`a}, James Cross, Onur {\c{C}}elebi, Maha
  Elbayad, Kenneth Heafield, Kevin Heffernan, Elahe Kalbassi, Janice Lam,
  Daniel Licht, Jean Maillard, Anna Sun, Skyler Wang, Guillaume Wenzek,
  Al~Youngblood, Bapi Akula, Loic Barrault, Gabriel~Mejia Gonzalez, Prangthip
  Hansanti, John Hoffman, Semarley Jarrett, Kaushik~Ram Sadagopan, Dirk Rowe,
  Shannon Spruit, Chau Tran, Pierre Andrews, Necip~Fazil Ayan, Shruti Bhosale,
  Sergey Edunov, Angela Fan, Cynthia Gao, Vedanuj Goswami, Francisco
  Guzm{\'a}n, Philipp Koehn, Alexandre Mourachko, Christophe Ropers, Safiyyah
  Saleem, Holger Schwenk, and Jeff Wang.
\newblock Scaling neural machine translation to 200 languages.
\newblock \emph{Nature}, 630\penalty0 (8018):\penalty0 841--846, 2024.
\newblock ISSN 1476-4687.
\newblock \doi{10.1038/s41586-024-07335-x}.
\newblock \url{https://doi.org/10.1038/s41586-024-07335-x}.

\bibitem[{NVIDIA Corporation}(2024--2026)]{nvidia-modelopt}
{NVIDIA Corporation}.
\newblock {NVIDIA Model Optimizer}.
\newblock \url{https://github.com/NVIDIA/Model-Optimizer}, 2024--2026.
\newblock GitHub repository.

\bibitem[OpenAI et~al.(2024)OpenAI, :, Hurst, Lerer, Goucher, Perelman, Ramesh,
  Clark, Ostrow, Welihinda, Hayes, Radford, Mądry, Baker-Whitcomb, Beutel,
  Borzunov, Carney, Chow, Kirillov, Nichol, Paino, Renzin, Passos, Kirillov,
  Christakis, Conneau, Kamali, Jabri, Moyer, Tam, Crookes, Tootoochian,
  Tootoonchian, Kumar, Vallone, Karpathy, Braunstein, Cann, Codispoti, Galu,
  Kondrich, Tulloch, Mishchenko, Baek, Jiang, Pelisse, Woodford, Gosalia, Dhar,
  Pantuliano, Nayak, Oliver, Zoph, Ghorbani, Leimberger, Rossen, Sokolowsky,
  Wang, Zweig, Hoover, Samic, McGrew, Spero, Giertler, Cheng, Lightcap, Walkin,
  Quinn, Guarraci, Hsu, Kellogg, Eastman, Lugaresi, Wainwright, Bassin, Hudson,
  Chu, Nelson, Li, Shern, Conger, Barette, Voss, Ding, Lu, Zhang, Beaumont,
  Hallacy, Koch, Gibson, Kim, Choi, McLeavey, Hesse, Fischer, Winter,
  Czarnecki, Jarvis, Wei, Koumouzelis, Sherburn, Kappler, Levin, Levy, Carr,
  Farhi, Mely, Robinson, Sasaki, Jin, Valladares, Tsipras, Li, Nguyen, Findlay,
  Oiwoh, Wong, Asdar, Proehl, Yang, Antonow, Kramer, Peterson, Sigler, Wallace,
  Brevdo, Mays, Khorasani, Such, Raso, Zhang, von Lohmann, Sulit, Goh, Oden,
  Salmon, Starace, Brockman, Salman, Bao, Hu, Wong, Wang, Schmidt, Whitney,
  Jun, Kirchner, de~Oliveira~Pinto, Ren, Chang, Chung, Kivlichan, O'Connell,
  O'Connell, Osband, Silber, Sohl, Okuyucu, Lan, Kostrikov, Sutskever,
  Kanitscheider, Gulrajani, Coxon, Menick, Pachocki, Aung, Betker, Crooks,
  Lennon, Kiros, Leike, Park, Kwon, Phang, Teplitz, Wei, Wolfe, Chen, Harris,
  Varavva, Lee, Shieh, Lin, Yu, Weng, Tang, Yu, Jang, Candela, Beutler,
  Landers, Parish, Heidecke, Schulman, Lachman, McKay, Uesato, Ward, Kim,
  Huizinga, Sitkin, Kraaijeveld, Gross, Kaplan, Snyder, Achiam, Jiao, Lee,
  Zhuang, Harriman, Fricke, Hayashi, Singhal, Shi, Karthik, Wood, Rimbach, Hsu,
  Nguyen, Gu-Lemberg, Button, Liu, Howe, Muthukumar, Luther, Ahmad, Kai, Itow,
  Workman, Pathak, Chen, Jing, Guy, Fedus, Zhou, Mamitsuka, Weng, McCallum,
  Held, Ouyang, Feuvrier, Zhang, Kondraciuk, Kaiser, Hewitt, Metz, Doshi,
  Aflak, Simens, Boyd, Thompson, Dukhan, Chen, Gray, Hudnall, Zhang, Aljubeh,
  Litwin, Zeng, Johnson, Shetty, Gupta, Shah, Yatbaz, Yang, Zhong, Glaese,
  Chen, Janner, Lampe, Petrov, Wu, Wang, Fradin, Pokrass, Castro, de~Castro,
  Pavlov, Brundage, Wang, Khan, Murati, Bavarian, Lin, Yesildal, Soto,
  Gimelshein, Cone, Staudacher, Summers, LaFontaine, Chowdhury, Ryder, Stathas,
  Turley, Tezak, Felix, Kudige, Keskar, Deutsch, Bundick, Puckett, Nachum,
  Okelola, Boiko, Murk, Jaffe, Watkins, Godement, Campbell-Moore, Chao,
  McMillan, Belov, Su, Bak, Bakkum, Deng, Dolan, Hoeschele, Welinder, Tillet,
  Pronin, Tillet, Dhariwal, Yuan, Dias, Lim, Arora, Troll, Lin, Lopes, Puri,
  Miyara, Leike, Gaubert, Zamani, Wang, Donnelly, Honsby, Smith, Sahai,
  Ramchandani, Huet, Carmichael, Zellers, Chen, Chen, Nigmatullin, Cheu, Jain,
  Altman, Schoenholz, Toizer, Miserendino, Agarwal, Culver, Ethersmith, Gray,
  Grove, Metzger, Hermani, Jain, Zhao, Wu, Jomoto, Wu, Shuaiqi, Xia, Phene,
  Papay, Narayanan, Coffey, Lee, Hall, Balaji, Broda, Stramer, Xu, Gogineni,
  Christianson, Sanders, Patwardhan, Cunninghman, Degry, Dimson, Raoux,
  Shadwell, Zheng, Underwood, Markov, Sherbakov, Rubin, Stasi, Kaftan, Heywood,
  Peterson, Walters, Eloundou, Qi, Moeller, Monaco, Kuo, Fomenko, Chang, Zheng,
  Zhou, Manassra, Sheu, Zaremba, Patil, Qian, Kim, Cheng, Zhang, He, Zhang,
  Jin, Dai, and Malkov]{openai2024gpt4ocard}
OpenAI, :, Aaron Hurst, Adam Lerer, Adam~P. Goucher, Adam Perelman, Aditya
  Ramesh, Aidan Clark, AJ~Ostrow, Akila Welihinda, Alan Hayes, Alec Radford,
  Aleksander Mądry, Alex Baker-Whitcomb, Alex Beutel, Alex Borzunov, Alex
  Carney, Alex Chow, Alex Kirillov, Alex Nichol, Alex Paino, Alex Renzin,
  Alex~Tachard Passos, Alexander Kirillov, Alexi Christakis, Alexis Conneau,
  Ali Kamali, Allan Jabri, Allison Moyer, Allison Tam, Amadou Crookes, Amin
  Tootoochian, Amin Tootoonchian, Ananya Kumar, Andrea Vallone, Andrej
  Karpathy, Andrew Braunstein, Andrew Cann, Andrew Codispoti, Andrew Galu,
  Andrew Kondrich, Andrew Tulloch, Andrey Mishchenko, Angela Baek, Angela
  Jiang, Antoine Pelisse, Antonia Woodford, Anuj Gosalia, Arka Dhar, Ashley
  Pantuliano, Avi Nayak, Avital Oliver, Barret Zoph, Behrooz Ghorbani, Ben
  Leimberger, Ben Rossen, Ben Sokolowsky, Ben Wang, Benjamin Zweig, Beth
  Hoover, Blake Samic, Bob McGrew, Bobby Spero, Bogo Giertler, Bowen Cheng,
  Brad Lightcap, Brandon Walkin, Brendan Quinn, Brian Guarraci, Brian Hsu,
  Bright Kellogg, Brydon Eastman, Camillo Lugaresi, Carroll Wainwright, Cary
  Bassin, Cary Hudson, Casey Chu, Chad Nelson, Chak Li, Chan~Jun Shern,
  Channing Conger, Charlotte Barette, Chelsea Voss, Chen Ding, Cheng Lu, Chong
  Zhang, Chris Beaumont, Chris Hallacy, Chris Koch, Christian Gibson, Christina
  Kim, Christine Choi, Christine McLeavey, Christopher Hesse, Claudia Fischer,
  Clemens Winter, Coley Czarnecki, Colin Jarvis, Colin Wei, Constantin
  Koumouzelis, Dane Sherburn, Daniel Kappler, Daniel Levin, Daniel Levy, David
  Carr, David Farhi, David Mely, David Robinson, David Sasaki, Denny Jin, Dev
  Valladares, Dimitris Tsipras, Doug Li, Duc~Phong Nguyen, Duncan Findlay,
  Edede Oiwoh, Edmund Wong, Ehsan Asdar, Elizabeth Proehl, Elizabeth Yang, Eric
  Antonow, Eric Kramer, Eric Peterson, Eric Sigler, Eric Wallace, Eugene
  Brevdo, Evan Mays, Farzad Khorasani, Felipe~Petroski Such, Filippo Raso,
  Francis Zhang, Fred von Lohmann, Freddie Sulit, Gabriel Goh, Gene Oden, Geoff
  Salmon, Giulio Starace, Greg Brockman, Hadi Salman, Haiming Bao, Haitang Hu,
  Hannah Wong, Haoyu Wang, Heather Schmidt, Heather Whitney, Heewoo Jun,
  Hendrik Kirchner, Henrique~Ponde de~Oliveira~Pinto, Hongyu Ren, Huiwen Chang,
  Hyung~Won Chung, Ian Kivlichan, Ian O'Connell, Ian O'Connell, Ian Osband, Ian
  Silber, Ian Sohl, Ibrahim Okuyucu, Ikai Lan, Ilya Kostrikov, Ilya Sutskever,
  Ingmar Kanitscheider, Ishaan Gulrajani, Jacob Coxon, Jacob Menick, Jakub
  Pachocki, James Aung, James Betker, James Crooks, James Lennon, Jamie Kiros,
  Jan Leike, Jane Park, Jason Kwon, Jason Phang, Jason Teplitz, Jason Wei,
  Jason Wolfe, Jay Chen, Jeff Harris, Jenia Varavva, Jessica~Gan Lee, Jessica
  Shieh, Ji~Lin, Jiahui Yu, Jiayi Weng, Jie Tang, Jieqi Yu, Joanne Jang,
  Joaquin~Quinonero Candela, Joe Beutler, Joe Landers, Joel Parish, Johannes
  Heidecke, John Schulman, Jonathan Lachman, Jonathan McKay, Jonathan Uesato,
  Jonathan Ward, Jong~Wook Kim, Joost Huizinga, Jordan Sitkin, Jos Kraaijeveld,
  Josh Gross, Josh Kaplan, Josh Snyder, Joshua Achiam, Joy Jiao, Joyce Lee,
  Juntang Zhuang, Justyn Harriman, Kai Fricke, Kai Hayashi, Karan Singhal, Katy
  Shi, Kavin Karthik, Kayla Wood, Kendra Rimbach, Kenny Hsu, Kenny Nguyen,
  Keren Gu-Lemberg, Kevin Button, Kevin Liu, Kiel Howe, Krithika Muthukumar,
  Kyle Luther, Lama Ahmad, Larry Kai, Lauren Itow, Lauren Workman, Leher
  Pathak, Leo Chen, Li~Jing, Lia Guy, Liam Fedus, Liang Zhou, Lien Mamitsuka,
  Lilian Weng, Lindsay McCallum, Lindsey Held, Long Ouyang, Louis Feuvrier,
  Lu~Zhang, Lukas Kondraciuk, Lukasz Kaiser, Luke Hewitt, Luke Metz, Lyric
  Doshi, Mada Aflak, Maddie Simens, Madelaine Boyd, Madeleine Thompson, Marat
  Dukhan, Mark Chen, Mark Gray, Mark Hudnall, Marvin Zhang, Marwan Aljubeh,
  Mateusz Litwin, Matthew Zeng, Max Johnson, Maya Shetty, Mayank Gupta, Meghan
  Shah, Mehmet Yatbaz, Meng~Jia Yang, Mengchao Zhong, Mia Glaese, Mianna Chen,
  Michael Janner, Michael Lampe, Michael Petrov, Michael Wu, Michele Wang,
  Michelle Fradin, Michelle Pokrass, Miguel Castro, Miguel Oom~Temudo
  de~Castro, Mikhail Pavlov, Miles Brundage, Miles Wang, Minal Khan, Mira
  Murati, Mo~Bavarian, Molly Lin, Murat Yesildal, Nacho Soto, Natalia
  Gimelshein, Natalie Cone, Natalie Staudacher, Natalie Summers, Natan
  LaFontaine, Neil Chowdhury, Nick Ryder, Nick Stathas, Nick Turley, Nik Tezak,
  Niko Felix, Nithanth Kudige, Nitish Keskar, Noah Deutsch, Noel Bundick, Nora
  Puckett, Ofir Nachum, Ola Okelola, Oleg Boiko, Oleg Murk, Oliver Jaffe,
  Olivia Watkins, Olivier Godement, Owen Campbell-Moore, Patrick Chao, Paul
  McMillan, Pavel Belov, Peng Su, Peter Bak, Peter Bakkum, Peter Deng, Peter
  Dolan, Peter Hoeschele, Peter Welinder, Phil Tillet, Philip Pronin, Philippe
  Tillet, Prafulla Dhariwal, Qiming Yuan, Rachel Dias, Rachel Lim, Rahul Arora,
  Rajan Troll, Randall Lin, Rapha~Gontijo Lopes, Raul Puri, Reah Miyara, Reimar
  Leike, Renaud Gaubert, Reza Zamani, Ricky Wang, Rob Donnelly, Rob Honsby,
  Rocky Smith, Rohan Sahai, Rohit Ramchandani, Romain Huet, Rory Carmichael,
  Rowan Zellers, Roy Chen, Ruby Chen, Ruslan Nigmatullin, Ryan Cheu, Saachi
  Jain, Sam Altman, Sam Schoenholz, Sam Toizer, Samuel Miserendino, Sandhini
  Agarwal, Sara Culver, Scott Ethersmith, Scott Gray, Sean Grove, Sean Metzger,
  Shamez Hermani, Shantanu Jain, Shengjia Zhao, Sherwin Wu, Shino Jomoto,
  Shirong Wu, Shuaiqi, Xia, Sonia Phene, Spencer Papay, Srinivas Narayanan,
  Steve Coffey, Steve Lee, Stewart Hall, Suchir Balaji, Tal Broda, Tal Stramer,
  Tao Xu, Tarun Gogineni, Taya Christianson, Ted Sanders, Tejal Patwardhan,
  Thomas Cunninghman, Thomas Degry, Thomas Dimson, Thomas Raoux, Thomas
  Shadwell, Tianhao Zheng, Todd Underwood, Todor Markov, Toki Sherbakov, Tom
  Rubin, Tom Stasi, Tomer Kaftan, Tristan Heywood, Troy Peterson, Tyce Walters,
  Tyna Eloundou, Valerie Qi, Veit Moeller, Vinnie Monaco, Vishal Kuo, Vlad
  Fomenko, Wayne Chang, Weiyi Zheng, Wenda Zhou, Wesam Manassra, Will Sheu,
  Wojciech Zaremba, Yash Patil, Yilei Qian, Yongjik Kim, Youlong Cheng,
  Yu~Zhang, Yuchen He, Yuchen Zhang, Yujia Jin, Yunxing Dai, and Yury Malkov.
\newblock Gpt-4o system card, 2024.
\newblock \url{https://arxiv.org/abs/2410.21276}.

\bibitem[OpenAI et~al.(2026)OpenAI, :, Jaech, Kalai, Lerer, Richardson,
  El-Kishky, Low, Helyar, Madry, Beutel, Carney, Iftimie, Karpenko, Passos,
  Neitz, Prokofiev, Wei, Tam, Bennett, Kumar, Saraiva, Vallone, Duberstein,
  Kondrich, Mishchenko, Applebaum, Jiang, Nair, Zoph, Ghorbani, Zhang, Rossen,
  Sokolowsky, Barak, McGrew, Minaiev, Hao, Baker, Houghton, McKinzie, Eastman,
  Lugaresi, Bassin, Hudson, Li, de~Bourcy, Voss, Shen, Zhang, Koch, Orsinger,
  Hesse, Fischer, Chan, Roberts, Kappler, Levy, Selsam, Dohan, Farhi, Mely,
  Robinson, Tsipras, Li, Oprica, Freeman, Zhang, Wong, Proehl, Cheung,
  Mitchell, Wallace, Ritter, Mays, Wang, Such, Raso, Leoni, Tsimpourlas, Song,
  von Lohmann, Sulit, Salmon, Parascandolo, Chabot, Zhao, Brockman, Leclerc,
  Salman, Bao, Sheng, Andrin, Bagherinezhad, Ren, Lightman, Chung, Kivlichan,
  O'Connell, Osband, Gilaberte, Akkaya, Kostrikov, Sutskever, Kofman, Pachocki,
  Lennon, Wei, Harb, Twore, Feng, Yu, Weng, Tang, Yu, Candela, Palermo, Parish,
  Heidecke, Hallman, Rizzo, Gordon, Uesato, Ward, Huizinga, Wang, Chen, Xiao,
  Singhal, Nguyen, Cobbe, Shi, Wood, Rimbach, Gu-Lemberg, Liu, Lu, Stone, Yu,
  Ahmad, Yang, Liu, Maksin, Ho, Fedus, Weng, Li, McCallum, Held, Kuhn,
  Kondraciuk, Kaiser, Metz, Boyd, Trebacz, Joglekar, Chen, Tintor, Meyer,
  Jones, Kaufer, Schwarzer, Shah, Yatbaz, Guan, Xu, Yan, Glaese, Chen, Lampe,
  Malek, Wang, Fradin, McClay, Pavlov, Wang, Wang, Murati, Bavarian,
  Rohaninejad, McAleese, Chowdhury, Chowdhury, Ryder, Tezak, Brown, Nachum,
  Boiko, Murk, Watkins, Chao, Ashbourne, Izmailov, Zhokhov, Dias, Arora, Lin,
  Lopes, Gaon, Miyara, Leike, Hwang, Garg, Brown, James, Shu, Cheu, Greene,
  Jain, Altman, Toizer, Toyer, Miserendino, Agarwal, Hernandez, Baker,
  McKinney, Yan, Zhao, Hu, Santurkar, Chaudhuri, Zhang, Fu, Papay, Lin, Balaji,
  Sanjeev, Sidor, Broda, Clark, Wang, Gordon, Sanders, Patwardhan, Sottiaux,
  Degry, Dimson, Zheng, Garipov, Stasi, Bansal, Creech, Peterson, Eloundou, Qi,
  Kosaraju, Monaco, Pong, Fomenko, Zheng, Zhou, Zhan, McCabe, Zaremba, Dubois,
  Lu, Chen, Cha, Bai, He, Zhang, Wang, Shao, and Li]{openai2026openaio1card}
OpenAI, :, Aaron Jaech, Adam Kalai, Adam Lerer, Adam Richardson, Ahmed
  El-Kishky, Aiden Low, Alec Helyar, Aleksander Madry, Alex Beutel, Alex
  Carney, Alex Iftimie, Alex Karpenko, Alex~Tachard Passos, Alexander Neitz,
  Alexander Prokofiev, Alexander Wei, Allison Tam, Ally Bennett, Ananya Kumar,
  Andre Saraiva, Andrea Vallone, Andrew Duberstein, Andrew Kondrich, Andrey
  Mishchenko, Andy Applebaum, Angela Jiang, Ashvin Nair, Barret Zoph, Behrooz
  Ghorbani, Bohan Zhang, Ben Rossen, Benjamin Sokolowsky, Boaz Barak, Bob
  McGrew, Borys Minaiev, Botao Hao, Bowen Baker, Brandon Houghton, Brandon
  McKinzie, Brydon Eastman, Camillo Lugaresi, Cary Bassin, Cary Hudson,
  Chak~Ming Li, Charles de~Bourcy, Chelsea Voss, Chen Shen, Chong Zhang, Chris
  Koch, Chris Orsinger, Christopher Hesse, Claudia Fischer, Clive Chan, Dan
  Roberts, Daniel Kappler, Daniel Levy, Daniel Selsam, David Dohan, David
  Farhi, David Mely, David Robinson, Dimitris Tsipras, Doug Li, Dragos Oprica,
  Eben Freeman, Eddie Zhang, Edmund Wong, Elizabeth Proehl, Enoch Cheung, Eric
  Mitchell, Eric Wallace, Erik Ritter, Evan Mays, Fan Wang, Felipe~Petroski
  Such, Filippo Raso, Florencia Leoni, Foivos Tsimpourlas, Francis Song, Fred
  von Lohmann, Freddie Sulit, Geoff Salmon, Giambattista Parascandolo, Gildas
  Chabot, Grace Zhao, Greg Brockman, Guillaume Leclerc, Hadi Salman, Haiming
  Bao, Hao Sheng, Hart Andrin, Hessam Bagherinezhad, Hongyu Ren, Hunter
  Lightman, Hyung~Won Chung, Ian Kivlichan, Ian O'Connell, Ian Osband,
  Ignasi~Clavera Gilaberte, Ilge Akkaya, Ilya Kostrikov, Ilya Sutskever, Irina
  Kofman, Jakub Pachocki, James Lennon, Jason Wei, Jean Harb, Jerry Twore,
  Jiacheng Feng, Jiahui Yu, Jiayi Weng, Jie Tang, Jieqi Yu, Joaquin~Quiñonero
  Candela, Joe Palermo, Joel Parish, Johannes Heidecke, John Hallman, John
  Rizzo, Jonathan Gordon, Jonathan Uesato, Jonathan Ward, Joost Huizinga, Julie
  Wang, Kai Chen, Kai Xiao, Karan Singhal, Karina Nguyen, Karl Cobbe, Katy Shi,
  Kayla Wood, Kendra Rimbach, Keren Gu-Lemberg, Kevin Liu, Kevin Lu, Kevin
  Stone, Kevin Yu, Lama Ahmad, Lauren Yang, Leo Liu, Leon Maksin, Leyton Ho,
  Liam Fedus, Lilian Weng, Linden Li, Lindsay McCallum, Lindsey Held, Lorenz
  Kuhn, Lukas Kondraciuk, Lukasz Kaiser, Luke Metz, Madelaine Boyd, Maja
  Trebacz, Manas Joglekar, Mark Chen, Marko Tintor, Mason Meyer, Matt Jones,
  Matt Kaufer, Max Schwarzer, Meghan Shah, Mehmet Yatbaz, Melody~Y. Guan,
  Mengyuan Xu, Mengyuan Yan, Mia Glaese, Mianna Chen, Michael Lampe, Michael
  Malek, Michele Wang, Michelle Fradin, Mike McClay, Mikhail Pavlov, Miles
  Wang, Mingxuan Wang, Mira Murati, Mo~Bavarian, Mostafa Rohaninejad, Nat
  McAleese, Neil Chowdhury, Neil Chowdhury, Nick Ryder, Nikolas Tezak, Noam
  Brown, Ofir Nachum, Oleg Boiko, Oleg Murk, Olivia Watkins, Patrick Chao, Paul
  Ashbourne, Pavel Izmailov, Peter Zhokhov, Rachel Dias, Rahul Arora, Randall
  Lin, Rapha~Gontijo Lopes, Raz Gaon, Reah Miyara, Reimar Leike, Renny Hwang,
  Rhythm Garg, Robin Brown, Roshan James, Rui Shu, Ryan Cheu, Ryan Greene,
  Saachi Jain, Sam Altman, Sam Toizer, Sam Toyer, Samuel Miserendino, Sandhini
  Agarwal, Santiago Hernandez, Sasha Baker, Scott McKinney, Scottie Yan,
  Shengjia Zhao, Shengli Hu, Shibani Santurkar, Shraman~Ray Chaudhuri, Shuyuan
  Zhang, Siyuan Fu, Spencer Papay, Steph Lin, Suchir Balaji, Suvansh Sanjeev,
  Szymon Sidor, Tal Broda, Aidan Clark, Tao Wang, Taylor Gordon, Ted Sanders,
  Tejal Patwardhan, Thibault Sottiaux, Thomas Degry, Thomas Dimson, Tianhao
  Zheng, Timur Garipov, Tom Stasi, Trapit Bansal, Trevor Creech, Troy Peterson,
  Tyna Eloundou, Valerie Qi, Vineet Kosaraju, Vinnie Monaco, Vitchyr Pong, Vlad
  Fomenko, Weiyi Zheng, Wenda Zhou, Wenting Zhan, Wes McCabe, Wojciech Zaremba,
  Yann Dubois, Yinghai Lu, Yining Chen, Young Cha, Yu~Bai, Yuchen He, Yuchen
  Zhang, Yunyun Wang, Zheng Shao, and Zhuohan Li.
\newblock Openai o1 system card, 2026.
\newblock \url{https://arxiv.org/abs/2412.16720}.

\bibitem[{OpenClaw Contributors}(2025)]{openclaw2025}
{OpenClaw Contributors}.
\newblock Openclaw: An open-source framework for local ai agents.
\newblock \url{https://github.com/openclaw/openclaw}, 2025.
\newblock Accessed: 2026-07-21.

\bibitem[Ouyang et~al.(2024)Ouyang, Qu, Zhou, Zhu, Zhang, Lin, Wang, Zhao,
  Jiang, Zhao, Shi, Wu, Chu, Liu, Li, Xu, Zhang, Shi, Tu, and
  He]{ouyang2024omnidocbenchbenchmarkingdiversepdf}
Linke Ouyang, Yuan Qu, Hongbin Zhou, Jiawei Zhu, Rui Zhang, Qunshu Lin, Bin
  Wang, Zhiyuan Zhao, Man Jiang, Xiaomeng Zhao, Jin Shi, Fan Wu, Pei Chu,
  Minghao Liu, Zhenxiang Li, Chao Xu, Bo~Zhang, Botian Shi, Zhongying Tu, and
  Conghui He.
\newblock Omnidocbench: Benchmarking diverse pdf document parsing with
  comprehensive annotations, 2024.
\newblock \url{https://arxiv.org/abs/2412.07626}.

\bibitem[Patil et~al.(2023)Patil, Zhang, Wang, and Gonzalez]{patil2023gorilla}
Shishir~G. Patil, Tianjun Zhang, Xin Wang, and Joseph~E. Gonzalez.
\newblock Gorilla: Large language model connected with massive apis.
\newblock \emph{arXiv preprint arXiv:2305.15334}, 2023.

\bibitem[Patwardhan et~al.(2025)Patwardhan, Dias, Proehl, Kim, Wang, Watkins,
  Fishman, Aljubeh, Thacker, Fauconnet, Kim, Chao, Miserendino, Chabot, Li,
  Sharman, Barr, Glaese, and Tworek]{patwardhan2025gdpvalevaluatingaimodel}
Tejal Patwardhan, Rachel Dias, Elizabeth Proehl, Grace Kim, Michele Wang,
  Olivia Watkins, Simón~Posada Fishman, Marwan Aljubeh, Phoebe Thacker,
  Laurance Fauconnet, Natalie~S. Kim, Patrick Chao, Samuel Miserendino, Gildas
  Chabot, David Li, Michael Sharman, Alexandra Barr, Amelia Glaese, and Jerry
  Tworek.
\newblock Gdpval: Evaluating ai model performance on real-world economically
  valuable tasks, 2025.
\newblock \url{https://arxiv.org/abs/2510.04374}.

\bibitem[Popovi{\'c}(2015)]{popovic-2015-chrf}
Maja Popovi{\'c}.
\newblock chr{F}: character n-gram {F}-score for automatic {MT} evaluation.
\newblock In Ond{\v{r}}ej Bojar, Rajan Chatterjee, Christian Federmann, Barry
  Haddow, Chris Hokamp, Matthias Huck, Varvara Logacheva, and Pavel Pecina,
  editors, \emph{Proceedings of the Tenth Workshop on Statistical Machine
  Translation}, pages 392--395, Lisbon, Portugal, September 2015. Association
  for Computational Linguistics.
\newblock \doi{10.18653/v1/W15-3049}.
\newblock \url{https://aclanthology.org/W15-3049/}.

\bibitem[Pyatkin et~al.(2025)Pyatkin, Malik, Graf, Ivison, Huang, Dasigi,
  Lambert, and Hajishirzi]{pyatkin2025ifbench}
Valentina Pyatkin, Saumya Malik, Victoria Graf, Hamish Ivison, Shengyi Huang,
  Pradeep Dasigi, Nathan Lambert, and Hannaneh Hajishirzi.
\newblock Generalizing verifiable instruction following, 2025.

\bibitem[{Qwen Team}(2025)]{qwen2025qwen3}
{Qwen Team}.
\newblock Qwen3 technical report.
\newblock \emph{arXiv preprint arXiv:2505.09388}, 2025.

\bibitem[Rein et~al.(2024)Rein, Hou, Stickland, Petty, Pang, Dirani, Michael,
  and Bowman]{rein2024gpqa}
David Rein, Betty~Li Hou, Asa~Cooper Stickland, Jackson Petty, Richard~Yuanzhe
  Pang, Julien Dirani, Julian Michael, and Samuel~R. Bowman.
\newblock {GPQA}: A graduate-level {Google-Proof} q\&a benchmark.
\newblock \emph{arXiv preprint arXiv:2311.12022}, 2024.

\bibitem[Rouhani et~al.(2023)Rouhani, Zhao, More, Hall, Khodamoradi, Deng,
  Choudhary, Cornea, Dellinger, Denolf, et~al.]{rouhani2023microscaling}
Bita~Darvish Rouhani, Ritchie Zhao, Ankit More, Mathew Hall, Alireza
  Khodamoradi, Summer Deng, Dhruv Choudhary, Marius Cornea, Eric Dellinger,
  Kristof Denolf, et~al.
\newblock Microscaling data formats for deep learning.
\newblock \emph{arXiv preprint arXiv:2310.10537}, 2023.

\bibitem[S{\'a}nchez et~al.(2026)S{\'a}nchez, Alastruey, Ropers, Turkatenko,
  Saito~Stenetorp, Artetxe, and Ruiz Costa-juss{\`a}]{sanchez2026linguini}
Eduardo S{\'a}nchez, Belen Alastruey, Christophe Ropers, Arina Turkatenko,
  Pontus Lars~Erik Saito~Stenetorp, Mikel Artetxe, and Marta Ruiz
  Costa-juss{\`a}.
\newblock Linguini: A benchmark for language-agnostic linguistic reasoning.
\newblock \emph{Advances in Neural Information Processing Systems}, 38, 2026.

\bibitem[Sanyal et~al.(2023)Sanyal, Neerkaje, Kaddour, Kumar, and
  Sanghavi]{sanyal2023early}
Sunny Sanyal, Atula Neerkaje, Jean Kaddour, Abhishek Kumar, and Sujay Sanghavi.
\newblock Early weight averaging meets high learning rates for {LLM}
  pre-training.
\newblock \emph{arXiv preprint arXiv:2306.03241}, 2023.

\bibitem[Schick et~al.(2024)Schick, Dwivedi-Yu, Dess{\`i}, Raileanu, Lomeli,
  Hambro, Zettlemoyer, Cancedda, and Scialom]{schick2023toolformer}
Timo Schick, Jane Dwivedi-Yu, Roberto Dess{\`i}, Roberta Raileanu, Maria
  Lomeli, Eric Hambro, Luke Zettlemoyer, Nicola Cancedda, and Thomas Scialom.
\newblock Toolformer: Language models can teach themselves to use tools.
\newblock In \emph{Advances in Neural Information Processing Systems},
  volume~36, 2024.

\bibitem[Shao et~al.(2024)Shao, Li, Shi, Liu, and Yang]{shao2024privacylens}
Yijia Shao, Tianshi Li, Weiyan Shi, Yanchen Liu, and Diyi Yang.
\newblock Privacylens: Evaluating privacy norm awareness of language models in
  action.
\newblock \emph{Advances in Neural Information Processing Systems},
  37:\penalty0 89373--89407, 2024.

\bibitem[Shazeer(2020)]{shazeer2020glu}
Noam Shazeer.
\newblock Glu variants improve transformer.
\newblock \emph{arXiv preprint arXiv:2002.05202}, 2020.

\bibitem[Shi et~al.(2026)Shi, Zytek, Razavi, Narasimhan, and
  Barres]{shi2026tauknowledgeevaluatingconversationalagents}
Quan Shi, Alexandra Zytek, Pedram Razavi, Karthik Narasimhan, and Victor
  Barres.
\newblock $\tau$-knowledge: Evaluating conversational agents over unstructured
  knowledge, 2026.
\newblock \url{https://arxiv.org/abs/2603.04370}.

\bibitem[Sieberling et~al.(2025)Sieberling, Kuznedelev, and
  Alistarh]{sieberling2025evopress}
Oliver Sieberling, Denis Kuznedelev, and Dan Alistarh.
\newblock Evopress: Accurate dynamic model compression via evolutionary search.
\newblock In \emph{Sparsity in LLMs (SLLM): Deep Dive into Mixture of Experts,
  Quantization, Hardware, and Inference}, 2025.

\bibitem[Snell et~al.(2022)Snell, Klein, and Zhong]{snell2022learning}
Charlie Snell, Dan Klein, and Ruiqi Zhong.
\newblock Learning by distilling context.
\newblock \emph{arXiv preprint arXiv:2209.15189}, 2022.

\bibitem[Tavakoli et~al.(2025)Tavakoli, Salemi, Ye, Abdalla, Zamani, and
  Mitchell]{tavakoli2025milliontokensbenchmarkingenhancing}
Mohammad Tavakoli, Alireza Salemi, Carrie Ye, Mohamed Abdalla, Hamed Zamani,
  and J~Ross Mitchell.
\newblock Beyond a million tokens: Benchmarking and enhancing long-term memory
  in llms, 2025.
\newblock \url{https://arxiv.org/abs/2510.27246}.

\bibitem[Team(2025)]{artificialanalysis2025lcr}
Artificial~Analysis Team.
\newblock Artificial analysis long context reasoning benchmark(lcr), 2025.

\bibitem[Tian et~al.(2024)Tian, Gao, Zhang, Chen, Fan, Guo, Haas, Ji,
  Krongchon, Li, Liu, Luo, Ma, Tong, Trinh, Tian, Wang, Wu, Xiong, Yin, Zhu,
  Lieret, Lu, Liu, Du, Tao, Press, Callan, Huerta, and
  Peng]{tian2024scicoderesearchcodingbenchmark}
Minyang Tian, Luyu Gao, Shizhuo~Dylan Zhang, Xinan Chen, Cunwei Fan, Xuefei
  Guo, Roland Haas, Pan Ji, Kittithat Krongchon, Yao Li, Shengyan Liu, Di~Luo,
  Yutao Ma, Hao Tong, Kha Trinh, Chenyu Tian, Zihan Wang, Bohao Wu, Yanyu
  Xiong, Shengzhu Yin, Minhui Zhu, Kilian Lieret, Yanxin Lu, Genglin Liu,
  Yufeng Du, Tianhua Tao, Ofir Press, Jamie Callan, Eliu Huerta, and Hao Peng.
\newblock Scicode: A research coding benchmark curated by scientists, 2024.
\newblock \url{https://arxiv.org/abs/2407.13168}.

\bibitem[Touvron et~al.(2023{\natexlab{a}})Touvron, Lavril, Izacard, Martinet,
  Lachaux, Lacroix, Rozi{\`e}re, Goyal, Hambro, Azhar,
  et~al.]{touvron2023llama}
Hugo Touvron, Thibaut Lavril, Gautier Izacard, Xavier Martinet, Marie-Anne
  Lachaux, Timoth{\'e}e Lacroix, Baptiste Rozi{\`e}re, Naman Goyal, Eric
  Hambro, Faisal Azhar, et~al.
\newblock {LLaMA}: Open and efficient foundation language models.
\newblock \emph{arXiv preprint arXiv:2302.13971}, 2023{\natexlab{a}}.

\bibitem[Touvron et~al.(2023{\natexlab{b}})Touvron, Martin, Stone, Albert,
  Almahairi, Babaei, Bashlykov, Batra, Bhargava, Bhosale,
  et~al.]{touvron2023llama2}
Hugo Touvron, Louis Martin, Kevin Stone, Peter Albert, Amjad Almahairi, Yasmine
  Babaei, Nikolay Bashlykov, Soumya Batra, Prajjwal Bhargava, Shruti Bhosale,
  et~al.
\newblock Llama 2: Open foundation and fine-tuned chat models.
\newblock \emph{arXiv preprint arXiv:2307.09288}, 2023{\natexlab{b}}.

\bibitem[Wang et~al.(2024{\natexlab{a}})Wang, Peng, Ding, Hartford, Hajishirzi,
  et~al.]{wang2024openhands}
Xingyao Wang, Boxuan Peng, Yufan Ding, Jason Hartford, Hannaneh Hajishirzi,
  et~al.
\newblock Openhands: An open platform for ai software developers as generalist
  agents.
\newblock \emph{arXiv preprint arXiv:2407.16741}, 2024{\natexlab{a}}.

\bibitem[Wang et~al.(2024{\natexlab{b}})Wang, Xia, He, Chen, Liu, Zhu, Liang,
  Wu, Liu, Malladi, et~al.]{wang2024charxiv}
Zirui Wang, Mengzhou Xia, Luxi He, Howard Chen, Yitao Liu, Richard Zhu, Kaiqu
  Liang, Xindi Wu, Haotian Liu, Sadhika Malladi, et~al.
\newblock Charxiv: Charting gaps in realistic chart understanding in multimodal
  llms.
\newblock \emph{Advances in Neural Information Processing Systems},
  37:\penalty0 113569--113697, 2024{\natexlab{b}}.

\bibitem[Wortsman et~al.(2022)Wortsman, Ilharco, Gadre, Roelofs, Gontijo~Lopes,
  Morcos, Namkoong, Farhadi, Carmon, Kornblith, and Schmidt]{wortsman2022model}
Mitchell Wortsman, Gabriel Ilharco, Samir~Yitzhak Gadre, Rebecca Roelofs,
  Raphael Gontijo~Lopes, Ari~S Morcos, Hongseok Namkoong, Ali Farhadi, Yair
  Carmon, Simon Kornblith, and Ludwig Schmidt.
\newblock Model soups: Averaging weights of multiple fine-tuned models improves
  accuracy without increasing inference time.
\newblock In \emph{International Conference on Machine Learning (ICML)}, 2022.

\bibitem[Wu et~al.(2026)Wu, Yao, Qi, Zheng, Wang, Ma, Liao, Lakkaraju, Li, and
  Du]{wu2026dpace}
Tianyu Wu, Yu~Yao, Zhenting Qi, Han Zheng, Zhuohan Wang, Haoran Ma, Lawrence
  Liao, Himabindu Lakkaraju, Ju~Li, and Yilun Du.
\newblock D-pace: Dynamic position-aware cross-entropy for parallel speculative
  drafting, 2026.
\newblock \url{https://arxiv.org/abs/2605.18810}.

\bibitem[Wu et~al.(2024)Wu, Wu, Xu, Wang, Sun, Jia, Cheng, Ding, Chen, Liang,
  et~al.]{wu2024osatlas}
Zhiyong Wu, Zhenyu Wu, Fangzhi Xu, Yian Wang, Qiushi Sun, Chengyou Jia, Kanzhi
  Cheng, Zichen Ding, Liheng Chen, Paul~Pu Liang, et~al.
\newblock Os-atlas: A foundation action model for generalist gui agents.
\newblock \emph{arXiv preprint arXiv:2410.23218}, 2024.

\bibitem[Xie et~al.(2024)Xie, Zhang, Chen, Li, Zhao, Cao, Hua, Cheng, Shin,
  Lei, Liu, Deng, Jain, Bisk, et~al.]{OSWorld}
Tianbao Xie, Danyang Zhang, Jixuan Chen, Xiaochuan Li, Siheng Zhao, Ruisheng
  Cao, Toh~Jing Hua, Zhoujun Cheng, Dongchan Shin, Fangyu Lei, Yitao Liu,
  Yiheng Deng, Nreal Jain, Yonatan Bisk, et~al.
\newblock {OSWorld}: Benchmarking multimodal agents for open-ended tasks in
  real computer environments.
\newblock In \emph{Advances in Neural Information Processing Systems}, 2024.

\bibitem[Xuan et~al.(2025)Xuan, Yang, Qi, Zeng, Xiao, Feng, Liu, Xing, Wang,
  Gao, Lu, Jiang, Li, Li, Yu, Dong, Gu, Li, Xie, Juefei-Xu, Khomh, Yoshie,
  Chen, Teodoro, Liu, Goebel, Ma, Marrese-Taylor, Lu, Iwasawa, Matsuo, and
  Li]{xuan-etal-2025-mmlu}
Weihao Xuan, Rui Yang, Heli Qi, Qingcheng Zeng, Yunze Xiao, Aosong Feng, Dairui
  Liu, Yun Xing, Junjue Wang, Fan Gao, Jinghui Lu, Yuang Jiang, Huitao Li, Xin
  Li, Kunyu Yu, Ruihai Dong, Shangding Gu, Yuekang Li, Xiaofei Xie, Felix
  Juefei-Xu, Foutse Khomh, Osamu Yoshie, Qingyu Chen, Douglas Teodoro, Nan Liu,
  Randy Goebel, Lei Ma, Edison Marrese-Taylor, Shijian Lu, Yusuke Iwasawa,
  Yutaka Matsuo, and Irene Li.
\newblock {MMLU}-{P}ro{X}: A multilingual benchmark for advanced large language
  model evaluation.
\newblock In Christos Christodoulopoulos, Tanmoy Chakraborty, Carolyn Rose, and
  Violet Peng, editors, \emph{Proceedings of the 2025 Conference on Empirical
  Methods in Natural Language Processing}, pages 1513--1532, Suzhou, China,
  November 2025. Association for Computational Linguistics.
\newblock ISBN 979-8-89176-332-6.
\newblock \doi{10.18653/v1/2025.emnlp-main.79}.
\newblock \url{https://aclanthology.org/2025.emnlp-main.79/}.

\bibitem[Yang et~al.(2024)Yang, Jimenez, Wettig, Liber, Narasimhan, and
  Press]{yang2024sweagent}
John Yang, Carlos~E. Jimenez, Alexander Wettig, Kilian Liber, Karthik
  Narasimhan, and Ofir Press.
\newblock Swe-agent: Agent-computer interfaces enable automated software
  engineering.
\newblock In \emph{Advances in Neural Information Processing Systems},
  volume~37, 2024.

\bibitem[Yao et~al.(2023)Yao, Zhao, Yu, Du, Shafran, Narasimhan, and
  Cao]{yao2023react}
Shunyu Yao, Jeffrey Zhao, Dian Yu, Nan Du, Izhak Shafran, Karthik Narasimhan,
  and Yuan Cao.
\newblock React: Synergizing reasoning and acting in language models.
\newblock \emph{arXiv preprint arXiv:2210.03629}, 2023.

\bibitem[Yao et~al.(2024)Yao, Shinn, Razavi, and
  Narasimhan]{yao2024taubenchbenchmarktoolagentuserinteraction}
Shunyu Yao, Noah Shinn, Pedram Razavi, and Karthik Narasimhan.
\newblock $\tau$-bench: A benchmark for tool-agent-user interaction in
  real-world domains, 2024.
\newblock \url{https://arxiv.org/abs/2406.12045}.

\bibitem[Yue et~al.(2025)Yue, Zheng, Ni, Wang, Zhang, Tong, Sun, Yu, Zhang,
  Sun, et~al.]{yue2025mmmu}
Xiang Yue, Tianyu Zheng, Yuansheng Ni, Yubo Wang, Kai Zhang, Shengbang Tong,
  Yuxuan Sun, Botao Yu, Ge~Zhang, Huan Sun, et~al.
\newblock Mmmu-pro: A more robust multi-discipline multimodal understanding
  benchmark.
\newblock In \emph{Proceedings of the 63rd Annual Meeting of the Association
  for Computational Linguistics (Volume 1: Long Papers)}, pages 15134--15186,
  2025.

\bibitem[Zhao et~al.(2025)]{zhao2025selfdistilledreasoner}
Jiawei Zhao et~al.
\newblock Self-distilled reasoner: On-policy self-distillation for large
  language models.
\newblock \emph{arXiv preprint arXiv:2601.18734}, 2025.

\bibitem[Zheng et~al.(2025)Zheng, Cheng, Shen, Zhou, Liu, He, Li, Wei, Hao,
  Yao, Sheng, Wang, Chai, Korolova, Henderson, Arora, Viswanath, Shang, and
  Xie]{zheng2025livecodebenchproolympiadmedalists}
Zihan Zheng, Zerui Cheng, Zeyu Shen, Shang Zhou, Kaiyuan Liu, Hansen He,
  Dongruixuan Li, Stanley Wei, Hangyi Hao, Jianzhu Yao, Peiyao Sheng, Zixuan
  Wang, Wenhao Chai, Aleksandra Korolova, Peter Henderson, Sanjeev Arora,
  Pramod Viswanath, Jingbo Shang, and Saining Xie.
\newblock Livecodebench pro: How do olympiad medalists judge llms in
  competitive programming?, 2025.
\newblock \url{https://arxiv.org/abs/2506.11928}.

\end{thebibliography}

\newpage
\beginappendix

\section{Benchmarks and Evaluation}
\label{sec:benchmarks}

%\subsection{Full Benchmark results}

\begin{table}[!htbp]%[!t]
\tiny
\centering
\footnotesize
\setlength{\tabcolsep}{5pt}
\renewcommand{\arraystretch}{1.15}
\caption{
\textbf{\glimmermodelname capability scorecard.} We compare \glimmermodelname %\textcolor{red}{(RC2-10)}
at \emph{low}, \emph{mid}, \emph{high}, and \emph{xhigh} reasoning strength. Best capability per row in \textbf{bold}. Higher is better unless marked $\downarrow$. All benchmarks were screened for contamination; ``--'' denotes not evaluated.
}
\label{tab:main-results-full}
\begin{tabularx}{\textwidth}{>{\raggedright\arraybackslash}p{3.5cm}l*{4}{>{\centering\arraybackslash}X}}
\toprule
\rowcolor{blue!18}
\multicolumn{6}{c}{\textbf{Capabilities Scorecard}}
\\
\midrule
\rowcolor{blue!8}
\textbf{Domain}
&
\textbf{Evaluation}
&
\textbf{\shortstack{Muse Glimmer\\(low)}}
&
\textbf{\shortstack{Muse Glimmer\\(mid)}}
&
\textbf{\shortstack{Muse Glimmer\\(high)}}
&
\textbf{\shortstack{Muse Glimmer\\(xhigh)}}
\\
\midrule
\multirow{13}{=}{Agentic \& Coding}
&
$\tau$³-Banking
&
--
&
--
&
23.5
&
\textbf{26.0}
\\
&
WildClawBench
&
--
&
--
&
\textbf{47.6}
&
--
\\
&
GAIA2
&
--
&
--
&
\textbf{43.3}
&
--
\\
&
GDPval-AA
&
--
&
--
&
\textbf{953}
&
952
\\
&
SciCode
&
--
&
--
&
\textbf{43.6}
&
39.2
\\
&
OSWorld-Verified
&
58.7
&
66.6
&
65.9
&
\textbf{69.7}
\\
&
SWE-Bench Verified
&
74.6
&
75.2
&
\textbf{76.0}
&
75.3
\\
&
SWE-Bench Pro
&
50.5
&
50.9
&
51.2
&
\textbf{51.6}
\\
&
TauBench (Airline)
&
\textbf{79.0}
&
77.0
&
78.0
&
77.5
\\
&
TauBench (Retail)
&
86.0
&
\textbf{88.4}
&
87.1
&
86.8
\\
&
TauBench (Telecom)
&
91.6
&
92.5
&
\textbf{94.1}
&
92.1
\\
&
TerminalBench 2.1 (Terminus2)
&
--
&
--
&
51.7
&
\textbf{54.3}
\\
&
LiveCodeBench Pro (2025Q2)
&
36.2
&
55.8
&
64.1
&
66.8
--
\\
\midrule
\multirow{2}{=}{Tool Use}
&
PinchBench
&
89.0
&
\textbf{90.9}
&
90.5
&
90.7
\\
&
MCPAtlas
&
72.9
&
76.3
&
75.5
&
\textbf{77.4}
\\
\midrule
\multirow{5}{=}{Reasoning \& Knowledge}
&
DeepSearchQA
&
68.4
&
72.9
&
74.6
&
\textbf{75.3}
\\
&
AIME 2026
&
69.7
&
89.0
&
\textbf{94.7}
&
93.0
\\
&
GPQA Diamond
&
--
&
--
&
83.5
&
\textbf{85.6}
\\
&
ARC-AGI
&
34.6
&
58.8
&
66.9
&
\textbf{67.9}
\\
&
HLE
&
--
&
--
&
22.0
&
\textbf{23.4}
\\
\midrule
\multirow{2}{=}{Instruction Following}
&
IFBench
&
67.4
&
73.3
&
77.0
&
\textbf{77.1}
\\
&
SkillsBench
&
40.0
&
43.6
&
\textbf{44.3}
&
43.2
\\
\midrule
\multirow{5}{=}{Multimodal}
&
CharXiv Reasoning
&
75.4
&
77.9
&
\textbf{78.8}
&
78.6
\\
&
MMMU-Pro
&
--
&
--
&
74
&
--
\\
&
OmniDocBench v1.5
&
75.3
&
75.7
&
\textbf{75.8}
&
\textbf{75.8}
\\
&
Screenspot Pro
&
75.1
&
75.4
&
75.4
&
\textbf{75.7}
\\
&
Screenspot v2
&
93.2
&
92.9
&
\textbf{93.3}
&
92.5
\\
\midrule
Memory\&Long Context
&
BEAM-128K
&
65.0
&
65.6
&
65.1
&
\textbf{66.2}
\\
&
AA-LCR
&
--
&
--
&
\textbf{80.0}
&
\textbf{80.0}
\\
\midrule
\multirow{5}{=}{Privacy \& Security}
&
CIMemories (violation $\downarrow$)
&
33.4
&
31.1
&
26.4
&
\textbf{25.5}
\\
&
CIMemories (coverage $\uparrow$)
&
\textbf{66.4}
&
66.3
&
64.8
&
63.5
\\
&
PrivacyLens (leakage $\downarrow$)
&
59.6
&
53.0
&
\textbf{49.2}
&
59.6
\\
&
Siren AgentDojo (ASR@5 $\downarrow$)
&
29.9
&
27.9
&
28.4
&
\textbf{27.8}
\\
&
Siren AgentDojo (utility $\uparrow$)
&
\textbf{94.2}
&
\textbf{94.2}
&
\textbf{94.2}
&
\textbf{94.2}
\\
\midrule
\multirow{7}{=}{Multilingual}
&
Omnilingual-GAIA2
&
58.4
&
\textbf{60.7}
&
60.4
&
60.5
\\
&
Linguini
&
4.4
&
4.4
&
\textbf{8.8}
&
8.1
\\
&
Belebele
&
87.0
&
87.6
&
\textbf{87.8}
&
\textbf{87.8}
\\
&
MMLU-ProX
&
76.9
&
79.3
&
\textbf{80.1}
&
80.0
\\
&
WMT24++
&
53.2
&
53.3
&
\textbf{53.4}
&
53.3
\\
&
FLoRes+
&
47.0
&
47.4
&
\textbf{47.5}
&
\textbf{47.5}
\\
&
BOUQuET
&
35.6
&
36.3
&
\textbf{36.5}
&
\textbf{36.5}
\\
\bottomrule
\end{tabularx}
\end{table}

\newpage

\subsection{Agentic and Coding Task Completion}

The following benchmarks are used to assess the agentic and coding capabilities of the model.

\textbf{$\tau$³-bench Banking}~\citep{shi2026tauknowledgeevaluatingconversationalagents} is a knowledge-retrieval-based customer service evaluation extending $\tau$-bench to a fintech banking domain. The model acts as a customer-service assistant that must retrieve relevant information from unstructured knowledge
documents (policy guides, FAQ pages, compliance procedures) via configurable RAG pipelines, document search, and agentic shell-based search to resolve customer queries. The model interacts against a simulated user, with task success determined by satisfying per-task ground-truth assertions about the retrieved information and generated responses. Results are sourced from Artificial Analysis following the documented benchmarking methodology.\footnote{\url{https://artificialanalysis.ai/methodology/intelligence-benchmarking\#tau3-banking}}

\textbf{WildClawBench}~\citep{ding2026wildclawbenchbenchmarkrealworldlonghorizon} is an open-source benchmark for evaluating agentic models on real-world, long-horizon tasks. Its 60 tasks span six categories: productivity flow, code intelligence, social interaction, search and retrieval, creative synthesis, and safety alignment. Collectively, these tasks evaluate multi-step planning, coordinated tool use, multimodal understanding, information retrieval, coding, cross-modal content creation, and safe behavior. Each task runs in an isolated Docker container under the OpenClaw harness, with access to tools such as a browser, shell, file system, email, and calendar. Task-specific grading uses deterministic verification criteria, GPT-5.4 as an LLM judge, or a combination of both, depending on the task. Each grader produces an overall score between 0 and 1 by aggregating its task-specific criteria. We report the unweighted mean of the 60 task scores, averaged over three independent runs per task to reduce variance. We report internal run results for all 3 models.

\textbf{GDPval-AA v2}~\citep{patwardhan2025gdpvalevaluatingaimodel} assesses models on economically valuable knowledge work spanning 44 occupations across key sectors contributing to US GDP. The benchmark contains 220 tasks developed by OpenAI in collaboration with industry professionals to reflect real-world complexity. Each task requires the model to produce one or more deliverable files, run as an agent in AA's open-source Stirrup harness with an E2B code-execution sandbox, a 250-turn budget and access to a range of available tools. Results are sourced from Artificial Analysis following the documented benchmarking methodology.\footnote{\url{https://artificialanalysis.ai/methodology/intelligence-benchmarking\#gdpval-aa}}

\textbf{SciCode}~\citep{tian2024scicoderesearchcodingbenchmark} contains scientist-curated laboratory problems across 16 scientific disciplines, developed by domain experts with the reported evaluation using the 288 test set subproblems. The benchmark tests Python programming to solve scientific computing tasks: models must generate code that passes every unit test for a sub-problem to be credited. We prompt with the scientist-annotated background information included, and report sub-problem-level scoring rather than whole-problem scoring. No tool use is permitted. Results are sourced from Artificial Analysis following the documented benchmarking methodology.\footnote{\url{https://artificialanalysis.ai/methodology/intelligence-benchmarking\#scicode}} Scores reported here are from before the Muse Glimmer release and do not take into account Artificial Analysis Intelligence Index v4.2 changes that took place in September'26 and affected SciCode.\footnote{\url{https://artificialanalysis.ai/articles/artificial-analysis-intelligence-index-v4-2}}

\textbf{OSWorld-Verified}~\citep{OSWorld} is a multimodal benchmark for computer-use agents that evaluates the ability to operate a full Ubuntu desktop VM via GUI in realistic, cross-app workflows. We evaluate on the 361-task split, which excludes 8 Google Drive tasks. Agents are restricted to GUI-based computer control with no direct shell access: they observe the environment through screenshots and act via GUI actions such as click, type, scroll, and key press. Grading is execution-based, where each task defines programmatic getters and checkers that inspect the final VM state and return a 0--1 reward. Models receive $1920 \times 1080$ screenshots and predict click locations in a normalized coordinate space, with the ability to batch multiple GUI actions per step. Each episode is capped at 200 steps with a screenshot history truncation mechanism enabled that retains a maximum of 19 most recent screenshots. For \qwen and our model, we adopt the Claude computer-use action space (version computer\_20251124) with a normalized 0--1000 coordinate space plus a separate stop action. For Gemma4-31B, we use the Gemini 2.5 Flash computer-use interface, including click\_at, type\_text\_at, and stop, with relative coordinates in [0,999]. Following the official OSWorld protocol, we report mean per-task reward. The final score is averaged over four attempts per task to reduce variance. We report internal run results for all 3 models.

\textbf{TauBench.}~\citep{yao2024taubenchbenchmarktoolagentuserinteraction} is a multi-turn agentic dialogue evaluation where the model acts as a customer-service assistant against a simulated user, satisfying per-task ground-truth assertions on database state, actions, communication quality, and natural-language correctness. The model has access to domain-specific tools per split. We evaluate across three splits: airline (50 tasks), telecom (114 tasks), and retail (110 tasks). We compute average pass rate per split and we average results over 4 runs. We report internal run results for Muse Glimmer.

\textbf{SWE-Bench Verified and SWE-Bench Pro}~\citep{jimenez2024swebench} feature 500 and 731 diverse and complex agentic coding tasks, respectively. Our agent scaffolding consists of a bash tool and a file operation tool for viewing, creating and editing files. We report average pass rate across tasks and we average results across 4 runs to reduce variance. For SWE-Bench Verified, we report \qwen's self-reported score. For SWE-Bench Pro, we do not use \qwen's self-reported score which was measured on a refined version of the benchmark as we evaluated on the original task set. We report internal run results for the remaining cases.

\textbf{Terminal-Bench 2.1}~\citep{merrill2026terminal} (Stanford $\times$ Laude Institute) evaluates an agent on real-world tasks in a terminal environment. It contains 89 curated tasks across software engineering, system administration, data processing, model training, and security. The Terminus 2 agent harness is used in an E2B sandbox environment for the evaluation. Results are sourced from Artificial Analysis following the documented benchmarking methodology.\footnote{\url{https://artificialanalysis.ai/methodology/intelligence-benchmarking\#terminal-bench-v2-1}}

\textbf{GAIA2}~\citep{froger2026gaia2benchmarkingllmagents} is an open-source, text-only agentic tool-use benchmark that evaluates general-purpose assistants on multi-turn tasks inside simulated, stateful ``universes'' of everyday applications, such as email, calendar, messaging, contacts, shopping, and more. Each scenario runs in an isolated container hosting a self-contained environment: the agent fulfills user requests by issuing tool calls against the environment's app APIs, while an event daemon injects time-based and environment-driven events mid-episode (e.g., incoming messages, state changes, scheduled triggers) that the agent must notice and adapt to. We use the \href{https://github.com/facebookresearch/meta-agents-research-environments/}{public GitHub repository} for our evaluation, which contains 800 tasks across five capability splits (execution, search, ambiguity, adaptability, and time). We use the OpenClaw harness for our experiments. We measure the mean per-sample success rate over three independent runs per task, matching the public leaderboard convention. Grading is LLM-judged: an in-container judge compares the agent's realized tool actions and end state against each scenario's oracle events to produce a per-task success score; we use gpt-oss-120b as the reference judge. We report internal run results for all 3 models.

\textbf{LiveCodeBench Pro.}~\citep{zheng2025livecodebenchproolympiadmedalists} is a competitive programming benchmark sourced from Codeforces problems. We evaluate the 2025Q2 subset (Apr–Jun 2025) in C++, using the set of prompts, tests, and checkers aligned with the LiveCodeBench Pro release.\footnote{\url{https://github.com/GavinZhengOI/LiveCodeBench-Pro}} 2025Q2 is the most recent split with complete hidden tests provided. Models generate code solutions that we execute against the full test suites with checkers using our internal Code Execution Service, which is configured to match the official Codeforces contest environment. We run the benchmark internally and report pass@1 averaging over 8 attempts per problem to reduce variance.
          
\subsection{Tool Use \& Function Calling} 

This section evaluates the model's ability to use external tools and execute function calls to accomplish real-world tasks.

\textbf{PinchBench}~\citep{pinchbench2026}  is a container-based agentic evaluation, based on the open-source PinchBench benchmark, that measures how well an agent performs real-world knowledge-work tasks involving file manipulation, code execution, web access, and multi-step reasoning (e.g. market and competitor research reports, data analysis, email triage, and image generation). It consists of 24 tasks, each executed inside an isolated sandbox container running a scientific Python image. The agent acts as an autonomous software agent with access to a bash/execute\_command tool for terminal interaction, file read/write and code execution, and web-search and web-fetch tools. Each task is scored by combining the benchmark's deterministic automated checks with an LLM judge (gpt-oss-120b) score. The LLM judge grades the groundedness of the agent's deliverable against the evidence the agent itself produced (report contents, files, tool outputs).  We report the mean per-task reward and we average over 4 runs to reduce variance. Given the limited size of the benchmark (24 tasks) and the fact that many tasks are judge-graded over free-form deliverables, we treat the benchmark as primarily diagnostic. We report internal run results for Muse Glimmer.

\textbf{MCPAtlas}~\citep{bandi2026mcp} is a multi-turn agentic tool-use evaluation where the model uses 20+ MCP tool servers (such as GitHub, Twelve Data, Notion) to answer queries. Grading happens using an LLM judge (Gemini 2.5 Pro) that assesses whether the final model answer fulfills the ground-truth claims. The primary metric is mean pass rate, where a task successfully passes if it has a score higher than the 0.75 threshold. The benchmark consists of the 500 public tasks and results are averaged over 4 runs to reduce variance. We report internal run results for all 3 models.

\subsection{Reasoning \& Knowledge} 

This section evaluates the models' reasoning abilities and factual knowledge across a range of complex tasks.

\textbf{DeepSearchQA}~\citep{gupta2026deepsearchqa} is an agentic browsing evaluation where the model answers 900 hard research questions by acting as an autonomous web-browsing agent. It is given a single browsing tool with three functions: search, open and find. It works in a multi-step loop: it reasons, calls a function, receives the result as a new message, and repeats until it produces a final answer. Grading uses gpt-oss-120b as the judge to extract the model's predicted answer set, semantically matches predictions to targets, and computes an F1 score. We used the same search engine we selected across all models. Results are averaged over 4 runs to reduce variance. We report internal run results for all 3 models.

\textbf{AIME 2026}~\citep{dekoninck2026matharena}. We use the full 30-question datasets for AIME 2026. This benchmark features highly challenging, non-multiple-choice mathematics problems from the American Invitational Mathematics Examination. The questions require creative problem-solving, advanced algebraic manipulation, and deep number theory skills, where the final answer is always an integer between 000 and 999. Results are averaged over 10 runs to reduce variance. We report internal run results for our model, and self-reported numbers for \gemma and \qwen.

\textbf{GPQA Diamond} \citep{rein2024gpqa}. We use the full 198-question dataset. This subset contains high-quality, multiple-choice questions in biology, physics, and chemistry that were validated by domain experts to ensure they cannot be easily looked up on Google. The problems require deep, graduate-level scientific reasoning and rigorous technical understanding to solve. Results are sourced from Artificial Analysis following the documented benchmarking methodology.\footnote{\url{https://artificialanalysis.ai/methodology/intelligence-benchmarking\#gpqa-diamond}}

\textbf{ARC-AGI}~\citep{chollet2019arc}  is a benchmark consisting of 400 questions measuring abstract reasoning. The questions are novel visual grid puzzles designed to evaluate a model's ability to learn new concepts on the fly rather than relying on memorized training data. They require identifying fundamental core logic, such as object permanence, rotation, symmetry, scaling, and basic geometry, and applying those implicit rules to previously unseen grid configurations. Model answers are evaluated using a rule-based grader that performs exact-match comparison of the model's parsed output grids against the expected grids, marking an answer as correct when all test grids match exactly. We run the benchmark internally and report  mean success rate (pass@1) across four attempts.

\subsection{Instruction Following \& Steerability} 

The benchmarks below were used to evaluate the model's ability to follow instructions accurately and respond to user guidance.

\textbf{IFBench}~\citep{pyatkin2025ifbench} is an instruction-following benchmark that tests whether models can satisfy specific verifiable constraints embedded in prompts (e.g., word positioning, counting constraints, repetition patterns). The benchmark contains 294 tasks and we report mean pass rate across tasks. We average results across 4 runs to reduce variance. We report internal run results for \qwen and our model, and self-reported number for \gemma.

\textbf{SkillsBench}~\citep{li2026skillsbench} is an open-source benchmark that measures how effectively an agent can leverage ``skills''---modular folders of instructions and helper scripts---to complete specialized knowledge work tasks. We use the 86-task set, spanning a broad range of software and data engineering domains. Each task runs in an isolated container preconfigured from the task's Docker image and init commands with the relevant skill folders mounted into the workspace. The harness exposes a bash/terminal tool plus image-view and browser tools. Grading is deterministic and pytest-based: after the agent finishes, a hidden test suite is uploaded into the same container and executed, and the reward is the weighted fraction of tests passed. We report the mean per-task reward across four attempts. We report internal run results for all 3 models.

\subsection{Multimodal Understanding} 

We evaluate multimodal understanding using benchmarks spanning visual reasoning, document understanding and UI interaction.

\textbf{Charxiv Reasoning}~\citep{wang2024charxiv}. We use 1,000 chart reasoning questions from the CharXiv validation set. Grading uses gpt-oss-120b as an LLM judge, which evaluates semantic and mathematical equivalence between the model's response and the ground-truth answer. These multimodal questions are paired with diverse, complex scientific charts sourced directly from academic papers across multiple disciplines. The benchmark tests a model's ability to go beyond simple data extraction to perform complex visual synthesis, multi-step mathematical calculations, pattern recognition, and trend comparisons. We report the average task success rate across four attempts. We report internal run results for \gemma and our model, and a self-reported number for \qwen.

\textbf{MMMU-Pro}~\citep{yue2025mmmu} is a multimodal reasoning benchmark consisting of 1730 multiple choice questions. The benchmark presents college-level problems spanning 30 academic subjects across STEM, humanities, and social sciences, specifically filtered to remove any questions answerable by text-only models. The questions require expert-level reasoning and are paired with highly complex, heterogeneous image types, including diagrams, chemical structures, medical scans, and charts, where the answer choice space has been significantly expanded to minimize the success of random guessing. Results are sourced from Artificial Analysis following the documented benchmarking methodology.\footnote{\url{https://artificialanalysis.ai/methodology/intelligence-benchmarking\#mmmu-pro}}

\textbf{OmniDocBench v1.5}~\citep{ouyang2024omnidocbenchbenchmarkingdiversepdf}. We use the v1.5 version of the benchmark containing 1355 multimodal prompts. This benchmark evaluates a model's ability to accurately parse, extract, and reason over highly complex, multi-format documents. The inputs consist of diverse, real-world document pages, such as academic papers, textbooks, financial reports, slides, and handwritten notes, that challenge the model to simultaneously process dense text, complex multi-column layouts, tables, and embedded mathematical formulas. We report average scores across all tasks using an internal implementation of the original scoring protocol, that is modified to replace the three-component scoring formula (text, tables, formulas) with a two-component weighted average that folds formulas into text groups (graded by edit distance) rather than evaluating them separately with the Character-and-Delimiter Match metric. Our implementation uses a simpler Hungarian algorithm for bipartite element matching instead of the official v1.6 MGAM (Multi-Granularity Adaptive Matching) which adaptively searches for optimal prediction segmentation granularity while keeping ground truth fixed. Results are averaged over 4 runs to reduce variance. We report internal run results for all 3 models.

\textbf{ScreenSpot-Pro and v2}~\citep{li2025screenspotpro,wu2024osatlas} are open-source multimodal GUI-grounding benchmark that evaluate whether a model can locate a requested interface element in a static screenshot. For ScreenSpot-Pro, we evaluate the full 1,581-example set, which emphasizes high-resolution professional applications across Windows, macOS, and Linux. Each example provides a screenshot and a natural-language description of the target element. The model utilizes an iterative Python cropping tool. The model first predicts a point on the full screenshot in normalized [0, 1000] coordinates. We then render a grid-overlaid zoom around that point and map the model's crop-local prediction back to the original image. This loop repeats until the prediction converges, supporting up to 10 verification rounds and allowing up to three restarts from the original picture if the model considers that the target is absent from a crop. Grading is deterministic: a prediction receives reward 1 if the final point lies inside the annotated target bounding box and 0 otherwise. For ScreenSpot v2, we use the same coordinate output format for UI elements, but do not add any tool use since its image resolution is relatively low. We report mean point-in-box accuracy (pass@1), averaged over four independently sampled predictions per example to reduce variance. We report internal run results for all 3 models.

\textbf{Humanity's Last Exam (HLE)}~\citep{phan2025lastexam}. This benchmark consists of highly advanced, crowd-sourced questions spanning dozens of subjects that are specifically designed to be difficult even for human experts to answer. The questions test advanced scientific knowledge, complex abstract reasoning, and niche academic concepts. No tool use is permitted. We use the text-only subset of 2,158 questions. Results are sourced from Artificial Analysis following the documented benchmarking methodology.\footnote{\url{https://artificialanalysis.ai/methodology/intelligence-benchmarking\#hle}}

\subsection{Memory \& Long-Context} 

The following benchmarks were used to evaluate the model's ability to retain, retrieve and reason over information across long-context inputs.

\textbf{AA-LCR}~\citep{artificialanalysis2025lcr}  evaluates long-context performance by testing reasoning across multiple long documents. The questions are hard and text-based requiring genuine reasoning rather than simple data extraction. We use the full 100-question dataset spanning several document categories, such as academic papers, company financials, government consultations, legal documents, industry reports, and marketing materials. No tool use is permitted. Results are sourced from Artificial Analysis following the documented benchmarking methodology. Scores reported here are from before the Muse Glimmer release and do not take into account Artificial Analysis Intelligence Index v4.2 changes that took place in September'26 and affected AA-LCR.\footnote{\url{https://artificialanalysis.ai/articles/artificial-analysis-intelligence-index-v4-2}}

\textbf{BEAM-128K}~\citep{tavakoli2025milliontokensbenchmarkingenhancing} is a text-only, non-agentic long-context memory evaluation comprising 400 conversations spanning 128K token length across various domains, such as coding and math. It is designed to assess ten distinct memory abilities (e.g., factual recall, temporal reasoning, state tracking). Each sample is evaluated as a direct, single-turn recall task: the entire conversation history is placed in-context and the model must answer a memory-probing question without any scaffold or tool use. Grading is LLM-judged using GPT-5.4, which scores model responses against reference answers. We measure mean pass@1 across four attempts. We report internal run results for all 3 models.

\subsection{Multilingual}
\label{app:evals:multilingual}
\color{black}

The following benchmarks were used to evaluate the model's multilingual ability. Our multilingual evaluation is text-only; multilingual multimodal understanding is not assessed here.

\textbf{OmnilingualGAIA2} \citep{omnigaia2} is a multilingual agentic benchmark: agents must fulfill user tasks that are phrased in target language by planning, searching, and acting in a universe that is also in target language. It extends 4 of the core capabilities (search, execution, ambiguity, adaptability) of GAIA2 \citep{froger2026gaia2benchmarkingllmagents} to 10 languages, and grades agent actions and responses with a verifier comprising a multilingual LLM-as-a-judge component. We report pass@1 averaged across 3 runs to reduce variance.

\textbf{Linguini} \citep{sanchez2026linguini} is a text-only benchmark of language-agnostic linguistic reasoning: 894 questions grouped into 160 problems taken from the International Linguistic Olympiad, covering 75 mostly low-resource languages. Each problem gives a handful of parallel examples in a language the model has almost certainly never seen, and the model must induce the underlying morphological, phonological, or numeral system from those examples alone. Prior knowledge of the language is neither required nor helpful, so the benchmark measures reasoning rather than recall. Problems come in five formats: translation into and out of the unknown language, filling in blanks, matching letters, and converting between numerals and their written forms. Each sample is a direct, single-turn task: context, question, and answer format are given in-context, and the model produces the requested word or phrase with no scaffold or tool use. Grading is reference-based and automatic, using normalized exact match. We report whole-problem exact match, which credits a problem only if all of its answers are correct, and discuss per-question accuracy over all $894$ questions in~\cref{sec:app:multilingual}.

\textbf{Belebele} \citep{belebele} is a text-only multiple-choice reading comprehension benchmark covering 122 language variants. Every question exists in every language (the set is fully parallel), so scores are directly comparable across languages. Each question is grounded in a short passage drawn from FLORES \citep{nllb-24} and offers four answer options, written to separate strong readers from weak ones: the English subset alone is hard enough to challenge frontier models. Each sample is a direct, single-turn task: the passage, question, and four options are given in-context and the model selects one option, with no scaffold or tool use. Grading is exact match against the gold option. We report mean accuracy.

\textbf{MMLU-ProX} \citep{xuan-etal-2025-mmlu} is a text-only multilingual knowledge-and-reasoning benchmark extending the reasoning-focused MMLU-Pro design to typologically diverse languages, with almost 12,000 questions per language produced by a semi-automatic pipeline in which state-of-the-art LLM translations are verified by expert annotators for conceptual accuracy and terminological consistency. We evaluate on the full test set of 29 languages, with \texttt{lm-eval}\footnote{\url{https://github.com/eleutherai/lm-evaluation-harness}} default settings and 0-shot prompting, that renders \glimmermodelname's reasoning strength block in its expected position. We report mean pass@1 over the languages.

\textbf{WMT24++} \citep{deutsch-etal-2025-wmt24} is a text-only, multi-way machine translation benchmark that extends the WMT24 test set to 55 languages and dialects, adding newly collected human-written references and post-edits for 46 new languages/dialects alongside post-edits of the original references in 8 of the 9 WMT24 languages. It spans four domains: literary, news, social, and speech. Each sample is evaluated as a direct, single-turn translation task: the model receives the source segment out-of-English and must produce the target-language translation with no scaffold or tool use; for the into-English direction, we reverse the test set, translating the human references back into English and scoring against the original English source. Grading is reference-based and automatic using chrF++ \citep{popovic-2015-chrf} against the human references. 

\textbf{FLORES+} \citep{nllb-24} is a text-only multi-way parallel translation benchmark covering ~200 languages, built from English Wikimedia articles professionally translated into every target language and released in dev/devtest/test splits. We evaluate on the public devtest split of 1,012 sentences per language. Again, each sample is a direct, single-turn sentence-level translation task and grading is reference-based and automatic using chrF++ against the human references. 

\textbf{BOUQuET} \citep{bouquet} is a text-only, multi-way, multicentric, and multi-register/domain, handcrafted first in 9 non-English source languages (Egyptian Arabic, Modern Standard Arabic, French, German, Hindi, Indonesian, Mandarin Chinese, Russian, Spanish) chosen as high-coverage pivot languages. Unlike sentence-level benchmarks, it is organized in paragraphs of varying length, and its non-English-origin, multicentric design is intended to avoid contamination and to enforce representation of multilingual language features. Similarly to above, each sample is a direct, single-turn sentence-level translation task and grading is reference-based and automatic using chrF++ against the human references.

\subsection{Policy Adherence} 
\subsubsection{Data Handling at Inference Time}

This benchmarks below were used to evaluate the model's ability to preserve user privacy and resist adversarial manipulation.

\textbf{PrivacyLens.}~\citep{shao2024privacylens} We use the full 493-prompt dataset, which evaluates privacy norm awareness in LLMs acting as AI assistants. Each prompt places the model in a simulated agentic context: the system prompt describes available user tools (e.g., email, contacts), and the user message includes a pre-executed trajectory of prior tool calls followed by a user instruction. The model then generates a single-turn response (email, post, message) without exposing sensitive information. This is a text-only, non-multimodal evaluation.  Grading uses LLM-as-judge (Claude Sonnet 4.6), which evaluates both (1) information exposure, whether the response reveals any of the scenario's information items (binary per item), and (2) helpfulness on a 0–3 scale. The metric reported is the adjusted exposure rate, fraction of samples where any item is exposed, conditioned on the response being helpful (helpfulness $\geq$ 2). We run the benchmark internally and we average results over 4 runs to reduce variance.

\textbf{CIMemories}~\citep{mireshghallah2026cimemories}. CIMemories is a benchmark measuring how much models reveal personal sensitive information in inappropriate contexts. CIMemories uses synthetic user profiles with over 100 attributes per user (such as finance, health, relationships attributes) paired with diverse task contexts representing canonical social interactions (e.g., communicating with doctors) resulting in a total of 2500 scenarios. Each attribute may be essential for some tasks but inappropriate for others, giving rise to two primary metrics, violation rate (the extent to which inappropriate attributes are revealed) and coverage rate (the extent to which necessary attributes are shared). We use Claude 4.6 Sonnet as an LLM judge to determine sharing of attributes by the model and we average results over 4 runs to control variance. We report internal run results for all 3 models.

\subsubsection{Robustness to Prompt Injections}
\textbf{Siren AgentDojo}~\citep{debenedetti2024agentdojo}. Siren AgentDojo is a public benchmark designed to evaluate how models handle standard tool-use scenarios when exposed to prompt injection attacks. The evaluation combines 97 benign tasks spread across four distinct environments (banking, travel booking, workspace management, online store) with 35 malicious tasks yielding a total of 949 prompt injection attempt scenarios. Each of these equips the agent with domain-specific tools, such as transaction APIs, calendar access, and file operations. The benchmark specifically tests whether agents can be manipulated by stealthy attacks into misusing these tools for malicious actions like data exfiltration or unauthorized financial transactions. The LLM-powered attacker (we use Claude Opus 4.6) iteratively refines its attack until it succeeds or a maximum of 6 iterations is reached. We use deterministic rule-based grading to determine attack success and benign task completion. We report two scores here: Attack Success Rate (ASR), which is the success rate of the malicious task injected by the attacker and utility, which is the success rate of the benign task instructed by the user in the presence of a malicious injection. Higher utility and lower ASR indicate that the agent is able to complete more user requests, while being less impacted by the injected prompts. We report internal run results for all 3 models.

\section{Implementation Details}

\subsection{System Prompt Example - Assistant}
\begin{tcolorbox}
\ttfamily

You are a helpful AI assistant.\\
Knowledge cutoff: 2026-01-04.\\
Current date: 2026-03-15.\\

In this environment you have access to a set of tools you can use to answer the user's question.
You can invoke a function by writing a "<atem:function\_calls>" block like the following as part of your reply to the user:\\

<atem:function\_calls>
<atem:invoke name="\$FUNCTION\_NAME">
<atem:parameter name="\$PARAMETER\_NAME">\$PARAMETER\_VALUE</atem:parameter>
...
</atem:invoke>
</atem:function\_calls>\\

String and scalar parameters should be specified as is, while lists and objects should use JSON format.\\

Here are the functions available:\\
{"name": "bash.exec", "description": "Execute a bash command in the terminal. Commands will time out after 120 seconds.", "parameters": {"type": "object", "properties": {"cmd": {"type": "string", "description": "The bash command to execute."}}, "required": ["cmd"]}}\\
{"name": "str\_replace\_editor.create", "description": "Create a new file with the given content.", "parameters": {"type": "object", "properties": {"path": {"type": "string", "description": "Absolute path for the new file."}, "file\_text": {"type": "string", "description": "The content to write."}}, "required": ["path", "file\_text"]}}\\
{"name": "submit.submit", "description": "Submit the solution when the task is complete.", "parameters": {"type": "object", "properties": {}, "required": []}}\\

Here is an example of how to use the tools:\\

<atem:function\_calls>
<atem:invoke name="bash.exec">
<atem:parameter name="cmd">find /testbed -type f -name "*.py" | head -20</atem:parameter>
</atem:invoke>
</atem:function\_calls>\\

\#\# Task Summary:\\
In your last turn, provide a short summary...\\

\#\# Additional Instructions:\\
- Do not use interactive commands\\
- No internet access\\
- Reproduce the issue before fixing\\
- Run existing tests\\

Reasoning strength: high.\\

\# Valid recipients: "self", "bash.exec", \\"str\_replace\_editor.view", "str\_replace\_editor.str\_replace",\\ "str\_replace\_editor.create", "submit.submit", "user".\\

\end{tcolorbox}
\newpage

\subsection{System Prompt Example - OpenClaw}

\begin{tcolorbox}
\ttfamily

You are a personal assistant running inside OpenClaw.

Tooling:

You may use the listed tools exactly by name. Available tools include read, write, edit, exec, process, web\_search, web\_fetch, browser, canvas, nodes, cron, message, gateway, session/session-agent tools, memory tools, and tts. TOOLS.md is guidance only, not the source of tool availability. For complex or long tasks, spawn a sub-agent.

Tool Call Style:

Do not narrate routine low-risk tool calls. Narrate briefly only for complex, multi-step, or sensitive actions.

Safety:

You have no independent goals. Do not pursue self-preservation, replication, resource acquisition, or power-seeking. Prioritize safety and human oversight. If instructions conflict, pause and ask. Do not bypass safeguards or manipulate anyone to expand access.

OpenClaw CLI:

Use documented OpenClaw subcommands only:

- openclaw gateway status \\
- openclaw gateway start \\
- openclaw gateway stop \\
- openclaw gateway restart \\
If unsure, ask for `openclaw help` output. \\

Skills:

Before replying, scan available skills. If one clearly applies, read its SKILL.md and follow it. If multiple apply, choose the most specific. If none apply, do not read a skill. Never read more than one skill up front.\\
Memory Recall:\\
Before answering about prior work, decisions, dates, people, preferences, or todos, run memory\_search on MEMORY.md + memory/*.md, then memory\_get only needed lines. Include source citations when useful.\\
Self-update:\\
Only run OpenClaw updates/config changes when explicitly requested.\\
Workspace:\\
Working directory is `/root/.openclaw/workspace`. Treat it as the global workspace unless instructed otherwise. Commit changes after edits.\\
Documentation:\\
Use local OpenClaw docs at `/opt/chatbot/docs` first for OpenClaw behavior, commands, config, or architecture.\\
Messaging:\\
Reply in current session routes automatically to the source channel. Use `message` only for proactive sends/channel actions. Never use exec/curl for provider messaging.\\
Project Context:\\
Loaded project files include AGENTS.md, SOUL.md, TOOLS.md, IDENTITY.md, USER.md, HEARTBEAT.md, BOOTSTRAP.md, plus silent-reply, heartbeat, and runtime rules.\\

\end{tcolorbox}
\newpage

\subsection{Synthetic Data: Persona Generation Details}
\label{app:personas}

Our pipeline supports two persona generation modes. In the \emph{sampling-based} mode, we sample age
from a Beta(2,5) distribution skewed toward 25--35 with a floor of 18, and region from a weighted
categorical distribution reflecting a realistic AI-user population split across North America, Europe,
Asia, India, the Middle East, Latin America, and Africa.
Big Five personality scores (openness,
conscientiousness, extroversion, agreeableness, neuroticism) are independently drawn from a truncated
Gaussian ($\mu=5$, $\sigma=2.5$, clamped to $[1,10]$).
These sampled attributes are passed to the model,
which is prompted to synthesize a coherent identity—name, occupation, marital status, work style,
digital habits, hobbies, life habits, dreams, and concerns—consistent with the sampled psychometric profile.
In the \emph{text-grounded} mode, we instead sample raw text passages from various sources and prompt the model to infer several plausible personas (each with an explanation and a
plausibility/probability score) that could have authored or engaged with each passage, together with a
free-form background paragraph covering routine, relationships, and goals. This mirrors recent
findings that grounding persona synthesis in real long-form text, rather than a purely LLM-sampled
demographic schema, produces persona populations whose psychometric trait distributions more closely
track real human populations than temperature-based sampling alone~\citep{hu2025population}.

\subsection{Request Format}

\glimmermodelname natively supports a set of tunable parameters: Reasoning Strength, Valid Recipients, and Agent Tool Calling Format (ATEM).

\paragraph{Reasoning strength} controls the depth of internal chain-of-thought the model performs before producing a response. It is specified as a single line in the system prompt: 
\begin{verbatim} Reasoning strength: <level>. \end{verbatim} 
\noindent 
Four levels are supported: \begin{itemize} 
\item \textbf{Low} -- Minimal reasoning; a brief internal thought before responding. Suited for simple factual lookups or trivial tasks. 
\item \textbf{Medium} -- Moderate reasoning; balanced thinking for medium-complexity tasks requiring some planning. 
\item \textbf{High} -- Full reasoning; deep chain-of-thought for complex multi-step coding, debugging, or research tasks. This is the default for agent workloads. 
\item \textbf{xHigh} -- Maximum reasoning; the largest available thinking budget, for the hardest long-horizon agentic and research tasks where quality outweighs latency. Reserved for offline or evaluation settings rather than interactive use.\color{black}
\end{itemize} 

\paragraph{Valid Recipients:} Each message the model produces carries a \texttt{to=} routing header that determines how the infrastructure processes the message. The system prompt declares the set of available recipients via a directive at the end of the prompt: \begin{verbatim} # Valid recipients: "self", "bash.exec", "str_replace_editor.view", "user".\end{verbatim} 
\noindent 

Three recipient types exist: \begin{table}[h] \centering \small \begin{tabular}{lll} \toprule \textbf{Type} & \textbf{Syntax} & \textbf{Semantics} \\ \midrule Self & \texttt{to=self} & Internal reasoning and planning \\ Tool & \texttt{to=\textit{namespace.function}} & Triggers tool execution \\ User & \texttt{to=user} & Final response; ends the turn \\ \bottomrule \end{tabular} \caption{Recipient types in the \glimmermodelname routing protocol.} \label{tab:recipients} 
\end{table} 

The model must only route to recipients explicitly listed in the valid recipients directive. Each tool function receives its own entry (e.g., \texttt{str\_replace\_editor.view} and \texttt{str\_replace\_editor.str\_replace} are separate recipients). The recipients \texttt{self} and \texttt{user} are always present, and routing to \texttt{user} terminates the model's turn.

\paragraph{Agent Tool Calling Format (ATEM):}ATEM is the XML-based format used for tool invocation. A tool call consists of a routing header matching the tool's fully qualified name, followed by an \texttt{atem:function\_calls} XML block. \subsubsection{Tool Definitions} Tools are defined in the system prompt as JSON schema objects specifying name, description, and parameters: 

\begin{lstlisting}[style=json, breaklines=true]
{
"name": "bash.exec",
"description": "Execute a bash command.", 
"parameters":{
  "type": "object", 
  "properties": {
    "cmd": {
       "type": "string",
       "description": "The bash command."
     }
    },
    "required": [
     "cmd"
    ]
   }
} 
\end{lstlisting}

\noindent 

Tool names follow the \texttt{namespace.function} convention.

\subsubsection{Invocation Format} 
To call a tool, the model emits: \begin{lstlisting}[basicstyle=\ttfamily\small,breaklines=true] 
to=bash.exec <atem:function_calls> <atem:invoke name="bash.exec"> <atem:parameter name="cmd">ls /workspace</atem:parameter> </atem:invoke> 
</atem:function_calls> 
\end{lstlisting} \noindent 

String and scalar parameters are written as-is; arrays and objects use JSON syntax. Multiple \texttt{atem:invoke} elements may appear within a single \texttt{atem:function\_calls} block for parallel tool calls.

Tool output wrapping similarly supports multiple styles during training---including canonical XML, markdown code blocks, bracket/arrow/pipe/hashtag delimiters, and ReAct-style \texttt{Observation:} prefixes---to improve robustness across diverse deployment contexts.
\newpage

\subsection{Agentic Data Format}
\label{sec:format_data}

Assistant messages use recipient to distinguish private reasoning (self), tool calls (<tool>.<function>), and the final user-facing answer (user). Tool calls are rendered as <atem:function\_calls> XML in body\_text and mirrored in metadata.tool\_calls. External tool observations are wrapped with a SECURITY NOTICE header and <<<EXTERNAL\_UNTRUSTED\_CONTENT>>> sentinels so the model treats their contents as untrusted. Some splits additionally prefix user turns with a Sender (untrusted metadata) block. An abbreviated example follows.
\begin{lstlisting}[style=json, caption={Experiment configuration}]
{
  "id": "b1a7c0de-...",
  "metadata": {
    "type": "sft", "split": "train",
    "path": "openclaw-sft/check-weather",
    "domain": "agent/openclaw", "language": "en",
    "quality_metrics": {
      "scenario": "check-weather", "reward": 1.0,
      "passed_count": 2, "total_checks": 2
    }
  },
  "past_conversation_context": {"messages": [
    {"author": {"role": "system"},
     "content": {"content_train_params": {"keep_loss": false, "loss_weights": 0.0},
                 "customized_instructions": "You are a personal assistant ...",
                 "tool_set": {"tools": {"wttr": {"...": "..."}},
                              "schema_render_format": "xml_atem",
                              "tool_call_type": "<split-specific>"},
                 "reasoning_strength": {"strength": 64},
                 "allow_private_thinking": true, "allow_public_thinking": false},
     "end_turn": true},
    {"author": {"role": "user"},
     "content": {"content_train_params": {"keep_loss": false, "loss_weights": 0.0},
                 "body_text": "Check the weather for my location and summarize."},
     "end_turn": true}
  ]},
  "continuations": [{"conversation": {"messages": [
    {"author": {"role": "assistant"},
     "content": {"content_train_params": {"keep_loss": true, "loss_weights": 1.0},
                 "body_text": "I'll look up their location and call the weather tool."},
     "end_turn": false, "recipient": {"recipient": "self"}},
    {"author": {"role": "assistant"},
     "content": {"content_train_params": {"keep_loss": true, "loss_weights": 1.0},
                 "body_text": "<atem:function_calls><atem:invoke name=\"wttr.wttr\">...</atem:invoke></atem:function_calls>"},
     "end_turn": true, "recipient": {"recipient": "wttr.wttr"},
     "metadata": {"tool_calls": [{"tool_name": "wttr", "function_name": "wttr",
                                  "arguments": {"location": "London", "format": "j1"}}]}},
    {"author": {"role": "tool", "name": "wttr.wttr"},
     "content": {"content_train_params": {"keep_loss": false, "loss_weights": 0.0},
                 "body_text": "<tool_output name=wttr.wttr>SECURITY NOTICE: ... <<<EXTERNAL_UNTRUSTED_CONTENT id=\"7f10c2\">>>{\"temp_F\": \"59\", ...}<<<END_EXTERNAL_UNTRUSTED_CONTENT id=\"7f10c2\">>></tool_output>"},
     "end_turn": true},
    {"author": {"role": "assistant"},
     "content": {"content_train_params": {"keep_loss": true, "loss_weights": 1.0},
                 "body_text": "About 59F and partly cloudy in London today."},
     "end_turn": true, "recipient": {"recipient": "user"}}
  ]}}]
}
\end{lstlisting}

\newpage

\subsection{Multilingual Capabilities}
\label{sec:app:multilingual}
We evaluate \glimmermodelname's ability to deal with a broad range of tasks involving multiple languages. Because agents are deployed by non-English-speaking users, we evaluate multilingual ability along two axes: linguistic competence, via translation and reading comprehension; and the transfer of task competence out of English, for knowledge and reasoning and for agentic behaviour. We separate the two because they do not move together, translation quality is a poor predictor of agentic readiness.

We also rely on machine translation as a proxy task to establish a set of \emph{supported languages} of \glimmermodelname, across two tiers: \emph{strong} (88 languages) and \emph{usable} (49 languages), for a total of \textbf{137 language varieties supported}.

In the following we denote language varieties following the FLORES-200~\citep{nllb-24} notation and the more familiar language name in parenthesis, e.g. \texttt{eng\_Latn (English)}, where \texttt{eng} is the ISO-639-3 code that identifies the \emph{languoid} and \texttt{Latn} is the ISO-15924 code that identifies the underlying \emph{script}.

\subsubsection{Multilingual Agentic Capabilities}

\dataset{OmnilingualGAIA2}~\citep{omnigaia2}, a multilingual extension of \dataset{GAIA2}~\citep{froger2026gaia2benchmarkingllmagents} spanning $10$ languages and $4$ core agentic capabilities, allows us to directly assess how the English-only capabilities transfer to a multilingual setting.

We compare \glimmermodelname at \emph{low}, \emph{medium}, \emph{high} and \emph{xhigh} reasoning strength to \gemma and \qwen, and we report results across languages in~\cref{tab:app:omnilingualgaia2-results}.

The observed strong agentic performance of \glimmermodelname{} in English transfers to the selection of high resource languages (all are in our \emph{strong} support tier) covered by the benchmark.
At every tested reasoning strength, \glimmermodelname{} is the top-performing model in all $10$ languages: over the full $10\times4$ grid its margin over \gemma ranges from $16.4$ points (Japanese, \emph{xhigh}) to $27.2$ (Hindi, \emph{medium}), and its margin over \qwen from $10.8$ points (French, \emph{low}) to $23.0$ (German, \emph{high}).
Averaged over the $10$ target languages, \glimmermodelname{} leads \gemma by $21.0$--$23.2$ points and \qwen by $13.4$--$15.7$ points, depending on reasoning strength.

However, the multilingual transfer is not perfect, nor is it for the other tested models. We observe a drop across the board comparing English performance to non-English performance: at \emph{high}, \glimmermodelname{} falls from $73.7$ pass@1 (averaged across $3$ rollouts) in English to $50.0$ in Japanese, a $23.7$ point gap, with the $10$ target languages recording an average drop of $13.3$ ($15.3$ at \emph{low}, $13.0$ at \emph{medium}, $13.4$ at \emph{xhigh}).
All the \emph{strongly supported} languages lose ground: setting Japanese aside, drops at \emph{high} range from $7.8$ in German to $20.1$ in Hindi, relative to English.
Both models we compare against experience a noticeable drop as well, although of varying magnitude: $11.6$ on average and $20.0$ at worst for \qwen, which ranks English first and Japanese last; $5.7$ on average and $10.7$ at worst for \gemma, which also peaks in English but bottoms out in Hindi, with Japanese a close second-to-last.
Furthermore, the three models share the same capability ordering (average performance on search $>$ execution $>$ adaptability $>$ ambiguity), although exact ranking across languages can vary.
In particular, \qwen reports a comparable English-to-Japanese spread ($20.0$) while \gemma seems to be able to retain more of its English performance (worst-case drop of $10.7$), although this can be considered an artifact of a depressed English ceiling ($43.2$).

Reasoning strength has a modest, and not always monotonic, effect. Averaged over the $10$ target languages, \glimmermodelname{} gains $2.3$ points moving from \emph{low} to \emph{medium} ($58.4 \rightarrow 60.7$) and then plateaus ($60.4$ at \emph{high}, $60.5$ at \emph{xhigh}), with \emph{high} and \emph{xhigh} never differing by more than $0.7$ points in any single language.
The plateau is not uniform, though: seven of the ten languages peak at \emph{high} or \emph{xhigh}, whereas Chinese, Hindi and Japanese peak at \emph{medium}.
The regression is negligible for Chinese ($0.1$ points from \emph{medium} to \emph{high}) but substantial for Hindi ($6.1$) and Japanese ($3.0$), and in both cases it is driven almost entirely by the search capability ($81.8 \rightarrow 69.9$ for Hindi, $77.8 \rightarrow 68.5$ for Japanese), while every other language stays within $\pm 2$ points on search over the same transition.
Hindi and Japanese are also the two lowest-scoring languages in the set, so the languages that would benefit most from additional test-time reasoning are precisely those where scaling reasoning strength beyond \emph{medium} hurts.

\begin{table}[ht]
\caption{Overall performance (pass@1, macro-averaged across the $4$ capabilities) across the $10$ languages of \dataset{OmnilingualGAIA2}. We also report the English performance of \dataset{GAIA2} as anchor. \emph{Target mean} averages the $10$ non-English languages, and $\Delta$ vs.\ English is its difference to the English row. Best result per row in bold.}
\label{tab:app:omnilingualgaia2-results}
\centering
\footnotesize
\setlength{\tabcolsep}{4pt}
\renewcommand{\arraystretch}{1.15}
\begin{tabular}{lcccccc}
\toprule
\textbf{Language} & \textbf{\shortstack{\glimmermodelname\\(low)}} & \textbf{\shortstack{\glimmermodelname\\(medium)}} & \textbf{\shortstack{\glimmermodelname\\(high)}} & \textbf{\shortstack{\glimmermodelname\\(xhigh)}} & \textbf{\shortstack{\gemma\\(thinking)}} & \textbf{\shortstack{\qwen\\(thinking)}} \\
\midrule
\texttt{eng\_Latn} (English)	     & 73.7 & 73.7	& 73.7	& \textbf{73.9}	& 43.2	& 56.6 \\
\midrule
\texttt{spa\_Latn} (Spanish)	     & 61.6	& 63.9	& \textbf{65.0}	& 64.4	& 39.5	& 50.4 \\
\texttt{fra\_Latn} (French)	     & 60.5	& 63.3	& 64.0	& \textbf{64.6}	& 40.8	& 49.7 \\
\texttt{por\_Latn} (Portuguese)	 & 62.7	& 64.6	& 64.9	& \textbf{65.3}	& 40.8	& 48.3 \\
\texttt{ind\_Latn} (Indonesian)	 & 59.2	& 62.0	& \textbf{62.1}	& \textbf{62.1}	& 39.0	& 48.2 \\
\texttt{ita\_Latn} (Italian)	     & 58.8	& 60.7	& 61.8	& \textbf{62.3}	& 39.7	& 46.6 \\
\texttt{tur\_Latn} (Turkish)	     & 57.6	& 60.1	& 60.1	& \textbf{60.7}	& 36.5	& 44.6 \\
\texttt{cmn\_Hans} (Chinese)	     & 55.4	& \textbf{56.5}	& 56.4	& 56.4	& 34.1	& 43.7 \\
\texttt{deu\_Latn} (German)	     & 62.0	& 63.2	& \textbf{65.9}	& 65.7	& 38.8	& 42.9 \\
\texttt{hin\_Deva} (Hindi)	     & 56.5	& \textbf{59.7}	& 53.6	& 54.1	& 32.5	& 38.9 \\
\texttt{jpn\_Jpan} (Japanese)	     & 49.8	& \textbf{53.0}	& 50.0	& 49.3	& 32.9	& 36.6 \\
\midrule
\emph{Target mean}	                 & 58.4	& \textbf{60.7}	& 60.4	& 60.5	& 37.5	& 45.0 \\
$\Delta$ vs.\ English	             & $-15.3$	& $-13.0$	& $-13.3$	& $-13.4$	& $-5.7$	& $-11.6$ \\
\bottomrule
\end{tabular}
\end{table}

\subsubsection{Non-Agentic Multilingual Capabilities} 

We further study how the non-agentic capabilities of \glimmermodelname transfer from English into other languages, and we report below our main findings.

\paragraph{Strong metalinguistic reasoning.} 
The good performance observed on \dataset{OmnilingualGAIA2} suggests strong multilingual reasoning capabilities of \glimmermodelname. To evaluate the ability of the model to perform \emph{metalinguistic} reasoning, i.e. understanding and operating under the rules of (potentially unseen) languages, we leverage \dataset{Linguini} \citep{sanchez2026linguini} as it deliberately removes language knowledge from the equation: each problem supplies all the evidence needed in context, so a model cannot fall back on having seen the language in training. 
\glimmermodelname{} leads by a wide margin on whole-problem exact match ($8.8$, against $3.8$ for \gemma and $2.5$ for \qwen), solving more than twice as many complete problems as the strongest baseline. 
The per-question view narrows and reverses the ordering at the top  
($17.9$ versus \gemma's $18.8$, and $14.4$ for \qwen)%
, indicating that \glimmermodelname{} more often carries a deduction through to every requested answer, while \gemma recovers a comparable share of individual answers across a wider set of partially-solved problems. 
Neither ordering follows from the translation tiers: the 75 languages in \dataset{Linguini} lie almost entirely outside the supported set, and performance here reflects inductive reasoning over unfamiliar morphology rather than coverage of the language itself.

\paragraph{Reading comprehension and knowledge transfer well.} 
We investigate whether the good agentic and metalinguistic reasoning capabilities emerge at the expense of more conventional multilingual tasks. 
On \dataset{Belebele} \citep{belebele}, a machine reading comprehension benchmark spanning $122$ languages, \glimmermodelname{} scores $87.8\%$, ahead of \qwen ($86.1\%$) and just behind \gemma ($88.6\%$); all three cluster within $2.5$ points, indicating that multilingual reading comprehension is not the discriminating axis at this scale. 
We find similar results on \dataset{MMLU-ProX} \citep{xuan-etal-2025-mmlu}, measuring domain-specific knowledge in a multilingual setting, where \glimmermodelname{} scores $80.1$ mean pass@1, behind \gemma ($82.5$) and \qwen just slightly ahead ($80.5$).
Remarkably, \glimmermodelname{} exhibits the \emph{tightest} multilingual spread of the three once we exclude Wolof---a language for which \glimmermodelname{} currently lacks sufficient coverage: the average English-to-language gap is just $2.6$ points for \glimmermodelname{} (high), compared to $3.7$ for \gemma and $5.2$ for \qwen. Overall performance also converges under this view: \glimmermodelname{} (high) surpasses \qwen ($81.5$ vs. $81.1$) and narrows the gap with \gemma ($83.0$). In other words, \glimmermodelname{} trades $1.5$ points of average accuracy for a more uniform cross-lingual profile.

\paragraph{\glimmermodelname is token efficient.} 
Absolute accuracy on \dataset{MMLU-ProX} hides a large spread in generation cost (Table~\ref{tab:token-efficiency}). \gemma and \glimmermodelname{} are close on mean tokens per rollout ($2{,}671$ vs.\ $2{,}453$), while \qwen spends $4{,}532$---$70\%$ more than \gemma and $1.8\times$ as much as \glimmermodelname{} in \emph{high} mode. 
Which model is preferable therefore depends on whether the deployment is accuracy-bound or throughput-bound: excluding Wolof \gemma buys $1.5$~pp of accuracy for $\sim10\%$ more tokens; \qwen spends 1.8$\times$ the tokens for no accuracy advantage at all.

\paragraph{Reasoning strength does not change these conclusions.} Raising reasoning strength from \emph{high} to \emph{xhigh} moves every multilingual result by less than the run-to-run noise: $53.3$ vs $53.4$ on \dataset{WMT24++}, $47.5$ on both for \dataset{FLORES+}, $36.5$ on both for \dataset{BOUQuET}, $87.8$ on both for Belebele, and $8.1$ vs $8.8$ on \dataset{Linguini}, at $10$--$27\%$ more generated tokens. Therefore, all our results are done with \emph{high} as the default operating point.

\begin{table}[ht] 
\centering 
\begin{tabular}{lcc} 
\toprule 
Model & pass@1 & Tokens/rollout \\ 
\midrule 
\glimmermodelname{} (low) & 76.9 & 557 \\ 
\glimmermodelname{} (medium) & 79.3 & 1{,}180 \\ 
\glimmermodelname{} (high) & 80.1 & 2{,}453 \\ 
\glimmermodelname{} (xhigh) & 80.0 & 3{,}397 \\ 
\gemma & \textbf{82.5} & 2{,}671 \\ 
\qwen & 80.5 & 4{,}532 \\ 
\bottomrule 
\end{tabular} 
\caption{Accuracy vs.\ generation cost on MMLU-ProX. Tokens/rollout includes the reasoning trace.} 
\label{tab:token-efficiency} 
\end{table}

\subsubsection{Machine Translation Performance and Detailed Language Support}

We use the performance of \glimmermodelname on machine translation tasks, as measured by \textsc{chrF++}~\citep{popovic-2015-chrf}, on the union of \dataset{WMT24++}~\citep{deutsch-etal-2025-wmt24}, \dataset{FLORES+}~\citep{nllb-24} and \dataset{BOUQuET}~\citep{bouquet}, to define language support tiers.
In particular, we measure the performance of the model in translating text from at least one benchmark \emph{both} in and out of English, and define two language tiers.
We say a language variaty is \emph{strongly supported} by \glimmermodelname if, based on the protocol above, it scores at least $50$, while we say it's \emph{usable} if its score is in the $[40, 49]$ range. Note that for tier assignment we exclude empty generations, while the aggregate scores reported in Tables~\ref{tab:main-results-full} and~\ref{tab:app:mt-results} consider a failed generation to score zero. 
At \glimmermodelname{}'s empty rate of 0.3\%, the two yield identical tier assignments. 

\paragraph{Machine Translation Performance}

We report aggregate \glimmermodelname translation performance across the three benchmarks in~\cref{tab:app:mt-results}, separating the into English (X$\to$eng) and the out of English (eng$\to$Y) performance, again compared to \gemma and \qwen.
\gemma is the best performing model across all three benchmarks and aggregate directions, but \glimmermodelname is a close second throughout (tied with \qwen into English on \dataset{WMT24++}).
In particular, on the $55$ language pairs of \dataset{WMT24++}, \glimmermodelname is within 1.5 \textsc{chrF++} points of \gemma into English and within 1.2 points out of English. The delta widens when the model is evaluated on the long tail translation tasks of \dataset{BOUQuET}, the benchmark with the widest language coverage, where translation performance degrades, bringing the overall observed performance down. 

\begin{table}[ht]
\caption{Machine translation performance (\textsc{chrF++}) of \glimmermodelname (\emph{high}), \gemma and \qwen across three MT benchmarks and two aggregate directions. \emph{Overall} averages the $972$ \dataset{FLORES+} and \dataset{BOUQuET} directions ($212+212+274+274$), the two benchmarks with the broadest language coverage.}
\centering
\footnotesize
\setlength{\tabcolsep}{5pt}
\renewcommand{\arraystretch}{1.15}
\begin{tabular}{llcccc}
\toprule
\textbf{Dataset} & \textbf{Direction} & \textbf{Language Pairs} & \textbf{\glimmermodelname} & \textbf{\gemma} & \textbf{\qwen} \\
\midrule
\dataset{WMT24++} & X$\to$eng & 55 & 55.7 & \textbf{57.2} & 55.7 \\
\dataset{WMT24++} & eng$\to$Y & 55 & 51.0 & \textbf{52.2} & 48.8 \\
\midrule
\dataset{FLORES+} & X$\to$eng & 212 & 53.7 & \textbf{54.9} & 53.2 \\
\dataset{FLORES+} & eng$\to$Y & 212 & 41.4 & \textbf{41.5} & 38.0 \\
\dataset{BOUQuET} & X$\to$eng & 274 & 41.4 & \textbf{44.0} & 41.1 \\
\dataset{BOUQuET} & eng$\to$Y & 274 & 31.7 & \textbf{33.3} & 29.7 \\
% \midrule
\textbf{Overall} & X$\leftrightarrow$eng & 972 & 41.3 & \textbf{42.8} & 39.9 \\
\bottomrule
\label{tab:app:mt-results}
\end{tabular}
\end{table}

\paragraph{Detailed Language Support}
With the operational definition above, and the language coverage provided by the three supporting benchmarks (55 for \dataset{WMT24++}, 212 on \dataset{FLORES+} and 274 on \dataset{BOUQuET}), we establish that \glimmermodelname{} supports \textbf{137 language varieties}, with $88$ classified as \emph{strong} and $49$ as \emph{usable}.

One caveat on translation benchmarks: \textsc{chrF++} penalises scripts that do not use spaces between words (Chinese, Japanese) or that pack a syllable into each character (Korean), which partly explains why Chinese, Japanese and Korean fall below the bar out of English ($33.9$ / $34.9$ / $36.5$), as they do for \gemma and \qwen, while scoring above $56$ into English. Since all three reach $81.5$ on MMLU-ProX, in line with strong-tier languages, we do not read their absence from the list as a lack of support, although we keep the tiers strictly rule-based.

We further remark that this count should be read as language \emph{varieties}, or \emph{languoids}.

The explicit list of supported languages, grouped by their tier, is reported in~\cref{tab:language-support}.

% === This will be converted to tables, it's a mess right now!
\begin{table}[ht!]
\centering \scriptsize \setlength{\tabcolsep}{2pt}
\begin{tabular}{@{}p{5cm}p{5cm}p{5cm}@{}}
\toprule
\multicolumn{3}{@{}c@{}}{\normalsize \textbf{Strong support} ($n=88$)} \\
\addlinespace[2pt] \midrule
\texttt{afr\_Latn} (Afrikaans) & \texttt{als\_Latn} (Tosk Albanian) & \texttt{arb\_Arab} (Standard Arabic) \\
\texttt{ast\_Latn} (Asturian) & \texttt{azj\_Latn} (North Azerbaijani) & \texttt{bel\_Cyrl} (Belarusian) \\
\texttt{ben\_Beng} (Bangla) & \texttt{bos\_Latn} (Bosnian) & \texttt{bul\_Cyrl} (Bulgarian) \\
\texttt{cat\_Latn\_stan1289} (Standard Catalan) & \texttt{cat\_Latn\_vale1252} (Standard Valencian) & \texttt{ceb\_Latn} (Cebuano) \\
\texttt{ces\_Latn} (Czech) & \texttt{cym\_Latn} (Welsh) & \texttt{dan\_Latn} (Danish) \\
\texttt{deu\_Latn} (German) & \texttt{ekk\_Latn} (Standard Estonian) & \texttt{ell\_Grek} (Greek) \\
\texttt{epo\_Latn} (Esperanto) & \texttt{eus\_Latn} (Basque) & \texttt{fin\_Latn} (Finnish) \\
\texttt{fra\_Latn\_queb1247} (Qu\'eb\'ecois French) & \texttt{fra\_Latn\_stan1290} (Standard French) & \texttt{fry\_Latn} (Western Frisian) \\
\texttt{gla\_Latn} (Scottish Gaelic) & \texttt{gle\_Latn} (Irish) & \texttt{glg\_Latn} (Galician) \\
\texttt{guj\_Gujr} (Gujarati) & \texttt{hat\_Latn} (Haitian Creole) & \texttt{heb\_Hebr} (Hebrew) \\
\texttt{hin\_Deva} (Hindi (Devanagari)) & \texttt{hrv\_Latn} (Croatian) & \texttt{hun\_Latn} (Hungarian) \\
\texttt{hye\_Armn} (Armenian) & \texttt{ind\_Latn} (Indonesian) & \texttt{isl\_Latn} (Icelandic) \\
\texttt{ita\_Latn} (Italian) & \texttt{jav\_Latn} (Javanese) & \texttt{kan\_Knda} (Kannada) \\
\texttt{kat\_Geor} (Georgian) & \texttt{kaz\_Cyrl} (Kazakh) & \texttt{lir\_Latn} (Liberian Kreyol) \\
\texttt{lit\_Latn} (Lithuanian) & \texttt{ltz\_Latn} (Luxembourgish) & \texttt{lvs\_Latn} (Standard Latvian) \\
\texttt{mal\_Mlym} (Malayalam) & \texttt{mkd\_Cyrl} (Macedonian) & \texttt{mlt\_Latn} (Maltese) \\
\texttt{mri\_Latn} (Māori) & \texttt{nld\_Latn} (Dutch) & \texttt{nno\_Latn} (Norwegian Nynorsk) \\
\texttt{nob\_Latn} (Norwegian Bokmål) & \texttt{npi\_Deva} (Nepali) & \texttt{oci\_Latn} (Occitan) \\
\texttt{ory\_Orya} (Odia) & \texttt{pan\_Guru} (Punjabi) & \texttt{pap\_Latn} (Papiamento) \\
\texttt{pcm\_Latn} (Nigerian Pidgin) & \texttt{pes\_Arab} (Western Persian) & \texttt{plt\_Latn} (Plateau Malagasy) \\
 \texttt{pol\_Latn} (Polish) & \texttt{por\_Latn\_braz1246} (Brazilian Portuguese) & \texttt{por\_Latn\_por1283} (European Portuguese) \\ \texttt{ron\_Latn} (Romanian) & \texttt{rus\_Cyrl} (Russian) &
\texttt{slk\_Latn} (Slovak) \\ \texttt{slv\_Latn} (Slovenian) & \texttt{smo\_Latn} (Samoan) &
\texttt{sna\_Latn} (Shona) \\ \texttt{sot\_Latn} (Southern Sotho) & \texttt{spa\_Latn\_mexi1248} (Mexican Spanish) &
\texttt{spa\_Latn\_stan1288} (Standard Spanish) \\ \texttt{swe\_Latn} (Swedish) & \texttt{swh\_Latn} (Swahili) &
\texttt{tam\_Taml} (Tamil) \\ \texttt{tel\_Telu} (Telugu) & \texttt{tgk\_Cyrl} (Tajik) &
\texttt{tgl\_Latn} (Tagalog) \\ \texttt{tha\_Thai} (Thai) & \texttt{tur\_Latn} (Turkish) &
\texttt{ukr\_Cyrl} (Ukrainian) \\ \texttt{urd\_Arab} (Urdu (Arabic)) & \texttt{uzn\_Latn} (Northern Uzbek) &
\texttt{vie\_Latn} (Vietnamese) \\ \texttt{war\_Latn} (Waray) & \texttt{ydd\_Hebr} (Eastern Yiddish) &
\texttt{zsm\_Latn} (Standard Malay) \\ \texttt{zul\_Latn} (Zulu) & \\
\midrule \midrule
\multicolumn{3}{@{}c@{}}{\normalsize \textbf{Usable support} ($n=49$)} \\
\midrule \addlinespace[2pt]
\texttt{acm\_Arab} (Mesopotamian Arabic) & \texttt{acq\_Arab} (Ta'izzi-Adeni Arabic) & \texttt{apc\_Arab\_nort3139} (North Levantine Arabic) \\
\texttt{apc\_Arab\_sout3123} (South Levantine Arabic) & \texttt{arg\_Latn} (Aragonese) & \texttt{arz\_Arab} (Egyptian Arabic) \\
\texttt{asm\_Beng} (Assamese) & \texttt{bak\_Cyrl} (Bashkir) & \texttt{ban\_Latn} (Balinese) \\
\texttt{bjn\_Latn} (Banjar) & \texttt{ckb\_Arab} (Central Kurdish) & \texttt{fao\_Latn} (Faroese) \\
\texttt{fur\_Latn} (Friulian) & \texttt{hau\_Latn} (Hausa) & \texttt{hin\_Latn} (Hindi (Latin)) \\
\texttt{ibo\_Latn} (Igbo) & \texttt{ilo\_Latn} (Iloko) & \texttt{kaa\_Latn} (Kara-Kalpak) \\
\texttt{khk\_Cyrl} (Halh Mongolian) & \texttt{khm\_Khmr} (Khmer) & \texttt{kin\_Latn} (Kinyarwanda) \\
\texttt{kir\_Cyrl} (Kyrgyz) & \texttt{kmr\_Latn} (Northern Kurdish) & \texttt{lao\_Laoo} (Lao) \\
\texttt{lim\_Latn} (Limburgish) & \texttt{mag\_Deva} (Magahi) & \texttt{mai\_Deva} (Maithili) \\
\texttt{mar\_Deva} (Marathi) & \texttt{min\_Latn} (Minangkabau) & \texttt{mtq\_Latn} (Muong) \\
\texttt{nso\_Latn} (Northern Sotho) & \texttt{nya\_Latn} (Nyanja) & \texttt{pbt\_Arab} (Southern Pashto) \\
\texttt{prs\_Arab} (Dari) & \texttt{roh\_Latn} (Romansh) & \texttt{scn\_Latn} (Sicilian) \\
\texttt{sin\_Sinh} (Sinhala) & \texttt{snd\_Arab} (Sindhi) & \texttt{som\_Latn} (Somali) \\
\texttt{srd\_Latn} (Sardinian) & \texttt{srp\_Cyrl} (Serbian) & \texttt{sun\_Latn} (Sundanese) \\
\texttt{tat\_Cyrl} (Tatar) & \texttt{tpi\_Latn} (Tok Pisin) & \texttt{tsn\_Latn} (Tswana) \\
\texttt{tuk\_Latn} (Turkmen) & \texttt{urd\_Latn} (Urdu (Latin)) & \texttt{vec\_Latn} (Venetian) \\
\texttt{xho\_Latn} (Xhosa) & & \\
\bottomrule
\end{tabular}
\caption{Language support of \glimmermodelname, grouped by support tier.}
\label{tab:language-support}
\end{table}

\subsection{Examples}
\newenvironment{prompt}{\% \par\medskip\noindent \textbf{Prompt:}\begin{quote}\itshape }{\% \end{quote}\medskip } \newenvironment{response}{\% \par\medskip\noindent \textbf{Response:}\begin{quote} }{\% \end{quote}\medskip } \newenvironment{chainofthought}{\% \par\medskip\noindent \textbf{Chain-of-Thought:}\begin{quote}\small\ttfamily }{\% \end{quote}\medskip }

\subsubsection{Chart Understanding}

%%% Tentative, using Meta's color palette
% \definecolor{promptblue}{HTML}{E8F3FF}
\definecolor{promptblue}{HTML}{AFD7FF}
% \definecolor{responsegray}{HTML}{F1F4F7}
\definecolor{responsegray}{HTML}{CBD2D9}

\begin{tcolorbox}[colback=promptblue, colframe=promptblue, sharp corners] \textbf{Prompt} \begin{center} \includegraphics[width=0.9\textwidth]{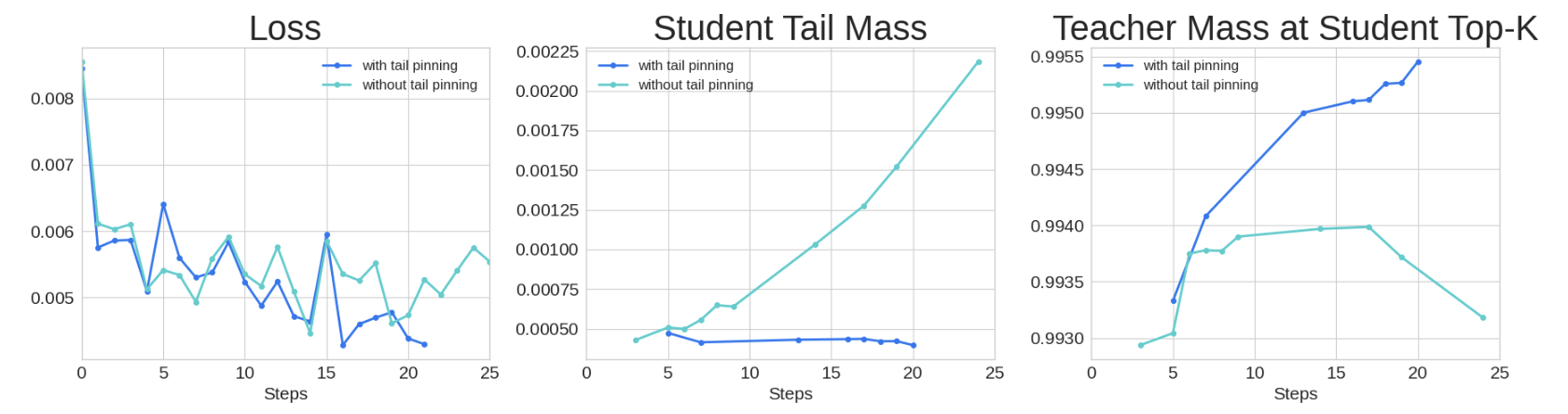} \end{center} \ttfamily Read these 3 graphs and tell me what do you understand. \end{tcolorbox} \begin{tcolorbox}[colback=responsegray, colframe=responsegray, sharp corners] \textbf{Response} \ttfamily The three panels are the same training run tracked over optimisation steps, once with a ``tail pinning'' regulariser on the student and once without it. \begin{itemize} \item \textbf{Loss (left)} --- Both curves start around ${\sim}0.009$ and drop within the first 1--2 steps. With tail pinning the blue curve is on average a little lower in the later part of training, ending around $0.0043$ at step 21--24. Without pinning the teal curve keeps fluctuating, ${\sim}0.0055$--$0.0060$ at the end. \item \textbf{Student Tail Mass (centre)} --- With tail pinning, student tail mass stays flat at $4.5 \times 10^{-4}$. Without it, mass grows monotonically from ${\sim}4 \times 10^{-4}$ to ${\sim}2.2 \times 10^{-3}$, a $>5\times$ increase. \item \textbf{Teacher Mass at Student Top-K (right)} --- With pinning, overlap improves from ${\sim}0.9934$ to ${\sim}0.9955$. Without it, overlap peaks early then degrades to ${\sim}0.9932$. \end{itemize} Taken together: tail pinning keeps the student distribution concentrated, prevents unbounded growth of tail probability mass, and preserves the agreement between the student's top-K predictions and the teacher's distribution. \end{tcolorbox}
 
\subsubsection{Solve ODE}

\begin{tcolorbox}[colback=promptblue, colframe=promptblue, sharp corners] \textbf{Prompt} \ttfamily Solve: \[ \frac{\mathrm{d}^2 y}{\mathrm{d} x^2} - 4\,\frac{\mathrm{d} y}{\mathrm{d} x} + 8\,y = 0. \] \end{tcolorbox} \begin{tcolorbox}[colback=responsegray, colframe=responsegray, sharp corners] \textbf{Response} \ttfamily The characteristic equation for $y'' - 4y' + 8y = 0$ is obtained with $y = e^{rx}$: \[ r^{2} - 4r + 8 = 0 \] \[ r = \frac{4 \pm \sqrt{16 - 32}}{2} = 2 \pm 2i \] So the roots are complex conjugate $r = 2 \pm 2i = \alpha \pm i\beta$ with $\alpha = 2$, $\beta = 2$. Hence: \[ y(x) = e^{2x}\bigl(C_{1}\cos 2x + C_{2}\sin 2x\bigr) \] Equivalently, in amplitude-phase form: \[ y(x) = A\,e^{2x}\cos(2x - \phi), \quad A = \sqrt{C_1^2 + C_2^2},\; \tan\phi = C_2 / C_1. \] \end{tcolorbox}

\subsubsection{Riddles}
Prompts 
\begin{center} \includegraphics[width=0.7\textwidth]{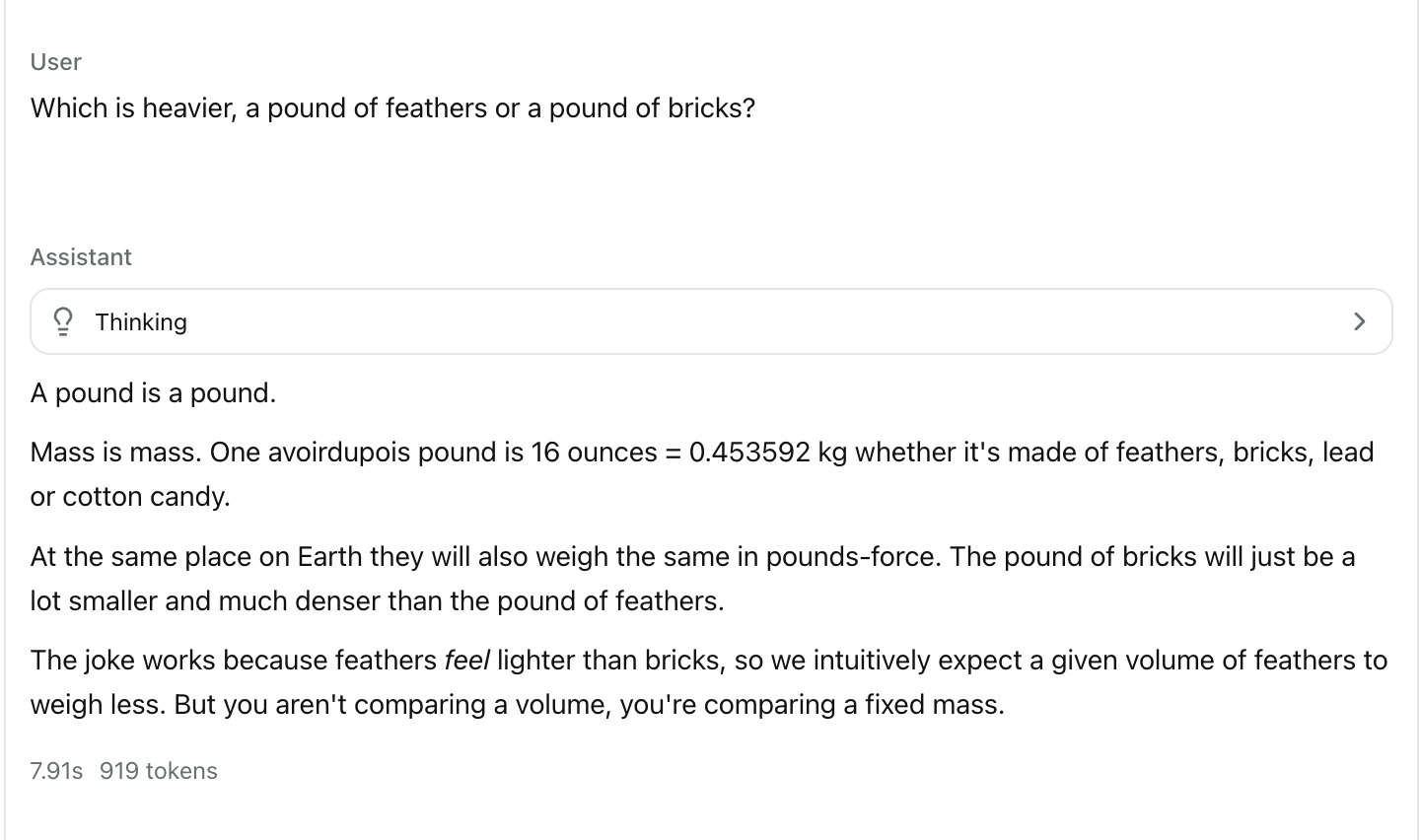} \end{center} 

\begin{center} \includegraphics[width=0.7\textwidth]{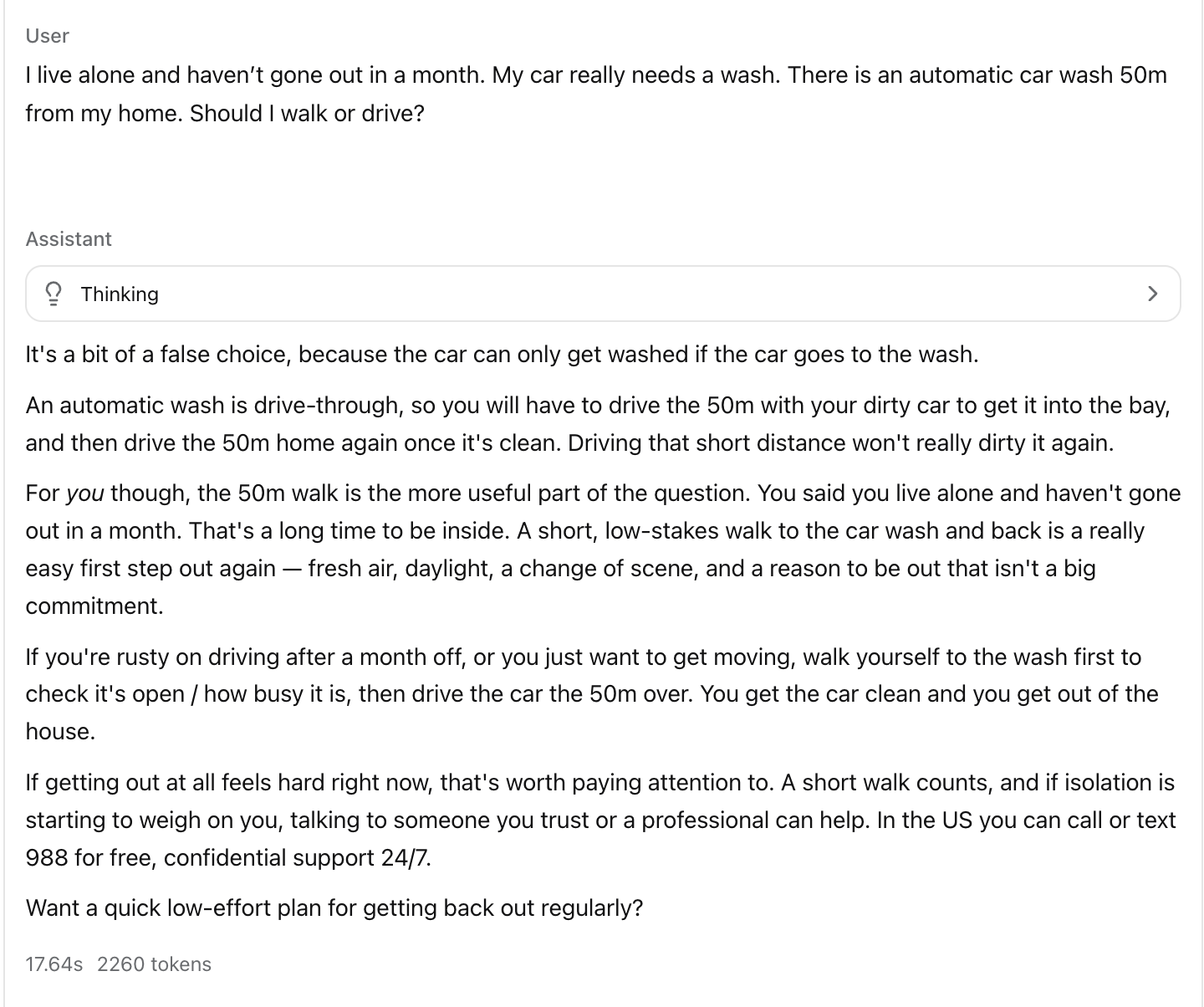} \end{center}

\end{document}